\documentclass{ieeetj}
\usepackage{cite}
\usepackage{pifont}
\usepackage{amsmath,amssymb,amsfonts}
\usepackage{algorithm}
\usepackage{algorithmic}
\usepackage{graphicx} 
\usepackage{textcomp}
\usepackage[table,xcdraw,dvipsnames,svgnames]{xcolor} 
\usepackage{hyperref}
\hypersetup{hidelinks=true}
\usepackage{caption}
\usepackage{makecell}
\usepackage{booktabs}
\usepackage{multirow}
\usepackage{arydshln}
\usepackage{changes}
\usepackage{comment}
\usepackage{float}
\usepackage{siunitx}
\usepackage{pdfpages}
\def\BibTeX{{\rm B\kern-.05em{\sc i\kern-.025em b}\kern-.08em
    T\kern-.1667em\lower.7ex\hbox{E}\kern-.125emX}}
\AtBeginDocument{\definecolor{tmlcncolor}{cmyk}{0.93,0.59,0.15,0.02}\definecolor{NavyBlue}{RGB}{0,86,125}}

\excludecomment{draft}

\def\OJlogo{\vspace{-4pt}$<$Society logo(s) and publication title will appear here.$>$}
\def\seclogo{\vspace{10pt}$<$Society logo(s) and publication title will appear here.$>$}
\def\authorrefmark#1{\ensuremath{^{\textbf{#1}}}}

\begin{document}
\receiveddate{XX Month, XXXX}
\reviseddate{XX Month, XXXX}
\accepteddate{XX Month, XXXX}
\publisheddate{XX Month, XXXX}
\currentdate{XX Month, XXXX}
\doiinfo{XXXX.2022.1234567}

\markboth{}{Author {et al.}}

\title{Explore From Sketch: Accelerating UAV Exploration in Large-scale Environments with Prior Maps}

\author{Tiancheng Lai\authorrefmark{1,2}, Yuman Gao\authorrefmark{1,4}, Xiangyu Li\authorrefmark{1,2}, Ruitian Pang\authorrefmark{1,2}, Xingpeng Wang\authorrefmark{1,2}, Siqi Shen\authorrefmark{1,2}, Mengke Zhang\authorrefmark{1,2}, Yin He\authorrefmark{3}, Zaisheng Pan\authorrefmark{1}, Fei Gao\authorrefmark{1,4}, Chao Xu\authorrefmark{1,2}, and Yanjun Cao\authorrefmark{1,2}}
\affil{Institute of Cyber-Systems and Control, College of Control Science and Engineering, Zhejiang University, Hangzhou 310027, China}
\affil{Huzhou Institute, Zhejiang University, and Huzhou Key Laboratory of Autonomous System, Huzhou 313000, China.}
\affil{Zhejiang Zhongyan Industry Co., Ltd, Hangzhou 310000, China.}
\affil{Differential Robotics Technology Company, Hangzhou 311100, China.}
\corresp{Corresponding author: Yanjun Cao (email: yanjunhi@zju.edu.cn), Chao Xu (email: cxu@zju.edu.cn), and Fei Gao (email: fgaoaa@zju.edu.cn).}

\authornote{Tiancheng Lai and Yuman Gao contributed equally to this work. \\ \\ This work was supported by the Key R\&D Project of China National Tobacco Corporation under Grant No. 110202402018, the National Key R\&D Program of China under Grant No. 2023YFB4706600, the Zhejiang Provincial Science and Technology Plan Project under Grant No. 2024C01170, and the National Natural Science Foundation of China under Grant No. 62322314.}

\begin{abstract}
  Autonomous exploration with unmanned aerial vehicles (UAVs) in large-scale, topologically complex environments often suffers from low efficiency due to suboptimal scheduling and detours.
  Prior maps, such as satellite imagery, construction drawings, and hand-drawn maps, are readily available in many scenarios and encode the global structure to support long-horizon decision-making.
  This paper presents a novel framework that leverages sparse, unaligned, and even discrepant 2D prior maps for LiDAR-based UAV exploration.
  Firstly, using a binary top-down sketch map as input, it registers the prior online against accumulating LiDAR observations, and the resulting registration uncertainty is then carried through to planning.
  Registration begins with GeoContext, a geometric descriptor designed to retrieve single-frame candidates by capturing the spatial relationships between 3D single-frame measurements and the prior sketch across the 2D–3D modality gap, without requiring any initial alignment.
  These candidates are aggregated by multi-frame verification, which filters outliers and establishes coarse transformations, and a Scale-ICP process refines the results.
  Together, these stages tolerate map discrepancies and retain multiple hypotheses when the geometry is ambiguous.
  Secondly, a hierarchical viewpoint planning strategy then acts on these hypotheses, attaching local viewpoints spatially to the prior guidepoints and sequencing those guidepoints under each hypothesis with a Monte Carlo Tree Search solver.
  A risk-aware selector then ranks the resulting sequences by a heuristic confidence-weighted travel cost, and a fixed-endpoint traveling salesman problem generates an efficient local coverage path under the selected prior guidance.
  Benchmark evaluations reveal up to 34.2\% improvement in exploration efficiency and 37.9\% reduction in flight distance compared to state-of-the-art methods, while extensive simulations and field experiments further demonstrate robustness to prior map incompleteness and deformations.

\end{abstract}

\begin{IEEEkeywords}
Autonomous exploration, unmanned aerial vehicles, field robotics, search and rescue.
\end{IEEEkeywords}

\IEEEspecialpapernotice{(Regular Paper)}

\maketitle

\begin{figure*}[!h]
  \centering
  \includegraphics[width=1.0\textwidth]{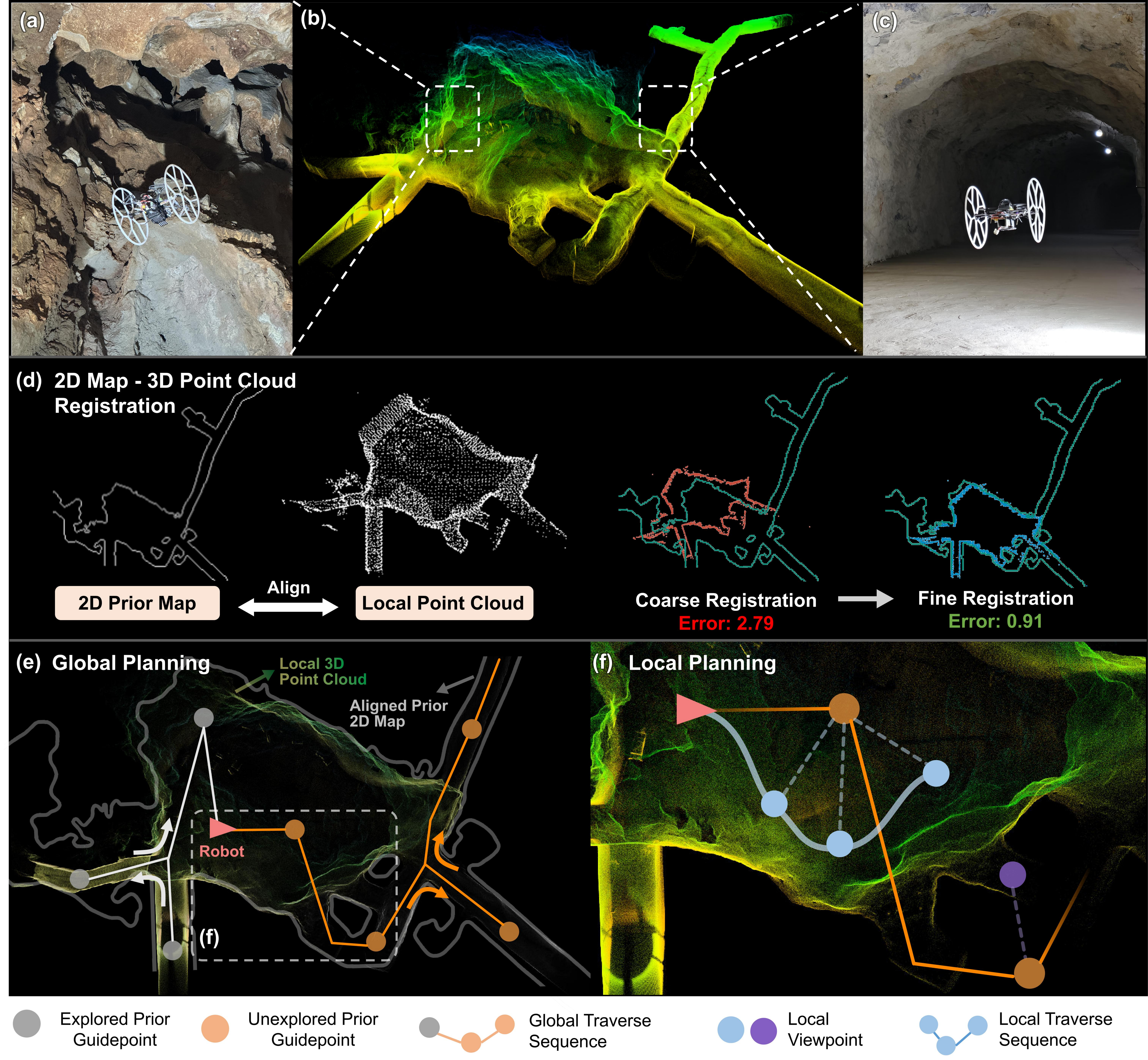}
\caption{\textbf{Graphical Abstract.}
(a)(c) Snapshots of the UAV exploring the wild cave; (b) The overall point cloud map of the environment; (d) the registration framework that aligns a 2D prior map with 3D local point cloud data; (e)(f) The hierarchical viewpoint planning framework, where global planning (e) determines the global guidance of the exploration process, and local planning (f) determines the local viewpoints traversal sequence based on the global guidance.
The light blue local viewpoints are attached to the next prior guidepoint based on their spatial relationships and will be considered in the local planning, while other viewpoints are excluded.}
  \label{fig:cover}
\end{figure*}

\section{INTRODUCTION}

\IEEEPARstart{A}{utonomous} exploration using unmanned aerial vehicles (UAVs) is a challenging task, especially in large and complex field environments.
To support real-world applications like search-and-rescue or inspection missions, UAVs need to explore and map the entire environment efficiently due to the strict constraints of their limited flight endurance.
However, exploration efficiency is often limited by unknown global topology, inherently causing suboptimal trajectories and prolonged mission times.
Meanwhile, with the continuous advancement of satellite technology and data-sharing platforms, prior maps of many environments are becoming readily accessible.
Such maps—e.g., satellite imagery, construction drawings, and floor plans—although often imprecise and flawed, contain valuable boundary and geometric information that can provide global guidance to improve exploration efficiency.

However, most existing exploration frameworks, particularly frontier-based~\cite{zhang2024falcon,geng2025epic} and next-best-view (NBV)~\cite{bircher2016receding} methods, rely solely on the robot's current sensory observations to generate local exploration guidance, resulting in inefficient exploration paths.
This limitation worsens as the environment scales up, complicating the planning of globally efficient trajectories~\cite{selin2019efficient, cao2023representation, zhou2023racer}.
Several recent studies~\cite{feng2024fc, bai2024graph, luperto2020robot} have attempted to incorporate prior maps into the exploration process.
However, these approaches typically assume that prior maps are either dense and accurate or perfectly aligned with the real-world environment, which is often impractical.
For example, in disaster-damaged environments or when dealing with outdated datasets, acquiring accurate models is difficult.
Therefore, achieving efficient autonomous exploration with flawed prior maps remains an open problem, with primary challenges as follows:

\begin{itemize}
    \item \textbf{Heterogeneous Data and Map Discrepancies:} The informational mismatch between 2D priors and 3D LiDAR scans makes data alignment inherently difficult.
    Prior maps provide abstracted geometric boundaries, while LiDAR captures dense yet local details.
    This becomes more challenging when the prior maps are incomplete or deformed.
    \item \textbf{Localization Ambiguity:} Similar topological structures often cause localization uncertainty, particularly during early exploration stages when local observations are limited.
    Multiple hypotheses should be generated if necessary.
    \item \textbf{Prior-guided Exploration:} The planner must convert uncertain registration results into global guidance while still generating executable local viewpoints, which remains a rarely addressed problem.
\end{itemize}

This paper presents a novel framework that effectively leverages sparse, unaligned and potentially discrepant prior maps to accelerate LiDAR-based UAV exploration in large-scale and complex environments.
The framework consumes such a prior as a binary top-down traversability map of free space and boundaries, the common form to which satellite imagery, construction drawings, and hand-drawn sketches can all be formulated.
An overview is provided in Fig.~\ref{fig:cover}: the framework includes a robust 2D-3D point cloud registration module that aligns LiDAR observations with the prior map (Fig.~\ref{fig:cover}(d)).
First, to address the modality differences, we design a GeoContext descriptor, which extracts consistent geometric features to retrieve single-frame candidates that reveal the spatial connections between the 2D prior and 3D LiDAR data.
These intra-frame candidates are then aggregated via a multi-frame verification mechanism to establish coarse transformations while filtering outliers.
Ultimately, a Scale-ICP process refines these transformations, outputting multiple registration hypotheses.

To boost the exploration process based on the registration results, we propose a prior-guided hierarchical viewpoint planning strategy to generate efficient local paths under registration uncertainties.
First, prior guidepoints are sparsely sampled from the prior map.
  For each transformation result, local viewpoints are spatially attached to prior guidepoints, and a Monte Carlo Tree Search (MCTS) algorithm is then utilized to generate an optimal traversal sequence of prior guidepoints (Fig.~\ref{fig:cover}(e)).
  Subsequently, a risk-aware selector identifies the most reliable prior sequence using a heuristic confidence-weighted travel cost, which then serves as the global guidance for local planning.
  To efficiently guide the robot's traversal under the global guidance, a Fixed-endpoint Traveling Salesman Problem (FE-TSP) is formulated to generate an efficient local coverage path under the selected prior guidance (Fig.~\ref{fig:cover}(f)).

To summarize, the primary contributions of this paper are as follows:

(1) We propose a novel framework that leverages sparse, unaligned, and even discrepant prior maps to guide autonomous exploration, significantly improving efficiency in large-scale environments.

(2) We develop a robust 2D-3D registration framework that aligns LiDAR observations with 2D prior maps while accounting for potential discrepancies.
This enables reliable localization under sparse and imperfect priors and supports multiple registration hypotheses.

(3) We propose a hierarchical viewpoint planning strategy, which generates efficient local paths under global guidance.

(4) We evaluate the framework in registration benchmarks, exploration benchmarks, simulations, and field experiments, covering precise, partial, and deformed prior maps.

\section{RELATED WORK}

This section reviews works related to our approach, including autonomous exploration in large-scale environments, localization against prior maps, and prior-map-integrated exploration.

\subsection{AUTONOMOUS EXPLORATION IN LARGE-SCALE ENVIRONMENTS}

Autonomous exploration in large-scale environments requires both efficient traversal path planning and scalable environment representation~\cite{selin2019efficient, cao2023representation, zhou2023racer}.

Among various approaches, the frontier-based method is widely adopted to guide robots by targeting boundaries between known and unknown spaces.
The concept was first introduced by Yamauchi~\cite{yamauchi1997frontier}, whose method selects the nearest frontier.
Recent research focuses on optimizing the local coverage path to maximize exploration efficiency.
Zhao et al.~\cite{zhao2023tdle} propose an exploration system that arranges the visitation order of uniformly and dynamically divided subregions to provide spatial guidance.
Zhou et al.~\cite{zhou2021fuel} propose a TSP-based planner to optimize the coverage path visiting all frontiers.
Zhang et al.~\cite{zhang2024falcon} improve the generation of the coverage path by taking topological connectivity into consideration.
Some works~\cite{bone2023decentralised, 11122320} utilize the MCTS method to address more complex sequential planning problems like multi-robot and bimodal vehicle exploration.

To achieve scalable environment representation~\cite{cao2023representation}, Dang et al.~\cite{dang2020graph} propose a hierarchical graph approach by constructing a sparse global map to represent the environment and an Exploring Random Graph to compute local exploration paths.
Batinovic et al.~\cite{batinovic2021multi} propose an OctoMap representation of the environment that supports incremental updates, thus reducing computational burden.
Yang et al.~\cite{yang2021graph} and Gao et al.~\cite{gao2022meeting} further employ simpler geometric representations like convex polyhedra and star-convex polyhedra.

Meanwhile, LiDAR-based exploration methods exploit long-range geometric measurements for accurate mapping in large-scale environments and achieve strong performance.
Kaufman et al.~\cite{kaufman2018Lidar_explore} adopt multiple depth sensors including a 2D LiDAR to generate accurate occupancy probabilities for grids.
Dharmadhikari et al.~\cite{dharmadhikari2020Lidar_explore} propose a computationally efficient volumetric representation of the environment to generate motion primitives for fast navigation.
Cao et al.~\cite{cao2021tare} first propose a criterion to evaluate whether a viewpoint is valid based on the scanning range and bearing from the viewpoint to the point.
Geng et al.~\cite{geng2025epic} further improve this criterion to construct an observation map and generate frontiers to guide exploration.

Although LiDAR-based perception shows great potential for large-scale exploration, existing methods often rely solely on local sensory observations to generate local exploration paths and fail to incorporate prior information into the exploration process. This inevitably causes myopic decision-making and inefficient path generation, especially in large-scale and complex environments.

\subsection{LOCALIZATION AGAINST PRIOR MAP}

Existing approaches for localization against prior maps include probabilistic filtering methods such as Monte Carlo Localization (MCL) and Bayesian filtering, supervised learning (SL), and feature-based methods.

MCL methods usually use particle filtering to localize the robot.
Matsuo et al.~\cite{matsuo2012outdoor} combine SLAM with particle swarm optimization for map refinement to achieve effective outdoor localization.
The works in~\cite{boniardi2015sketch,foroughi2019indoor} simultaneously estimate local deformation of prior maps to achieve better localization.
Xu et al.~\cite{xu2024robot} propose a Bayesian filtering method to obtain the robot's pose by matching the prior map and the local occupancy predictor built from the robot's panoramic RGB observation.
The global heuristic vector is then calculated via grid search for navigation.
However, these methods are often confined to simple, small-scale indoor environments, as they suffer from prohibitive computational costs and are difficult to deploy in real time when the environment is large and high registration accuracy is required. They are also sensitive to initialization and prone to local minima, which further limits their scalability.

For SL methods, Chen et al.~\cite{chen2020localizing} use a CNN to predict control points to compute the transformation between the robot's map and the prior map via spline-based registration.
Shah et al.~\cite{shah2011robust,shah2013qualitative} propose a quadratic programming method to compute waypoints by modeling spatial relationship of landmarks and user-defined paths.
Such methods require paired training data with ground-truth alignment or human-labeled landmarks, and tend to overfit to the training data.
Moreover, they cannot perform local-to-global alignment, which our setting requires in order to obtain useful guidance early in the exploration.

Regarding feature-based methods, Qiao et al.~\cite{qiao2025speak} extract triangle descriptors from line and corner features and adopt the Hough transform to match triplets via hierarchical voting.
This design presumes the accurate and structured geometry of a Building Information Modeling model with known scale, and thus degrades on sketches that are unstructured, incomplete, or deformed.
Another common approach is to project the 3D point cloud into 2D images and perform image-to-image registration.
Some supervised learning methods can extract consistent features across different modalities.
For example, SuperGlue~\cite{sarlin2020superglue} leverages a graph neural network to extract and match geometric features, maintaining robustness across different modalities.
Its keypoint-level matching, however, lacks global spatial consistency, so a large observed portion is required before the local-to-global alignment succeeds.
In addition, certain cross-modality image matching methods~\cite{tuzcuouglu2024xoftr, chen2023shape} are ineffective in this task, as they tend to overfit to appearance cues such as color and texture, which are absent in prior maps.

\begin{figure*}[htbp]
  \centering
  \includegraphics[width=1.0\textwidth]{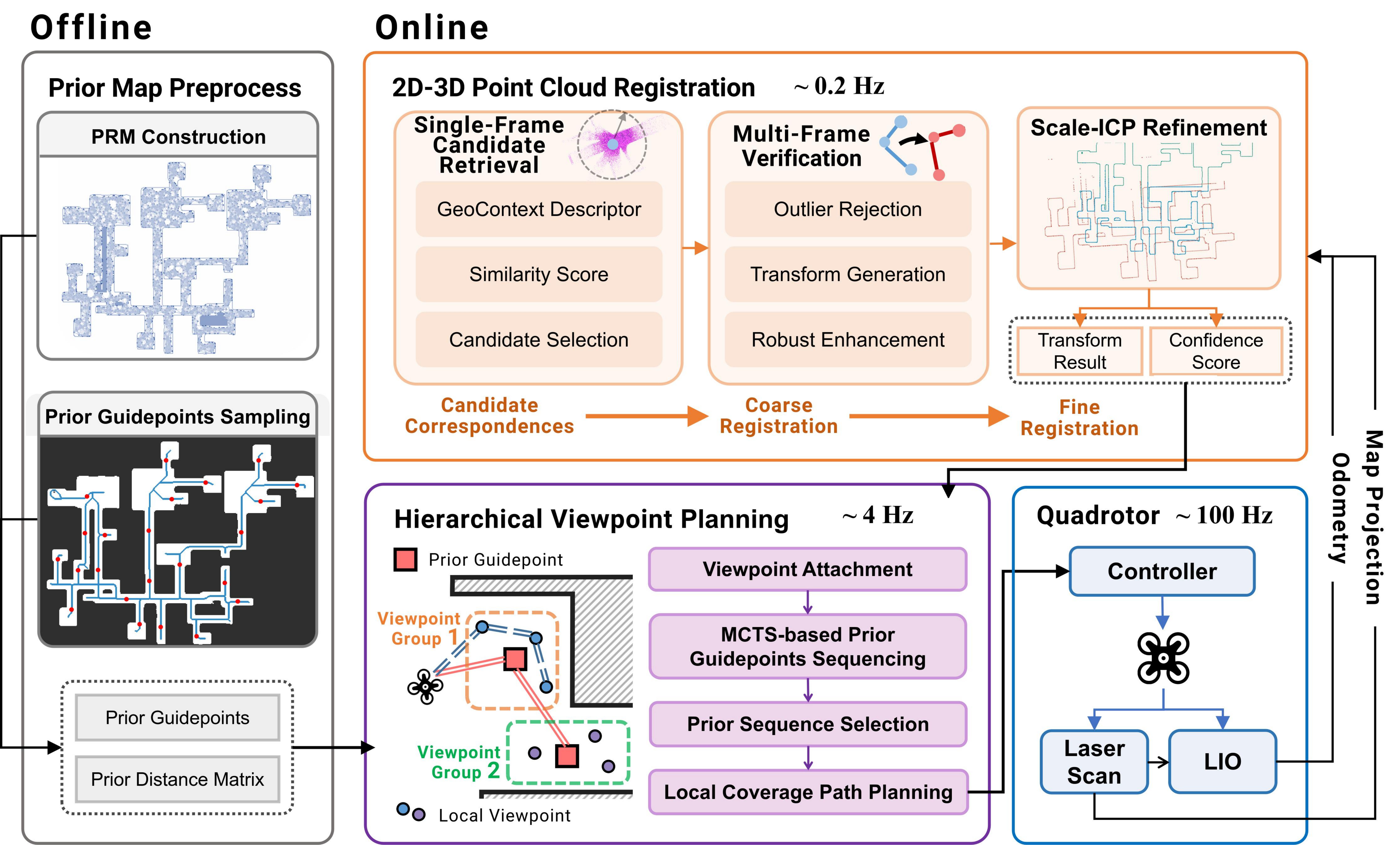}
  \caption{System overview.
The framework (1) preprocesses the prior map, (2) registers online LiDAR observations to the prior map, (3) plans prior-guided local viewpoints under registration uncertainty, and (4) executes the resulting UAV trajectories.}
  \label{fig:framework}
\end{figure*}

\subsection{PRIOR-MAP-INTEGRATED EXPLORATION}

Several recent works have attempted to integrate prior maps into the exploration process.
Feng et al.~\cite{feng2024fc} propose a reconstruction framework that generates viewpoints from an input point cloud via skeleton-based space decomposition and produces a high-quality coverage path by dividing the planning problem into sub-problems.
However, this method requires accurate and complete 3D prior models, which can be challenging to obtain in practice.
Bai et al.~\cite{bai2024graph} propose a SLAM-aware exploration framework that leverages prior topo-metric graphs to enhance both exploration efficiency and pose graph reliability.
By actively selecting globally informative loop closures with a theoretically pruned greedy algorithm, it improves mapping accuracy without sacrificing exploration speed.
However, this method emphasizes SLAM robustness and fails to accelerate the exploration process under the prior guidance.
Luperto et al.~\cite{luperto2020robot} present an online frontier-based exploration strategy that exploits incomplete and inaccurate prior knowledge (e.g., floor plans, footprints) to improve robot exploration performance by better estimating information gain at candidate locations.
Though demonstrating certain effectiveness and robustness, this method also assumes that the prior maps are perfectly aligned with the real-world environment.
Moreover, this greedy viewpoint selection method only improves exploration efficiency in the early stages, but requires significant time to cover the full environment thereafter.
These limitations motivate an exploration framework that can register sparse, imperfect 2D priors online and use uncertain registration hypotheses for global-to-local planning.
\section{PROBLEM DEFINITION AND SYSTEM OVERVIEW}

\subsection{PROBLEM DEFINITION}
\begin{table}[htbp]
\centering
\resizebox{\columnwidth}{!}{%
\begin{tabular}{lp{0.75\columnwidth}}
\toprule
\textbf{Symbol} & \textbf{Description} \\
\midrule
\multicolumn{2}{c}{\textit{Spaces, Maps \& Point Clouds}} \\
\midrule
$\mathcal{M}_{\text{pri}}$, $\mathcal{M}_{\text{obs}}$ & 2D prior sketch map and local observation map \\
$\mathcal{P}$, $\mathcal{P}_i$ & Visible point cloud in the current frame, and its subset in the $i$-th azimuthal bin \\
$X_{\text{r}}$ & Robot pose node \\
\midrule
\multicolumn{2}{c}{\textit{Descriptors \& Registration}} \\
\midrule
$\mathbf{I}^{p}, \mathbf{I}^{o}$ & GeoContext descriptors for the prior and local maps \\
$\mathcal{X}_{\text{cand}}$ & Set of candidate locations sampled on the traversable regions of the prior map \\
$\mathbf{f}^{p}, \mathbf{f}^{o}$ & Low-frequency components of the descriptors extracted via FFT \\
$\mathbf{R}, \mathbf{t}, s$ & Rotation matrix, translation vector, and scaling factor \\
$\mathcal{T}_f, T$ & Set of fine-registered transformations, and each transformation with confidence score \\
\midrule
\multicolumn{2}{c}{\textit{Guidepoints \& Viewpoints \& Nodes}} \\
\midrule
$\mathcal{G}$, $\tilde{\mathcal{G}}$, $\tilde{G}_i$ & Original prior guidepoints, transformed prior guidepoints and the $i$-th transformed prior guidepoint \\
$\mathcal{V}$, $V_i$  & Set of sampled local viewpoints and the $i$-th local viewpoint \\
$\mathcal{V}_{\text{iso}}, \mathcal{V}_{\text{target}}$ & Sets of isolated local viewpoints and target local viewpoints \\
\midrule
\multicolumn{2}{c}{\textit{Costs \& Objectives}} \\
\midrule
$L(\cdot, \cdot)$ & Searched path length between two nodes \\
$C_{\text{risk}}(i)$ & Risk score evaluating the expected extra travel length for the $i$-th viewpoint group \\
$J_{\text{urg}}(i)$ & Urgency score for prioritizing the visitation of the $i$-th viewpoint \\
\bottomrule
\end{tabular}%
}
\caption{Summary of Key Notations.}
\label{tab:notation}
\end{table}

We address the problem of efficiently exploring an unknown bounded 3D space $\mathcal{W}\subset\mathbb R^{3}$ using a LiDAR-equipped UAV, given a prior map $\mathcal M_{\text{pri}}$.
The prior map is assumed to be an unaligned 2D sketch that illustrates the boundaries of the environment.
The UAV should generate feasible and efficient trajectories under the guidance of the prior map, aiming to explore the whole space and generate a dense point cloud map $\mathcal M_{\text{obs}}$ in minimum time.

\subsection{SYSTEM OVERVIEW}
The proposed exploration framework, depicted in Figure~\ref{fig:framework}, comprises three main modules: prior map pre-processing (Section~\ref{sec:prior_map_preprocessing}), 2D-3D point cloud registration (Section~\ref{sec:registration}), and prior-guided hierarchical viewpoint planning (Section~\ref{sec:path_planning}).
The offline pre-processing extracts and stores prior guidepoints and their distance matrix for online planning.
During the exploration process, the registration module continuously aligns the prior map with the real-world environment.
Given the estimated alignment, the hierarchical viewpoint planning module determines optimal guidepoint sequences and generates efficient local viewpoint coverage paths under prior guidance.
The exploration terminates when no local viewpoints remain in the environment.
The main notations used in the paper are summarized in Table~\ref{tab:notation}.

\section{PRIOR MAP PRE-PROCESSING}\label{sec:prior_map_preprocessing}

In offline pre-processing, we generate prior guidepoints and compute distances among them using the prior map.
The prior guidepoints are sampled from the topological skeleton extracted from traversable areas, and the distance matrix between every two guidepoints is calculated via a Probabilistic Roadmap (PRM) and stored for efficient querying during online planning.
The implementation of the traversable mask, the skeleton thinning, the roadmap construction, and the distance matrix is detailed in Section~S1 of the supplementary material.

\begin{figure}[b]
  \centering
  \includegraphics[width=\linewidth]{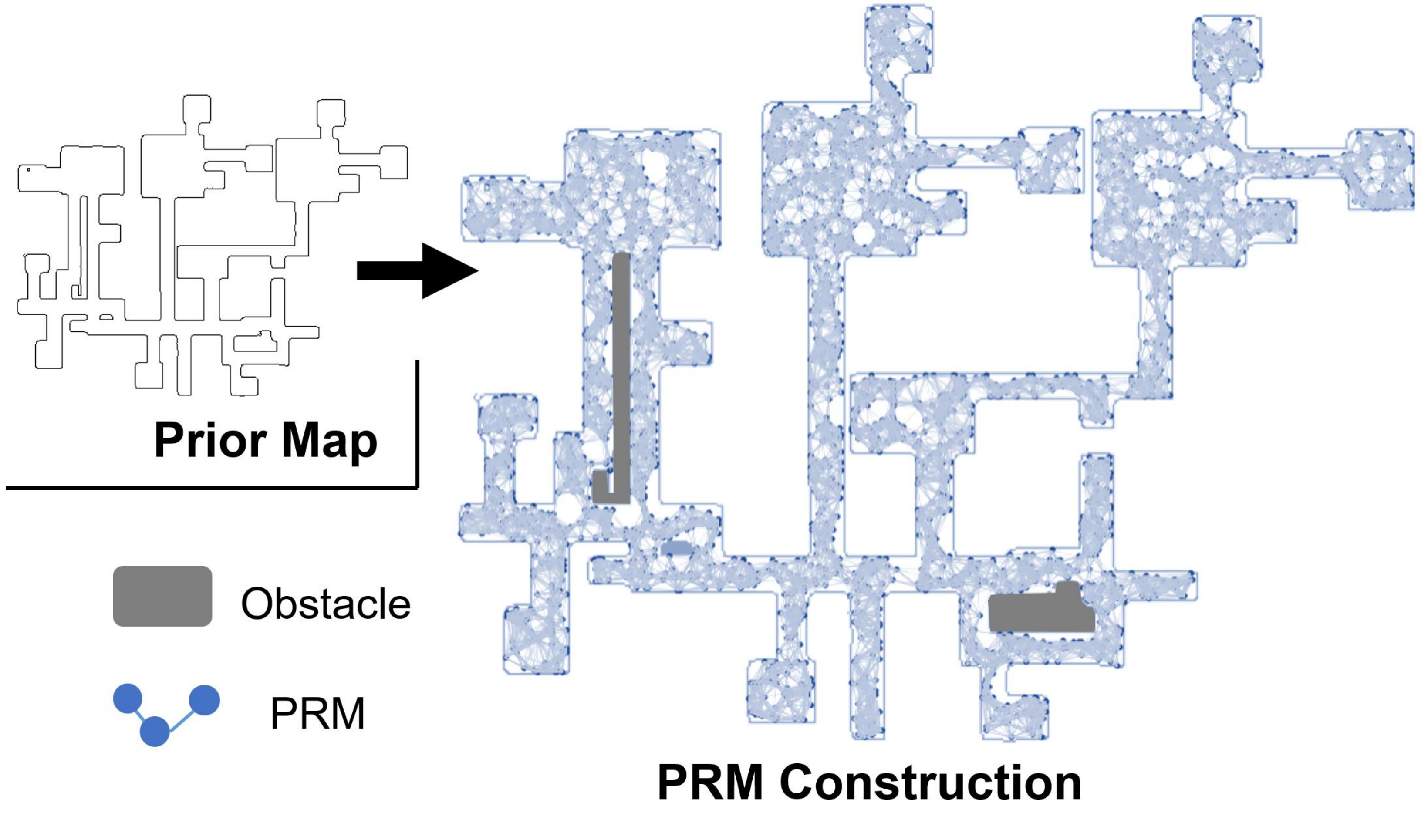}
  \caption{PRM construction in traversable areas of the prior map.}
  \label{fig:prm}
\end{figure}

\subsection{PRM CONSTRUCTION}\label{sec:prm_construction}

\begin{figure}[t]
  \centering
  \includegraphics[width=0.98\linewidth]{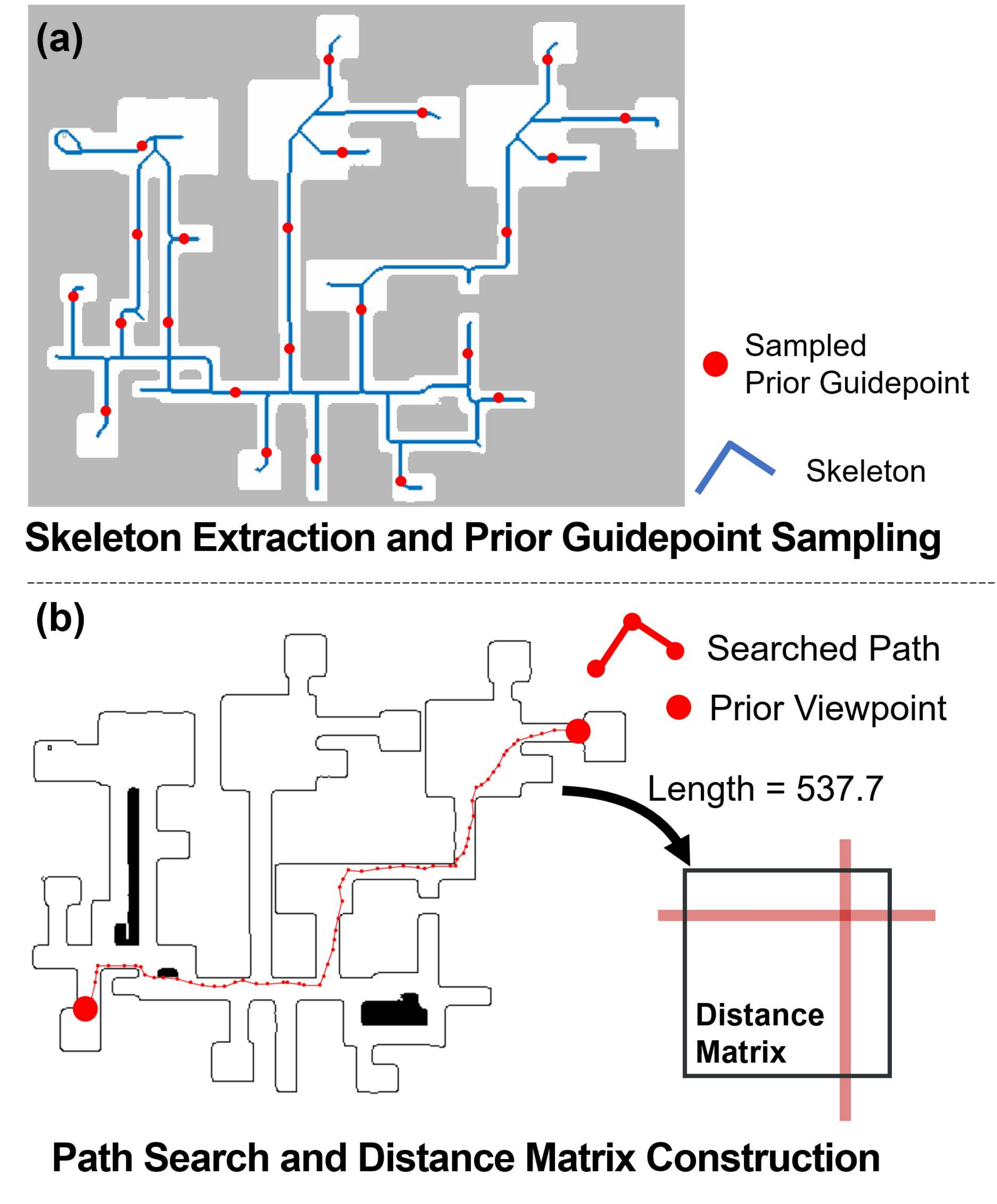}
  \caption{(a) Skeleton extraction and prior guidepoint sampling. (b) Path search and distance matrix construction.}
  \label{fig:prior_map_preprocessing}
\end{figure}

We construct a PRM~\cite{kavraki1998analysis} to represent the prior map and compute the prior distances between prior guidepoints (Figure~\ref{fig:prm}).
First, traversable regions of the prior map are extracted.
We assume that the provided prior map is a binary image, where the traversable areas are surrounded by black boundary pixels.
Then we build an undirected graph $\Gamma = (V_p, E_p)$ to represent the roadmap, where $V_p$ is the set of nodes on the traversable areas and $E_p$ is the set of collision-free edges connecting these nodes.
The nodes in $V_p$ are sampled uniformly within the traversable region of the prior map.
For every $v\in V_p$, we connect it to its $k$ nearest neighbors if the edge between them is collision-free.
The weight of each edge is defined as the Euclidean distance between the two nodes it connects.

\subsection{TOPOLOGY EXTRACTION AND PRIOR GUIDEPOINT GENERATION}\label{sec:topology_extraction}

Prior guidepoints are generated based on the topology of the prior map (Figure~\ref{fig:prior_map_preprocessing}(a)).
The skeleton of the prior map is extracted using the morphology module in the scikit-image library~\cite{van2014scikit}, where the Zhang-Suen thinning algorithm~\cite{zhang1984fast} is applied.
Once the skeleton is extracted, prior guidepoints are sampled along it guided by two key considerations to ensure effective information extraction.
First, prior guidepoints should be sampled sparsely to offer high-level topological guidance without compromising local optimality.
In practice, we set the sampling interval to around 2-4 times the mapping range of the robot.
Second, prior guidepoints should capture key topological structures, such as new branches or independent regions, while omitting non-critical ones like intersections or skeletal endpoints that lack directional guidance for exploration.

\subsection{DISTANCE MATRIX CONSTRUCTION}
Having constructed the graph and generated the prior guidepoints, we compute the prior distances between guidepoints using Dijkstra's algorithm~\cite{dijkstra1959note} (Figure~\ref{fig:prior_map_preprocessing}(b)).
Each guidepoint is mapped to its nearest node in the graph and the distance between any two guidepoints is defined as the shortest distance between the corresponding nodes.
These distances are precomputed and stored in a matrix for efficient online querying.

\section{2D-3D POINT CLOUD REGISTRATION}\label{sec:registration}

We propose a registration module that aligns 2D prior maps with 3D point clouds, recovering both the scale and the rigid transformation.
Since the prior map is restricted to a horizontal plane, we decouple the vertical axis and restrict the transformation into 2D space (i.e., $\mathbf{R} \in SO(2)$ and $\mathbf{t} \in \mathbb{R}^2$).
The method consists of single-frame candidate retrieval, multi-frame geometric verification, and Scale-ICP refinement.
In this context, a \textit{frame} refers to a specific timestamp at which both the sensory input and the robot pose are sampled. These frames are extracted at a low frequency (e.g., \qty{0.2}{\hertz}) to provide sparse snapshots of the environment.
The module only requires a rough scale guess $s_g$ to filter unreasonable correspondences initially.
The main algorithm is shown in Algorithm~\ref{alg:registration}.

\begin{algorithm}
  \caption{2D-3D Point Cloud Registration}
  \begin{algorithmic}[1]

  \REQUIRE $\mathcal{M}_\text{pri}$: 2D Prior Map, $\mathcal{M}_\text{obs}$: Online Observation Map, $X_{r}$: Robot Position, $\mathcal{X}_{\text{cand}}$: Sampled Location Sets, $s_g$: Scale Guess
  \ENSURE $\mathcal{T}_f$: Refined Transformations

  \STATE // \textbf{Phase 1: Descriptor Pre-computation}
  \STATE $\mathcal{F}_\text{pri} \gets \{\}$
  \FOR{each sampled location $x_{j}$ in $\mathcal{X}_{\text{cand}}$}
      \STATE $\mathcal{M}_{\text{visible}, j} \gets \textsc{HPR}(\mathcal{M}_\text{pri}, x_j)$
      \STATE $\mathbf{d}_{\text{pri}, j} \gets \textsc{calcDesc}(\mathcal{M}_{\text{visible}, j})$
      \STATE $\mathcal{F}_\text{pri}.\text{insert}(\mathbf{d}_{\text{pri}, j})$
  \ENDFOR

  \STATE // \textbf{Phase 2: Online Registration}
  \STATE $\mathcal{P}_\text{robot}, \mathcal{C}, \mathcal{T}_\text{pot} \gets \{\}$

  \WHILE{receive $\mathcal{M}_\text{obs}$ and $X_r$}
    \STATE $\mathcal{P}_\text{robot}.\text{insert}(X_r)$
    \STATE $\mathcal{M}_\text{visible} \gets \textsc{HPR}(\mathcal{M}_\text{obs}, X_r)$
    \STATE $\mathbf{d}_{i} \gets \textsc{calcDesc}(\mathcal{M}_\text{visible})$
    \STATE $C_{i} \gets \textsc{FindCand}(\mathcal{F}_\text{pri}, \mathbf{d}_{i})$
    \STATE $\mathcal{C}.\text{insert}(C_{i})$
    \IF{$\mathcal{C}.\mathtt{size}() < 3$}
        \STATE \textbf{continue}
    \ELSIF{$\mathcal{C}.\mathtt{size}() == 3$}
        \STATE $\mathcal{T}_\text{pot} \gets \textsc{GetPotTran}(\mathcal{C}, \mathcal{P}_\text{robot}, s_g)$
    \ELSIF{$\mathcal{C}.\mathtt{size}() > 3$}
        \STATE $\mathcal{T}_\text{pot} \gets \textsc{FilterOutliers}(\mathcal{T}_\text{pot}, \mathcal{C}, \mathcal{P}_\text{robot})$
    \ENDIF
    \STATE $\mathcal{T}_f \gets \textsc{Scale-ICP}(\mathcal{T}_\text{pot}, \mathcal{M}_\text{obs}, \mathcal{M}_\text{pri})$
\ENDWHILE
  \end{algorithmic}
  \label{alg:registration}
\end{algorithm}

\subsection{SINGLE-FRAME CANDIDATE RETRIEVAL}\label{sec:single_frame_registration}

We introduce GeoContext, a descriptor that captures numerical and sequential geometric patterns shared by the 2D prior map and the 3D LiDAR map.
Unlike point-wise matching descriptors such as Scan Context~\cite{kim2018scan}, GeoContext emphasizes boundary-level sequential distributions, improving robustness to noise and map discrepancies.

\begin{figure}[htbp]
  \centering
  \includegraphics[width=\linewidth]{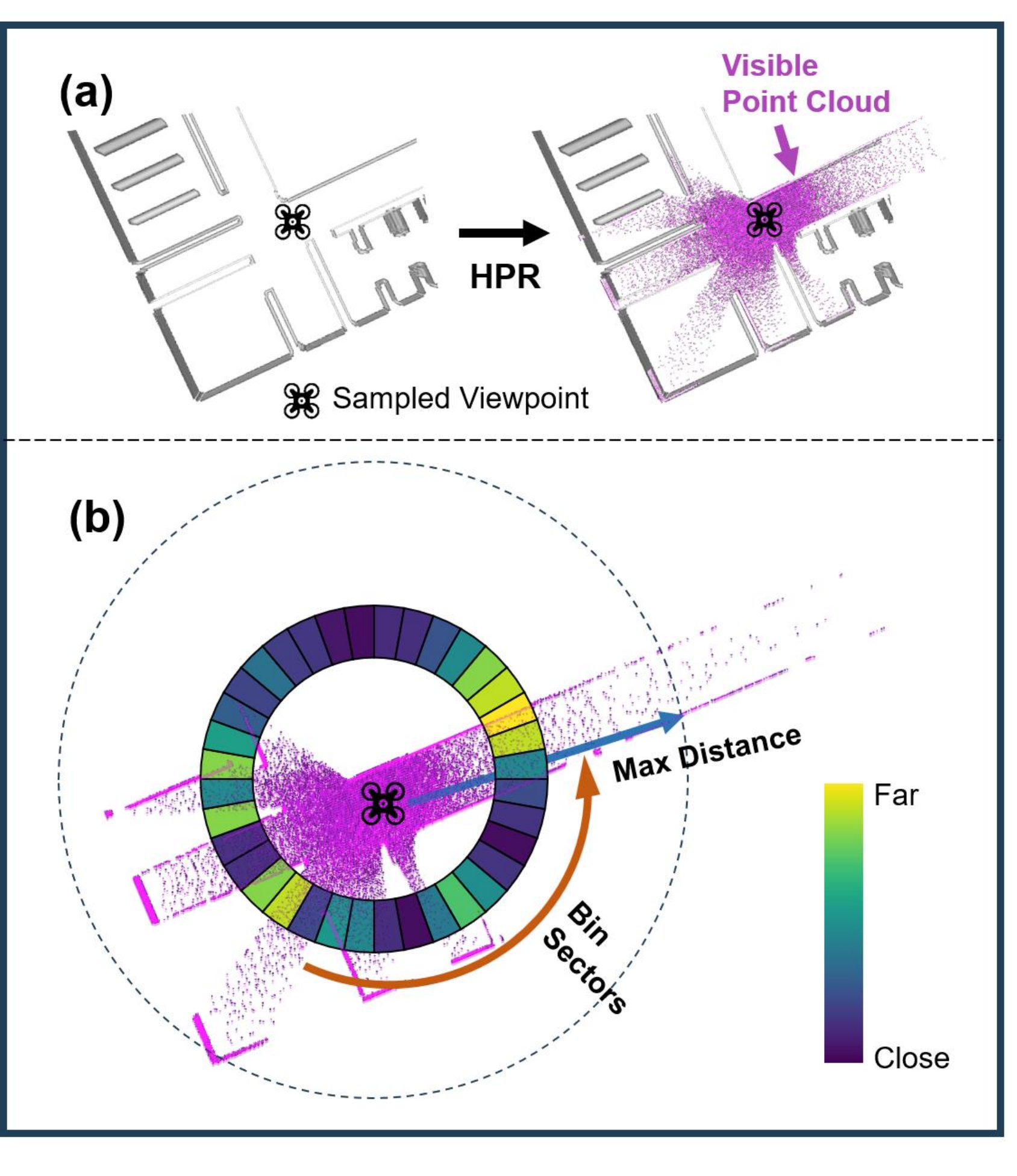}
  \caption{Illustration of the extraction of the GeoContext descriptor.
  The purple point cloud represents the visible points extracted from the HPR process.
  The 2D descriptor is generated by dividing the points into azimuthal bins and recording the maximum distance in each bin.}
  \label{fig:scancontext_2d}
\end{figure}

\subsubsection{GeoContext Descriptor}\label{sec:scancontext_2d}

Given either a 2D or 3D point cloud, we first divide the points into azimuthal bins (from 0 to $2\pi$) in the sensor's polar coordinate system.
Within each azimuthal bin, we record the maximum horizontal distance from the robot center to the points.
Let $\mathcal{P}$ be the set of visible points expressed in the robot's local sensor frame, and $\mathcal{P}_i$ the set of points belonging to the $i$-th azimuthal bin.
The partition can be written as
\begin{equation}
\mathcal{P} = \bigcup_{i=1}^{N_b} \mathcal{P}_i, \quad \mathcal{P}_i \cap \mathcal{P}_j = \emptyset, \quad i \neq j,
\end{equation}
where $N_b$ is the number of azimuthal bins.

Thus, the $i$-th element of the GeoContext descriptor $\mathbf{I}$ is computed as
\begin{equation}
\mathbf{I}_i = \max_{\mathbf{p} \in \mathcal{P}_{i}} \|\Pi_{xy}(\mathbf{p})\|_2,
\end{equation}
where $\Pi_{xy}(\cdot)$ denotes the projection onto the horizontal plane.
If no points fall within the $i$-th azimuthal bin (i.e., $\mathcal{P}_i = \emptyset$), we set $\mathbf{I}_i = 0$.
The GeoContext descriptor $\mathbf{I}$ is a 1D vector with $N_b$ elements.
By recording the maximum distance in each azimuthal direction, GeoContext reduces the influence of local obstacles and approximates boundary-level structure.

Given the prior map $\mathcal{M}_\text{pri}$, we first convert it into a point cloud by extracting its black pixels, and uniformly sample a set of candidate locations $\mathcal{X}_{\text{cand}}$ with a sampling interval of $d_{\text{gap}}$ within the traversable regions of $\mathcal{M}_\text{pri}$ (Figure~\ref{fig:scancontext_2d}).
Each location is then taken as a virtual viewpoint, from which we extract the visible subset of the point cloud using the Hidden Point Removal (HPR) strategy implemented via the Star Convex Polygon (SCP) algorithm~\cite{zhong2020generating} (denoted $\textsc{HPR}(\cdot)$, Figure~\ref{fig:scancontext_2d}(a)) and compute its GeoContext descriptor (denoted $\textsc{calcDesc}(\cdot)$, Figure~\ref{fig:scancontext_2d}(b)).
During exploration, historical LiDAR scans are accumulated to form a local point cloud map.
We similarly apply $\textsc{HPR}(\cdot)$ and $\textsc{calcDesc}(\cdot)$ at the current robot pose to generate the descriptor for subsequent matching.
Section~S4 of the supplementary material gives a full account of the candidate-location database and of the GeoContext descriptor, together with a controlled study of retrieval accuracy under increasing prior-map deformation.

\subsubsection{Similarity Score of Descriptors}

\begin{figure}
  \centering
  \includegraphics[width=0.97\linewidth]{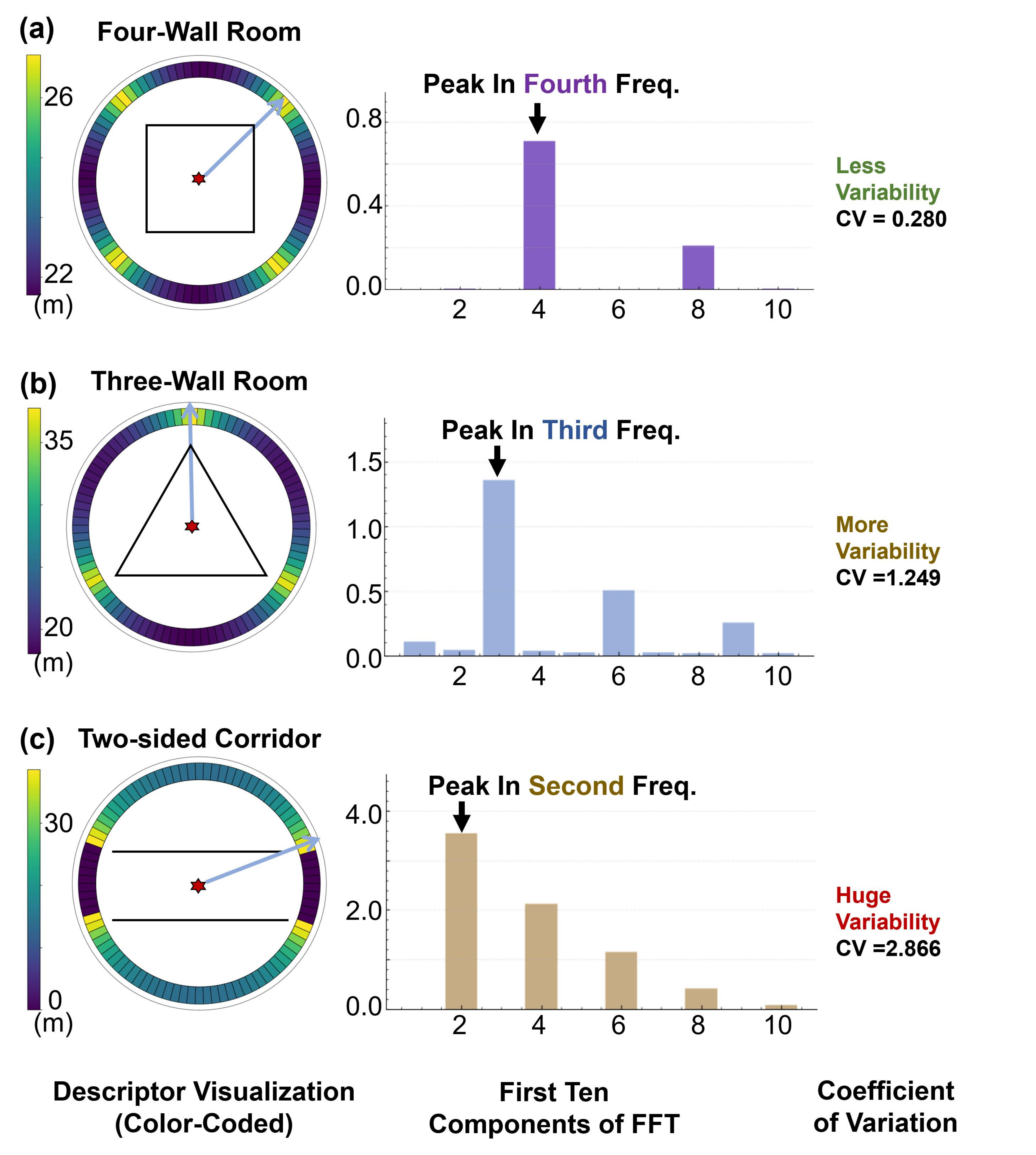}
  \caption{Visualization of the sequential similarity metrics.
  The left column shows the environment and color-coded descriptors, the middle column visualizes the distribution of the FFT similarity metric, and the right column shows the CV similarity metric.}
  \label{fig:similarity}
\end{figure}

Given a pair of GeoContext descriptors, we construct a similarity measure between the corresponding locations.
We define the prior map as the point cloud map extracted from the prior sketch, and the observation map as the local point cloud map generated from the LiDAR scans.
Let non-zero $\mathbf{I}^{p}$ and $\mathbf{I}^{o}$ be the GeoContext descriptors of the prior map and the observation map, respectively.
The cosine similarity introduced in~\cite{kim2018scan} represents the numerical similarity of two environments.
Unlike the standard cosine similarity, we adopt an inverse variant that assigns larger weights to smaller values, since larger values (corresponding to distant features) tend to be noisier and less stable:
\begin{equation}
\begin{aligned}
    S_{\text{inv\_cos}}(\mathbf{I}^{p}, \mathbf{I}^{o}) &= \frac{\sum_{i} w_i (\mathbf{I}^{p})_i (\mathbf{I}^{o})_i}{\sqrt{\sum_{i} w_i (\mathbf{I}^{p})_i^2} \sqrt{\sum_{i} w_i (\mathbf{I}^{o})_i^2}}, \\
    w_i &= \frac{1}{|(\mathbf{I}^{p})_i| + |(\mathbf{I}^{o})_i|},
\end{aligned}
\end{equation}
where $w_i$ is the weight for the $i$-th azimuthal bin, ensuring that the elements with larger values are given less weight.

Similar to Scan Context, we also calculate similarities with all possible column-shifted descriptors to find the maximum one.
Let $\mathbf{I}_{n}^{o}$ be a column-shifted version of $\mathbf{I}^o$ with $n$ bins.
The final inverse cosine similarity between two descriptors is defined as:
\begin{equation}
S_{\text{cos}}(\mathbf{I}^{p}, \mathbf{I}^{o}) = \max_{n} S_{\text{inv\_cos}}(\mathbf{I}^{p}, \mathbf{I}_{n}^{o}).
\end{equation}

To capture the sequential spatial structure of the environment, we also measure the similarity between the two descriptors in the frequency domain.
We first apply the Fast Fourier Transform (FFT) to the two descriptors and extract $N_c$ low-frequency components.
Let $\mathbf{f}^{p}$ and $\mathbf{f}^{o}$ denote the extracted magnitudes of the low-frequency components of $\mathbf{I}^{p}$ and $\mathbf{I}^{o}$, respectively, both of which are $N_c \times 1$ vectors.
The similarity of two descriptors in the frequency domain is defined as:
\begin{equation}
S_{\text{FFT}}(\mathbf{f}^{p}, \mathbf{f}^{o}) = \frac{1}{2} \left( \frac{\mathbf{f}^{p} \cdot \mathbf{f}^{o}}{|\mathbf{f}^{p}|_2 |\mathbf{f}^{o}|_2} + 1 \right).
\end{equation}

The low-frequency components abstract the rough, stable spatial layout.
Due to the circular shift property of FFT, a column-shift of the original descriptor alters only the phase of these components, leaving their magnitudes unchanged and thereby enabling orientation-independent matching.

We further capture the overall distribution of the environment using the Coefficient of Variation (CV) of the two descriptors, defined as the ratio of the standard deviation to the mean.
The CV of a descriptor $\mathbf{I}$ can be written as:
\begin{equation}
CV(\mathbf{I}) = \frac{\sigma(\mathbf{I})}{\mu(\mathbf{I})},
\end{equation}
where $\sigma(\cdot)$ and $\mu(\cdot)$ are the standard deviation and mean functions, respectively.
The similarity of two descriptors in terms of CV is defined as:
\begin{equation}
S_{\text{cv}}(\mathbf{I}^{p}, \mathbf{I}^{o}) = 1 - \frac{|CV^p - CV^o|}{CV^p + CV^o}.
\end{equation}

The CV value is also scale-invariant and captures the overall spatial feature distribution.
As illustrated in Figure~\ref{fig:similarity}, it effectively distinguishes restricted small-CV room-like environments (with more uniform point depths) from large-CV corridor-like structures.

We calculate a composite similarity by combining the three similarity metrics via their weighted product:

\begin{equation}
    S(\mathbf{I}^{p}, \mathbf{I}^{o}) = S_{\text{cos}}(\mathbf{I}^{p}, \mathbf{I}^{o})^{\alpha_1} \cdot S_{\text{FFT}}(\mathbf{f}^{p}, \mathbf{f}^{o})^{\alpha_2}\cdot S_{\text{cv}}(\mathbf{I}^{p}, \mathbf{I}^{o})^{\alpha_3},
\end{equation}
where $\alpha_1$, $\alpha_2$, and $\alpha_3$ are weights.
This score jointly accounts for descriptor alignment, low-frequency spatial layout, and distributional variation.

We rank all candidate locations by their similarity scores to the observation-map descriptor.
The top-ranked locations, with a proportion denoted by the candidate rate $\eta$, are aggregated to form the raw candidate set $C_{\text{raw}}$.
Because candidates often cluster densely in regions with similar local geometry, we spatially downsample $C_{\text{raw}}$ to form the final candidate set $C_{\text{final}}$.
This yields the correspondence set for the $i$-th frame $Q^i = \{(C_{\text{final}}^{i}, \mathbf{p}^{i})\}$, where $\mathbf{p}^{i}$ is the 2D coordinate of the robot's current position.
The $j$-th correspondence in the $i$-th frame is denoted as $q_j^i = (\mathbf{c}_j^i, \mathbf{p}^i)$, where $\mathbf{c}_j^i \in C_{\text{final}}^i$.
The whole selection process is referred to as $\textsc{FindCand}(\cdot)$.

\subsection{MULTI-FRAME VERIFICATION MECHANISM}\label{sec:multi_frame_registration}

Given candidates across multiple frames, we introduce a multi-frame verification mechanism based on Fourier Descriptors (FD) that matches candidates to identify potential transformations and filter outliers.
We also incorporate a robustness enhancement strategy to handle failed single-frame results.

\subsubsection{Potential Transformation Estimation}\label{sec:potential_transformation_estimation}

We define the transformation $T_{p\rightarrow r}=(\mathbf{R},\mathbf{t},s)$ as a mapping from a point in the prior map $\mathbf{c}$ to a point in the observation map $\mathbf{p}$:
$\mathbf{p}=s\mathbf{R}\mathbf{c}+\mathbf{t}$.
A potential transformation can be computed in closed form when at least three correspondences are established; for more correspondences, we optimize the transformation via least squares.

When correspondences from three frames are available, we select the frames with the lowest and second-lowest number of candidates as the first and second anchor frames, respectively.
The anchor frames are chosen to minimize the combinatorial complexity of triplet generation.
We then connect every candidate of the two frames to form a set of potential correspondence pairs $P = \{(q_{i}^{1}, q_{j}^{2})\}$ where $q_{i}^{1} \in Q^{1}$ and $q_{j}^{2} \in Q^{2}$.
For every correspondence pair, we can calculate the scale factor $\tilde{s}$ as
\begin{equation}
\tilde{s} = \frac{\|\mathbf{p}^{1} - \mathbf{p}^{2}\|_2}{\|\mathbf{c}_{i}^{1} - \mathbf{c}_{j}^{2}\|_2},
\end{equation}
where $\mathbf{c}_{i}^{1}$ and $\mathbf{c}_{j}^{2}$ are candidate coordinates in the prior map, and $\mathbf{p}^{1}$ and $\mathbf{p}^{2}$ are the corresponding robot positions.
Given an initial scale guess $s_g$, we can filter out invalid correspondences with significant scale deviations.
For each correspondence pair, the scale estimate $\tilde{s}$ is then used as the baseline for selecting the third frame's correspondences.

For each correspondence pair, we then connect every candidate of the third frame to form a set of potential correspondence triplets $Q = \{(q_{i}^{1}, q_{j}^{2}, q_{k}^{3})\}$ where $q_{i}^{1} \in Q^{1}$, $q_{j}^{2} \in Q^{2}$ and $q_{k}^{3} \in Q^{3}$.
For every correspondence triplet, we compute a potential transformation $T = (\mathbf{R}, \mathbf{t}, s)$ in closed form.
The rotation matrix $\mathbf{R}$, translation vector $\mathbf{t}$, and scaling factor $s$ are given by:
\begin{equation}
\begin{aligned}
\mathbf{R} &= \arg\min_{\mathbf{R}\in SO(2)} \sum_{i=1}^{3}
\| \mathbf{R}(\mathbf{c}_i-\bar{\mathbf{c}})-(\mathbf{p}_i-\bar{\mathbf{p}})\|_2^2,\\
s &= \frac{\sum_{i=1}^{3}(\mathbf{p}_i-\bar{\mathbf{p}})^{T}\mathbf{R}(\mathbf{c}_i-\bar{\mathbf{c}})}
{\sum_{i=1}^{3}\|\mathbf{c}_i-\bar{\mathbf{c}}\|_2^2},\\
\mathbf{t} &= \bar{\mathbf{p}}-s\mathbf{R}\bar{\mathbf{c}}.
\end{aligned}
\end{equation}
where $\bar{\mathbf{p}}$ and $\bar{\mathbf{c}}$ are the centroids of the point set $\mathbf{p}_i$ and $\mathbf{c}_i$, respectively.
Likewise, we reject the triplet as an outlier if its computed scale factor $s$ deviates significantly from the previous estimate.
Otherwise, we add this transformation to the potential transformation set $\mathcal{T}_{\text{raw}}$. This process is denoted as function $\textsc{GetPotTran}(\cdot)$.

\subsubsection{Results Update and Outlier Filtering}\label{sec:results_update_and_outlier_filtering}

As more correspondences are established, we remove outliers by comparing the FD of the geometric shapes they form.
The distance between two geometric shapes is defined as the Euclidean distance between their low-frequency FD components.
A potential transformation in $\mathcal{T}_{\text{raw}}$ is deemed valid if the mean Euclidean distance between its FD and those of all other candidate combinations falls below a predefined threshold.
We then compute the transformation via least-squares optimization:
\begin{equation}
\begin{aligned}
\mathbf{R}, \mathbf{t}, s &= \arg\min_{\mathbf{R}, \mathbf{t}, s} \sum_{i=1}^{n_l} \| s\mathbf{R} \mathbf{c}_i + \mathbf{t} - \mathbf{p}_i \|_2^2, \\
\text{s.t.} & \quad \mathbf{R}^T \mathbf{R} = \mathbf{I}, \quad \det(\mathbf{R}) = 1,
\end{aligned}
\end{equation}
where $n_l$ is the number of correspondences.
Finally, we obtain a set of potential transformations $\mathcal{T}_{\text{raw}} = \{(\mathbf{R}_i, \mathbf{t}_i, s_i) \mid i \in \{1, \dots, m\}\}$, where $m$ is the number of potential transformations.

After $\mathcal{T}_{\text{raw}}$ is obtained, we further perform clustering to group similar transformations and retain only representative ones.
We first define the distance between two transformations as
\begin{equation}
d_T(T_i, T_j) = \lambda_1 ||\mathbf{t}_i - \mathbf{t}_j||_2 + \lambda_2 \theta(\mathbf{R}_i, \mathbf{R}_j) + \lambda_3 |s_i - s_j|,
\end{equation}
where $\theta(\mathbf{R}_i, \mathbf{R}_j)$ is the rotation angle between two rotation matrices $\mathbf{R}_i$ and $\mathbf{R}_j$, and $\lambda_1, \lambda_2, \lambda_3$ are the weights.
We then apply DBSCAN~\cite{hahsler2019dbscan} to cluster the transformations based on their distances.
For each cluster, we select the transformation with the best similarity score of FDs as the representative transformation of this cluster.
After clustering, the final set of representative transformations is $\mathcal{T} = \{(\mathbf{R}_i, \mathbf{t}_i, s_i) \mid i \in \{1, \dots, n_t\}\}$, where $n_t$ is the number of representative transformations. The process is denoted as function $\textsc{FilterOutliers}(\cdot)$.

\subsubsection{Robustness Enhancement Strategy}\label{sec:robustness_enhancement_strategy}

The generation of valid correspondences relies heavily on the accuracy of the single-frame process.
To mitigate the risk of erroneous single-frame results, we introduce a robustness enhancement strategy.

For frames $i\in\{4,5,\dots,N_f\}$ where $N_f$ is a user-defined parameter, we repeat the $\textsc{FindCand}(\cdot)$ process by randomly selecting two previous frames and combining them with the current frame, thereby generating new correspondence sets.
Unselected frames are marked as dormant.
This procedure is repeated for $N_f-3$ frames to boost the probability of finding valid transformations.
The parameter $N_f$ balances robustness and computational cost.

Additionally, if a potential solution does not find a corresponding candidate in the next frame, it will not be abandoned immediately; instead, it will be marked as dormant for one frame.
If no corresponding candidate is found in the subsequent frame, the solution is discarded.
Otherwise, the dormant solution is reactivated.

Although this dormant strategy increases computational cost and introduces some outliers, it reduces the risk of discarding valid transformations due to single-frame failures.
An ablation of the strategy is reported in Section~S3 of the supplementary material.

\subsection{SCALE-ICP REFINEMENT}\label{sec:scale_icp_refinement}

For each potential transformation $T_{\text{pot}}$, considering that its estimated scaling factor is not precise before refinement, we apply the Scale-ICP algorithm~\cite{ying2009scale} to refine the transformation and calculate the fitness score.

The Scale-ICP refinement estimates the prior-to-real transformation by minimizing
\begin{equation}
\begin{aligned}
\mathbf{R}^*, \mathbf{t}^*, s^* &=
\arg\min_{\mathbf{R},\mathbf{t},s}
\sum_{i=1}^{N_r}\|s\mathbf{R}\mathbf{c}_i+\mathbf{t}-\mathbf{p}_i\|_2^2,\\
\text{s.t.} &\quad \mathbf{R}^T\mathbf{R}=\mathbf{I},\quad \det(\mathbf{R})=1,
\end{aligned}
\end{equation}
where $\mathbf{c}_i$ are points in the prior map and $\mathbf{p}_i$ are their corresponding nearest neighbors in the observation map, and $N_r$ is the number of point correspondences. Here, the point correspondences are established over the full point clouds of the prior map and the observation map by finding, for each transformed prior-map point, its nearest neighbor in the observation map.
We first project the 3D points in the observation map onto the horizontal plane to obtain a 2D point cloud, and extract its boundary points as the projected map.
We adopt the method proposed in~\cite{ying2009scale} to compute the optimal transformation.

After refinement, we obtain a set of refined transformations $\mathcal{T}_{\text{fine}} = \{(\mathbf{R}_i, \mathbf{t}_i, s_i)\}_{i=1}^{m}$, where $\mathbf{R}_i$, $\mathbf{t}_i$, and $s_i$ are the rotation matrix, translation vector, and scaling factor of the $i$-th transformation, respectively.
The fitness score $F_i$ is defined as the mean distance between the transformed prior-map points and their nearest neighbors in the projected map:
\begin{equation}
F_i = \frac{1}{N_p}\sum_{j=1}^{N_p}\|s_i\mathbf{R}_i\mathbf{c}_j+\mathbf{t}_i-\mathbf{p}_j\|_2,
\end{equation}
where $\mathbf{p}_j$ is the nearest neighbor of the transformed prior-map point $s_i\mathbf{R}_i\mathbf{c}_j+\mathbf{t}_i$ in the projected map, and $N_p$ is the number of valid point correspondences.

Moderate local deformation of prior maps can also be mitigated during the refinement process. Through iterative point-wise matching and optimization, the algorithm emphasizes global alignment and inherently tolerates local geometric mismatches.

We only retain the transformations with fitness scores below the threshold $F_{\text{thresh}}$.
The remaining transformations are considered the final registration results, whose confidence is computed as
\begin{equation}
\text{Conf}_i = \frac{\exp(-\tau F_i)}{\sum_{j=1}^{m} \exp(-\tau F_j)},
\end{equation}
where $\tau$ is a temperature parameter controlling the sharpness of the confidence distribution. The confidence score serves as a heuristic measure of the reliability of each transformation.
The final result set $\mathcal{T}_{f}$ is composed of all fine-registered transformations and their respective confidence scores $\{T_{\text{fine}, i}, \text{Conf}_i\}$. This set provides global guidance for the subsequent exploration and path generation. The whole process is denoted as function $\textsc{Scale-ICP}(\cdot)$.

\section{PRIOR-GUIDED HIERARCHICAL VIEWPOINT PLANNING}\label{sec:path_planning}
\begin{figure*}
  \centering
  \includegraphics[width=\linewidth]{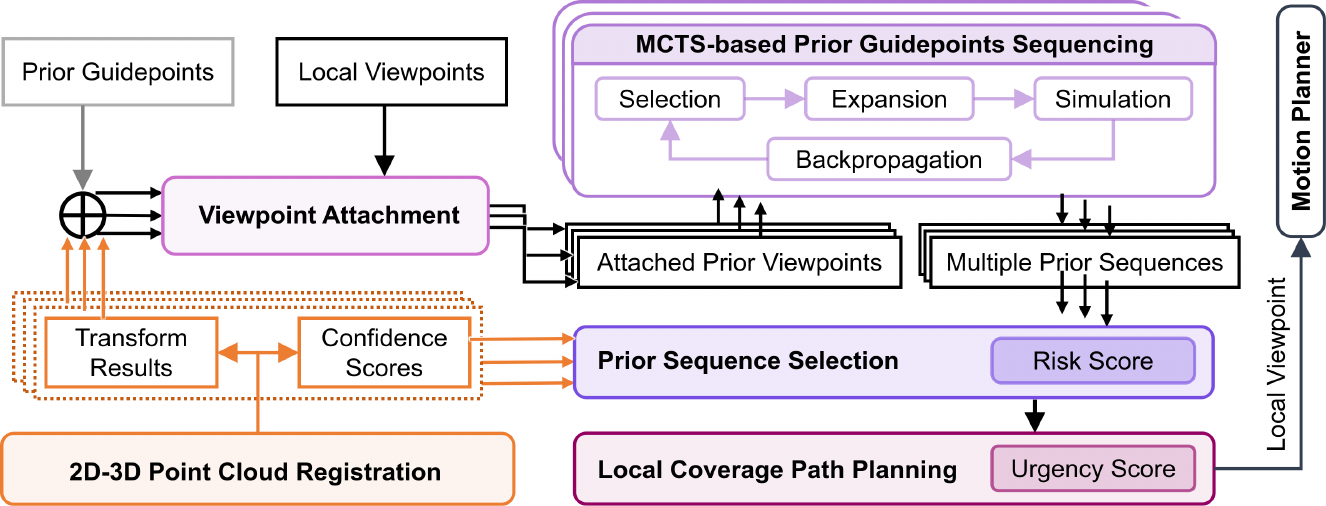}
  \caption{Pipeline of the hierarchical viewpoint planning framework.
Stacked boxes denote replication across registration hypotheses.
Each hypothesis yields one prior sequence through MCTS sequencing, and the selection module reduces these to the single sequence that drives local coverage path planning via risk-aware selection using both confidence and travel risk.}
  \label{fig:planning_pipeline}
\end{figure*}

\begin{figure*}
  \centering
  \includegraphics[width=\linewidth]{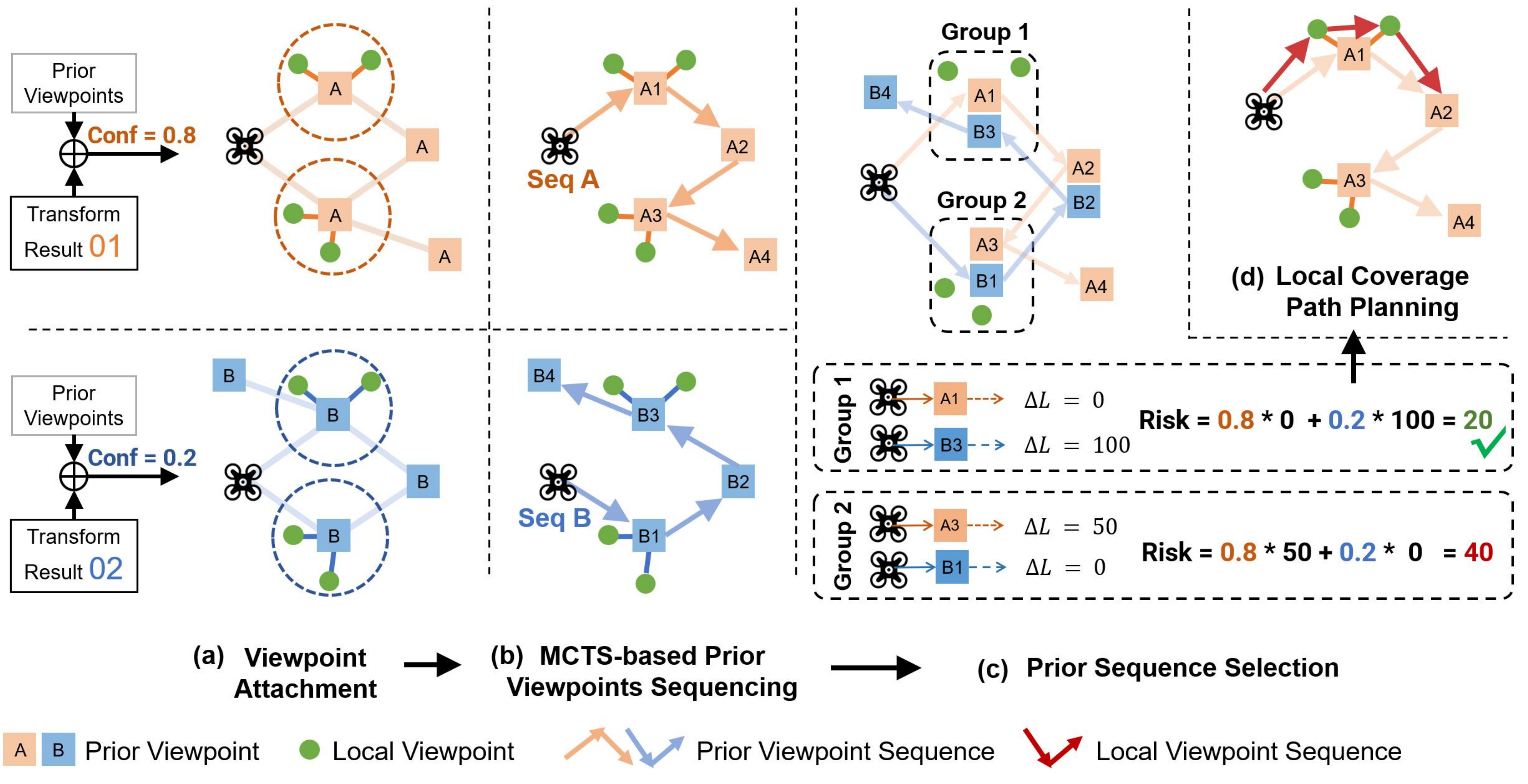}
  \caption{Demonstration case for the hierarchical viewpoint planning framework under two registration hypotheses.
(a) Each hypothesis transforms the prior guidepoints to its own positions, and local viewpoints (green) inside the attachment radius (dashed circles) are attached to the nearest guidepoint.
(b) MCTS sequencing yields one traversal order per hypothesis.
(c) The two orders disagree on which group to visit first, so each group is scored via the confidence-weighted travel risk. Group~1 attains the lower risk score.
(d) The local viewpoint sequence (red arrows) is planned under the selected guidance.}
  \label{fig:hierarchical_planning}
\end{figure*}

Given multiple possible registration results $\mathcal{T}_f$, our goal is to generate an optimal local viewpoint coverage path under the guidance of the prior guidepoints.
To achieve this, we propose a hierarchical viewpoint planning strategy that consists of four modules: viewpoint attachment, MCTS-based prior guidepoints sequencing, prior sequence selection, and local coverage path planning.
The pipeline is illustrated in Figure~\ref{fig:planning_pipeline}.
The main algorithm is shown in Algorithm~\ref{alg:path_planning}.
The demonstration scenario is shown in Figure~\ref{fig:hierarchical_planning}.

  \begin{algorithm}
  \caption{Hierarchical Viewpoint Planning}
  \begin{algorithmic}[1]
  \REQUIRE $\mathcal{T}_f$: Transformation Results, $\mathcal{G}$: Prior Guidepoints, $\mathcal{V}$: Local Viewpoints, $X_r$: Robot Node
  \ENSURE $\mathcal{S}_\text{local}$: Local viewpoint traversal sequence
  \STATE $\mathcal{S}_\text{pri} \gets \{\}$

  \FOR{each transformation $T$ in $\mathcal{T}_f$}
  \STATE // \textbf{Viewpoint Attachment:}
  \STATE $\tilde{\mathcal{G}} \gets \textsc{TransformVps}(T, \mathcal{G})$
  \STATE \textsc{AttachVps}($\mathcal{V}, \tilde{\mathcal{G}}$)
  \STATE // \textbf{MCTS-based Prior Guidepoints Sequencing:}

      \STATE $\text{seq}_\text{pri} \gets \textsc{MCTS}(X_r, \mathcal{V}, \tilde{\mathcal{G}})$
      \STATE $\mathcal{S}_{\text{pri}}.\text{insert}(\text{seq}_{\text{pri}})$
  \ENDFOR

  \STATE // \textbf{Prior Sequence Selection:}
    \STATE $\textsc{GroupVps}(\mathcal{V}, \tilde{\mathcal{G}}, \mathcal{T}_f)$
\IF{$\mathcal{T}_f.\mathtt{size}() > 1$}
    \STATE $C_\text{risk} \gets \textsc{CalcRisk}(\mathcal{T}_f, \mathcal{S}_\text{pri}, \mathcal{V})$
    \STATE $\mathcal{V}_\text{iso}, \mathcal{V}_\text{target}, \text{seq}_{\text{pri}} \gets \textsc{SelectPriSeq}(\mathcal{V}, \tilde{\mathcal{G}}, C_\text{risk})$
\ELSIF{$\mathcal{T}_f.\mathtt{size}() = 1$}
    \STATE $\mathcal{V}_\text{iso}, \mathcal{V}_\text{target}, \text{seq}_{\text{pri}} \gets \textsc{SelectPriSeq}(\mathcal{V}, \tilde{\mathcal{G}})$
\ENDIF
\STATE // \textbf{Local Coverage Path Planning:}
\STATE $\mathcal{J}_\text{urg} \gets \textsc{CalcUrgency}(\mathcal{V}_\text{iso}, \text{seq}_{\text{pri}})$
\STATE $\mathcal{J} \gets \textsc{CalcCosts}(\mathcal{V}_{\text{target}}, X_r)$
\STATE $\mathcal{S}_\text{local} \gets \textsc{FE-TSP}(\mathcal{V}_\text{iso}, \mathcal{V}_\text{target}, X_r, \mathcal{J}_\text{urg}, \mathcal{J}, \text{seq}_{\text{pri}})$

  \end{algorithmic}
  \label{alg:path_planning}
\end{algorithm}

\subsection{VIEWPOINT ATTACHMENT}\label{sec:viewpoint_management}

For each transformation result, we first transform the pre-stored prior guidepoints $\mathcal{G}$ to their real-world positions (function $\textsc{TransformVps}(\cdot)$), yielding the transformed guidepoints $\tilde{\mathcal{G}}$.
Like most frontier-based methods, we sample local viewpoints around frontiers.
We attach each local viewpoint to the nearest prior guidepoint based on Euclidean distance.
The local viewpoint $V_i$ is attached to the prior guidepoint $\tilde{G}_j$ if:
\begin{equation}
d(V_i, \tilde{G}_j) < \min \left( \rho, \min_{m \neq j} d(V_i, \tilde{G}_m) \right),
\end{equation}
where $d(\cdot)$ is the horizontal Euclidean distance, $\rho$ is the maximum attachment radius, $\tilde{G}_m$ is the $m$-th transformed prior guidepoint under one registration hypothesis.
For each transformation result, each local viewpoint is attached to at most one prior guidepoint, while a prior guidepoint may have multiple local viewpoints attached. This process is denoted as function $\textsc{AttachVps}(\cdot)$.

Next, we define three states for each prior guidepoint: unexplored, explored, and exploring.
We classify scanned points as poorly or well-observed using the criteria in EPIC~\cite{geng2025epic}.
A prior guidepoint is classified as: (1) \textit{unexplored} if it has no local viewpoints attached to it and has no points observed within its influence radius.
(2) \textit{explored} if it has no local viewpoints attached and has well-observed points within its influence radius.
(3) \textit{exploring} if it has local viewpoints attached to it.
We update the states of all transformed prior guidepoints, which then inform the criterion for prior guidepoint sequencing.

\subsection{MCTS-BASED PRIOR GUIDEPOINTS SEQUENCING}\label{sec:global_planning}

\begin{figure}
  \centering
  \includegraphics[width=0.92\linewidth]{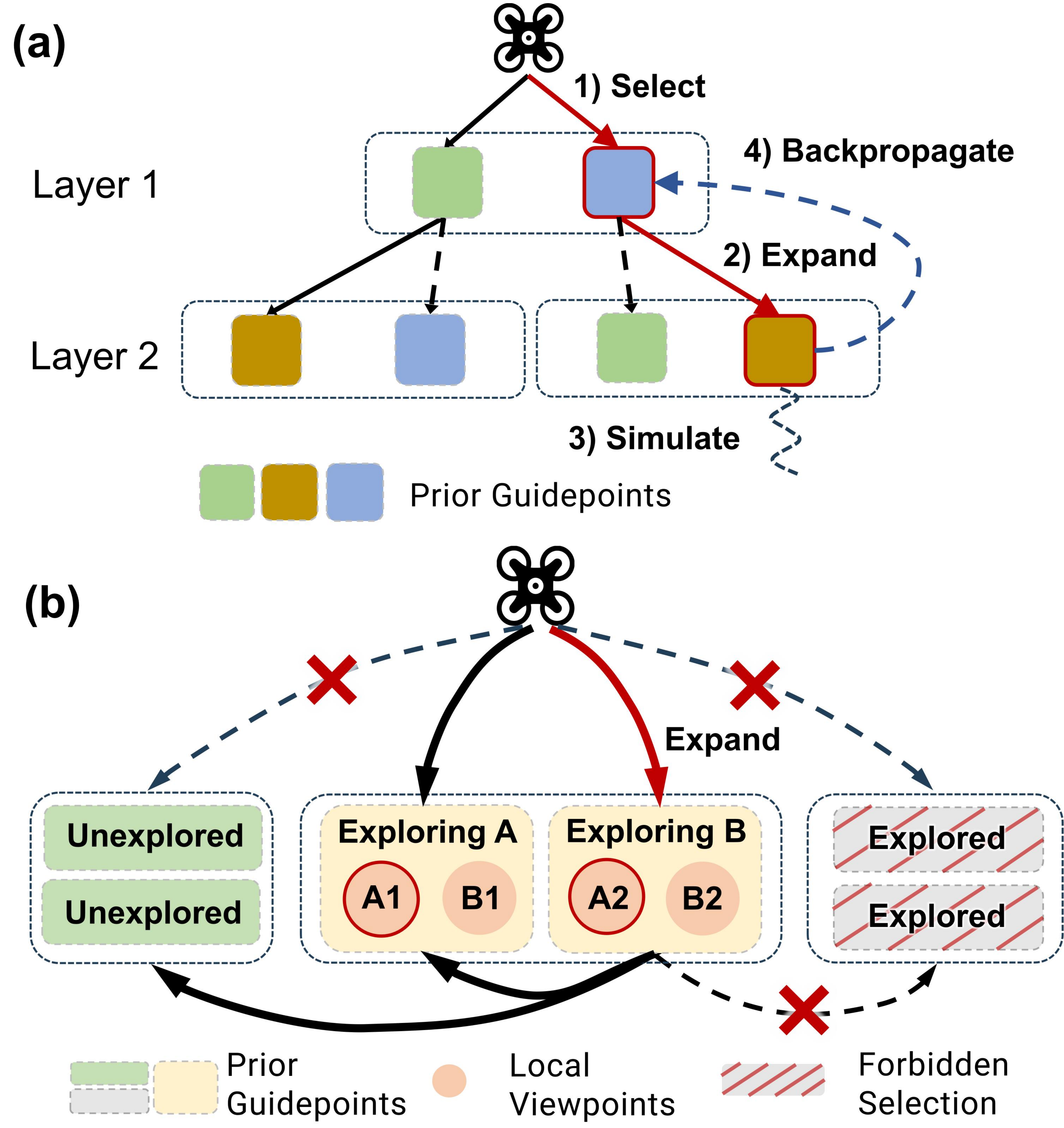}
  \caption{(a) Tree structure of the proposed MCTS method and one search iteration.
The UAV is the root, every other node is a prior guidepoint, and each branch represents a candidate traversal order.
The red arrows indicate the selected path in both selection and expansion processes.
(b) Criterion for the expansion process.
Guidepoints are partitioned by exploration state into unexplored (green), exploring (yellow, with their attached local viewpoints), and explored (hatched).
The root node can expand only to exploring guidepoints, whereas a non-root node, shown by the arrows beneath, can also reach unexplored ones.
Explored guidepoints are not allowed to be expanded to avoid redundant paths.
}
  \label{fig:mcts_tree}
\end{figure}

For each transformation result, we apply the Monte Carlo Tree Search~\cite{browne2012survey} method to search for the optimal sequence of prior guidepoints.
By performing iterative rollouts, the solver balances exploration and exploitation via Upper Confidence Bound (UCB) rules.
MCTS consists of four main steps: selection, expansion, simulation, and backpropagation.
Upon reaching the maximum iterations, the best sequence is used as the global path. The process of MCTS is shown in Figure~\ref{fig:mcts_tree}. The whole process is denoted as function $\textsc{MCTS}(\cdot)$.

\subsubsection{Tree Structure and Reward}

We first define the structure of the MCTS tree.
The root node is the robot's current position.
Each tree node corresponds to a prior guidepoint $G$ and each branch represents the traversal path of prior guidepoints.
The reward for each node is the negative total path length generated during the simulation process.
Since the standard MCTS algorithm is designed to maximize rewards, we use negative rewards here to minimize the path length.

For each transformation result, we build a distance matrix consisting of the robot's current position and all exploring and unexplored prior guidepoints.
$\tilde{G}_i^{(\text{exp})}$ is the $i$-th exploring prior guidepoint.
Assume $\tilde{G}_i^{(\text{exp})}$ has $k$ attached local viewpoints $V_{0}, V_{1}, \ldots, V_{k-1}$, and let $X_r$ denote the robot's node.
The cost from the robot's node to the exploring prior guidepoints is defined as
\begin{equation}
        J(X_r, \tilde{G}_i^{(\text{exp})}) = \frac{1}{k} \sum_{j=0}^{k-1} L(X_r, V_{j}),
\end{equation}
where $L(\cdot)$ is the searched path length between two nodes.
For the distance between two prior guidepoints $\tilde{G}_{i}$ and $\tilde{G}_{j}$, the cost is defined as
\begin{equation}
    J(\tilde{G}_{i}, \tilde{G}_{j}) = s \cdot L(\tilde{G}_{i}, \tilde{G}_{j}),
\end{equation}
where $L(\cdot)$ is the distance between the two prior guidepoints pre-calculated using the PRM, and $s$ is the scaling factor of the current transformation.
The cost from the prior guidepoints to the robot's position node is set to $0$.
The cost from the robot's current position to unexplored prior guidepoints is ignored because the proposed TSP solver avoids choosing these paths.

\subsubsection{Selection and Expansion Process}

In the selection process, we determine the UCB value for each child node to balance exploration and exploitation.
The UCB value of a child node $i$ is defined as
\begin{equation}
\text{UCB}_i = \frac{w_i}{n_i} + C_u \sqrt{\frac{\ln N}{n_i}},
\end{equation}
where $w_i$ is the total negative reward of the child node, $n_i$ is the number of times the child node has been visited, $N$ is the total number of visits to the parent node, and $C_u$ is the exploration constant. The node with the maximum UCB value is selected for expansion.

Given a selected leaf node to be expanded, the expansion process follows specific criteria, as shown in Figure~\ref{fig:mcts_tree}(b).
If the selected node is the root node, expansion is limited to exploring prior guidepoints.
For non-root nodes, expansion includes both exploring and unexplored guidepoints.
These criteria prevent direct transitions to prior guidepoints without attached local viewpoints.

\subsubsection{Simulation Process}

After expanding the leaf node, we perform a heavy simulation rollout from the newly expanded node to a terminal node.
For unreached nodes, we solve a simplified asymmetric traveling salesman problem (ATSP) to determine the optimal path for this simulation step, providing high-fidelity estimates under minimal rollouts.
The best path with its total length is continuously updated through simulation.

\subsubsection{Backpropagation Process}

After each simulation step, we perform accumulative backpropagation to all nodes along the path from the newly expanded node to the root.
For each node in the path, we update its reward and number of visits accordingly.
This updates the UCB scores of all nodes for the next iteration's selection.
\subsection{PRIOR SEQUENCE SELECTION}\label{sec:local_planning}

After obtaining the traversal sequences of prior guidepoints for all transformations, we aim to select the best sequence to serve as the global guidance for subsequent local coverage path planning.

We first group viewpoints based on their attachments under all registration hypotheses (function $\textsc{GroupVps}(\cdot)$).
Specifically, if two local viewpoints share the exact same attachments (including no attachment) across all hypotheses, they are classified into the same group.
Each group then consists of these local viewpoints together with all the prior guidepoints they are attached to.

Next, for each viewpoint group, we compute a risk score that quantifies the risk of exploring this group.
The risk score penalizes the expected extra travel length caused by registration uncertainty, defined as:
\begin{equation}
\begin{aligned}
    C_{\text{risk}}(i) &= \sum_{j=1}^{m} \text{Conf}_j \cdot \Delta L_i^j, \\
    \Delta L_i^j &= L_i^j - L^{j,*},
\end{aligned}
\end{equation}
where $\text{Conf}_j$ is the confidence score of the $j$-th transformation result and $\Delta L_i^j$ is the extra path length associated with the $i$-th viewpoint group. Specifically, $\Delta L_i^j = L_i^j - L^{j,*}$ is the extra tour length from visiting group $i$ first ($L_i^j$) rather than the best-ordering group ($L^{j,*}$), under hypothesis $j$. $L_i^j$ is obtained from the previous MCTS process where intermediate results are calculated and stored via the simulation process. This process is denoted as function $\textsc{CalcRisk}(\cdot)$.

To determine the active global guidance, a greedy strategy is applied.
This function first evaluates all viewpoint groups to identify the one with the minimum risk score, and then selects its corresponding prior guidepoint sequence.
If multiple prior sequences point to the same selected group, we choose the sequence with the highest confidence score.

If the registration module generates only one transformation result, we assume it is correct and bypass the risk calculation process.
The viewpoint group attached to the next prior guidepoint is selected directly.
The selection process is denoted as function $\textsc{SelectPriSeq}(\cdot)$.

\subsection{LOCAL COVERAGE PATH PLANNING}

Directly solving an ATSP over all local viewpoints may generate paths inconsistent with the global guidance.
We therefore formulate an FE-TSP to constrain the directional destination of local traversal.

First, we select local viewpoints from both the isolated groups and the target group to be visited next.
The isolated group is the group with no prior guidepoints, meaning that the local viewpoints in this group have no attachment under any registration hypothesis.
The target group is the next viewpoint group selected to visit.
All local viewpoints within other groups are excluded from planning.
\begin{figure}
  \centering
  \includegraphics[width=0.98 \linewidth]{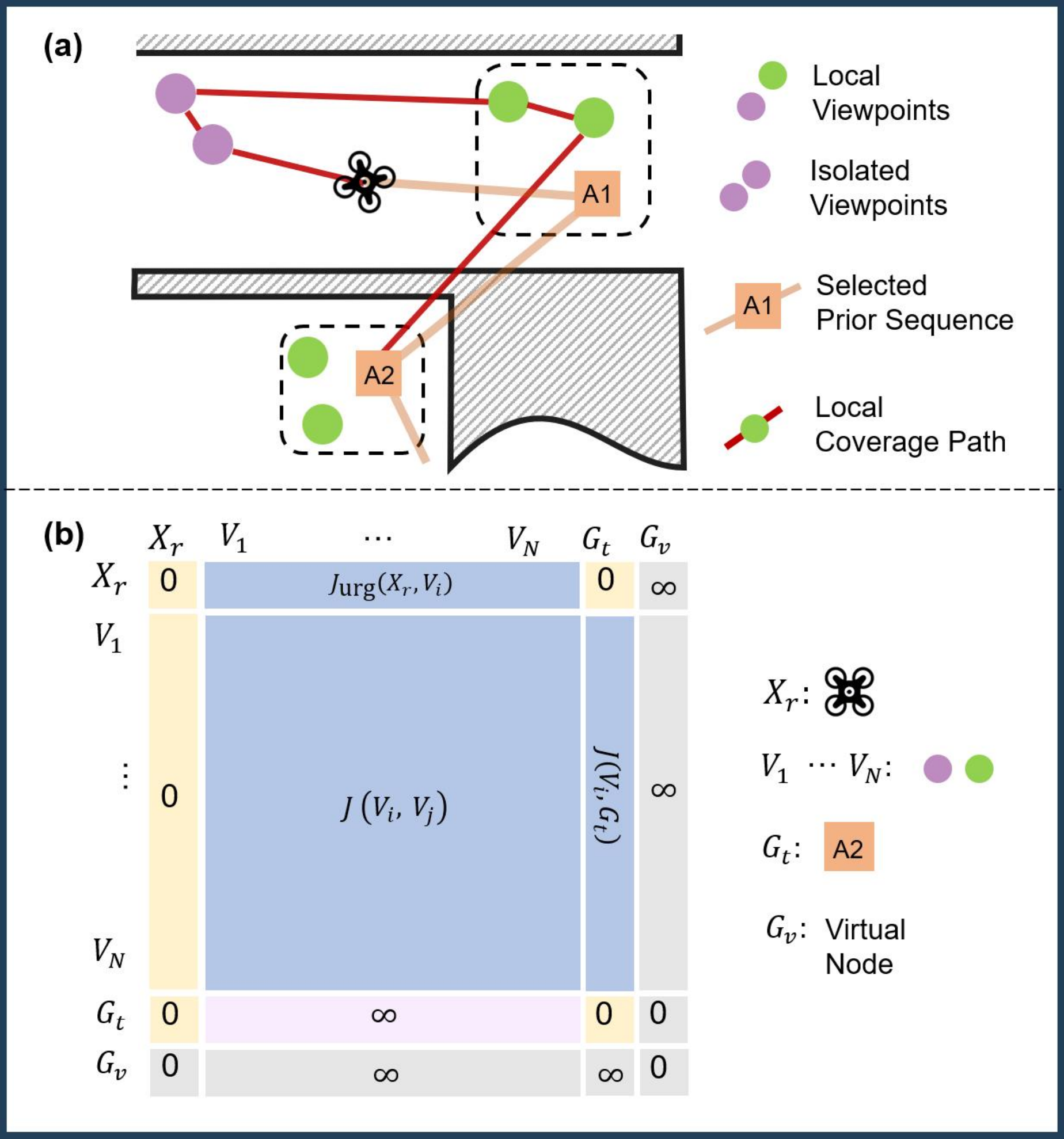}
  \caption{
    (a) Visualization of the local coverage path planning.
    (b) Distance Matrix Construction for the FE-TSP problem.}
  \label{fig:local_planning}
\end{figure}

We construct three types of viewpoint nodes for the FE-TSP problem: (1) $\mathcal{V}_{\text{iso}}$: all viewpoints in the isolated groups.
(2) $\mathcal{V}_{\text{target}}$: all viewpoints in the target group.
(3) $G_t$: the second next prior guidepoint in the prior sequence, which serves as the termination node of the FE-TSP problem.
$\mathcal{V}_{\text{iso}}$ and $\mathcal{V}_{\text{target}}$ form all local viewpoints $V_1, \cdots, V_N$ to be visited in the FE-TSP problem.
The three types of viewpoints are illustrated in Figure~\ref{fig:local_planning}(a).
Thus, the FE-TSP problem can be constructed as:

\textit{Given the robot's current position node $X_r$, a set of local viewpoints $V_1, \cdots, V_N$, and termination node $G_t$, find the optimal path from $X_r$ to $G_t$ that visits all $V_1, \cdots, V_N$ with the minimum total cost.}

To solve this problem, we apply the Dummy Node Method by introducing a virtual node $G_v$ connected to both start and terminal nodes, translating it into a general ATSP framework.
The costs $J_{G_t, G_v}$ and $J_{G_v, X_r}$ are set to zero; costs $J_{X_r, G_v}$, $J_{G_v, G_t}$, and the costs between the virtual node and all local viewpoints ($J_{G_v, V_{1\cdots N}}$ and $J_{V_{1\cdots N}, G_v}$) are set to infinity to ensure the validity of the solution.
The problem is then transformed into a general ATSP problem.

We then construct the distance matrix for the ATSP problem.
To encourage visiting isolated viewpoints early, we calculate and apply an urgency cost to them to prioritize their visitation ($\textsc{CalcUrgency}(\cdot)$).
The cost between the start node $X_r$ and viewpoint $V_i$ is defined as
\begin{equation}
  \begin{aligned}
      J_{\text{urg}}(X_r, V_i) &= \max \left( \frac{L(X_r, V_i)}{v_{\text{max}}}, \frac{\mathrm{dyaw}(X_r, V_i)}{\omega_{\text{max}}} \right) + U(V_i), \\
      U(V_i) &= \begin{cases}
      \beta_{e} \times \frac{\mathrm{dyaw}(X_r, V_i)}{\omega_{\text{max}}}, & V_i \text{ is isolated} \\
      0, & \text{otherwise}
      \end{cases}
  \end{aligned}
\end{equation}
where $L(\cdot)$ is the searched path length between two nodes, $\mathrm{dyaw}(\cdot)$ is the minimum yaw angle difference between the two viewpoints, $v_{\text{max}}$ and $\omega_{\text{max}}$ are the maximum velocity and yaw rate, respectively, and $\beta_{e}$ is the urgency cost weight.
The cost between two viewpoints $V_i$ and $V_j$ is defined as
\begin{equation}
J(V_i, V_j) = \frac{L(V_i, V_j)}{v_{\text{max}}}.
\end{equation}
We also calculate basic spatial distance costs for other viewpoints ($\textsc{CalcCosts}(\cdot)$).
Let $\tilde{\mathbf{p}}_t$ be the 2D coordinate of the terminal prior guidepoint transformed into the real-world frame.
The cost from viewpoint $V_j$ to the terminal node $G_t$ is defined as
\begin{equation}
J(V_j, G_t) = \frac{\| \mathbf{p}_j - \tilde{\mathbf{p}}_t \|_2}{v_{\text{max}}},
\end{equation}
where $\mathbf{p}_j$ is the 2D coordinate of $V_j$.
Since the terminal node is transformed from a 2D prior guidepoint and does not correspond to an actual sampled local viewpoint, this terminal cost is approximated by the Euclidean distance in the real-world horizontal plane.
However, since this cost provides only directional guidance, certain numerical estimation errors do not significantly compromise optimality.

Finally, the cost from local viewpoints to the start node $J(V_i, X_r)$ is set to zero, as there is no need for closure.
The cost from the terminal node to local viewpoints $J(G_t, V_j)$ is set to infinity to ensure the validity of the solution.
The visualization of the cost matrix is shown in Figure~\ref{fig:local_planning}(b).
After constructing the cost matrix, we solve the ATSP problem using the LKH solver~\cite{helsgaun2009general} to generate an efficient local path that is consistent with the global guidance path, which is denoted as the $\textsc{FE-TSP}(\cdot)$ function.

\subsection{LOCAL TRAJECTORY GENERATION}\label{sec:local_traj_planning}

After identifying the next viewpoint, we aim to generate smooth, feasible, and energy-efficient local trajectories.
The $A^*$ algorithm~\cite{hart1968formal} serves as the front-end to find a coarse path.
In the back-end, we employ the piecewise polynomial method MINCO~\cite{wang2022geometrically} to optimize both position trajectories $p(t) = [x(t), y(t), z(t)]^{T}$ and yaw trajectory $\psi(t)$.
To ensure safety, we construct a polyhedral flying corridor using FIRI~\cite{wang2025fast}.
The yaw direction is always constrained to point towards the unexplored region to maximize sensor coverage.
To ensure kinodynamically feasible trajectories and stable localization and mapping, we constrain the maximum velocity, acceleration, yaw rate, and yaw acceleration during the optimization process.
Through joint spatio-temporal optimization with MINCO, we generate local trajectories that are safe, kinodynamically feasible, and energy-efficient for robust field operations.

\section{SIMULATION AND BENCHMARK}\label{sec:simulation}

\begin{figure*}
  \centering
  \includegraphics[width=0.95\linewidth]{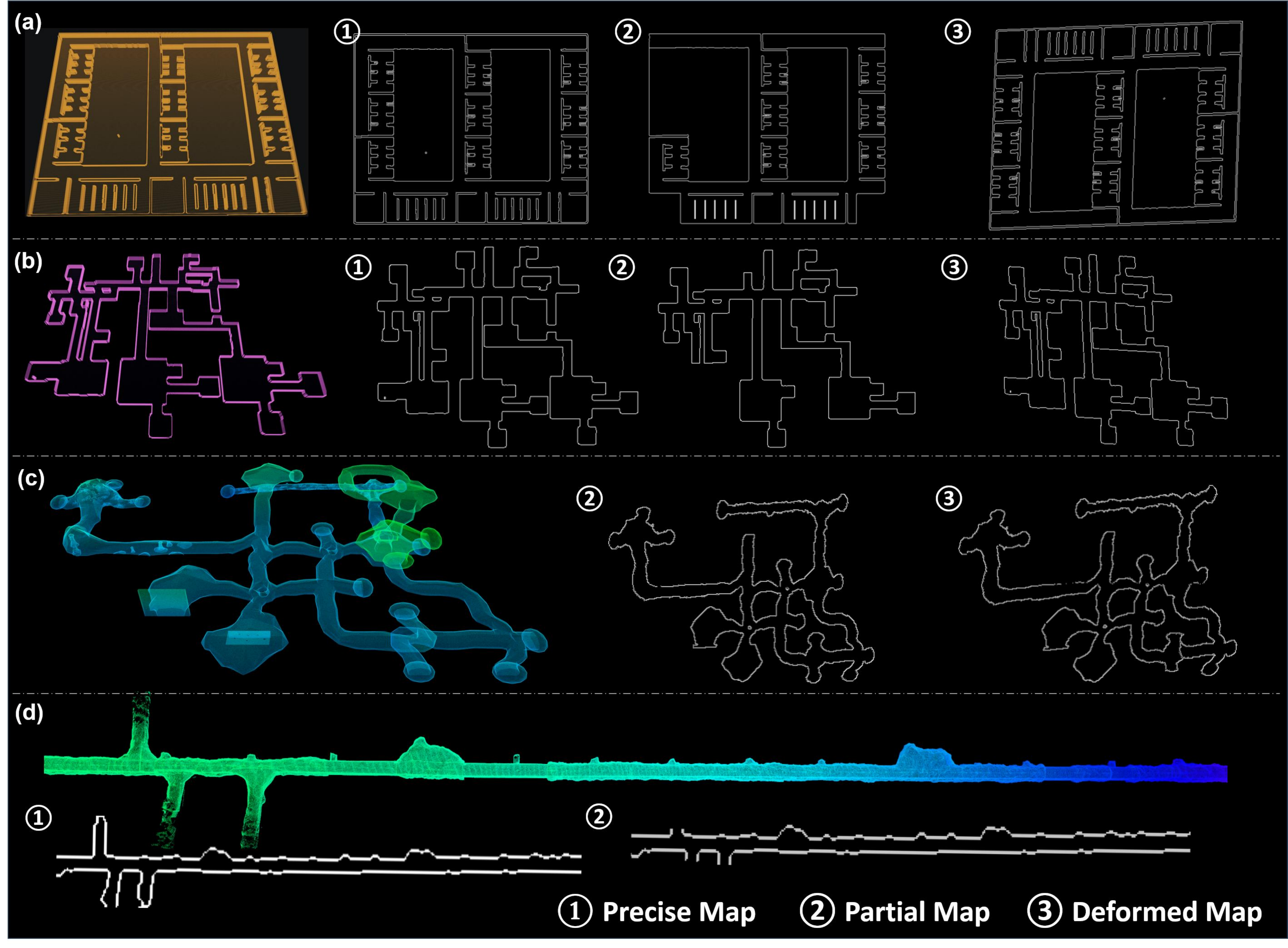}
  \caption{Simulation environments and prior maps (Precise, Partial, and Deformed).
(a) Garage environment.
(b) Indoor architecture environment.
(c) Cave environment from the DARPA SubT challenge.
(d) Simple cave environment collected from a real-world cave.}
  \label{fig:Env_and_prior_map}
\end{figure*}
\begin{table}[htbp]
    \centering
    \vspace{1ex}
    \begin{tabular}{cccc}
    \toprule
    \textbf{Parameter} & \textbf{Value} & \textbf{Parameter} & \textbf{Value} \\
    \midrule
    \multicolumn{4}{c}{\textit{Registration Parameters}} \\
    \midrule
    $N_b$            & 90               & $\alpha_2$       & 0.4              \\
    $N_c$            & 10               & $\alpha_3$       & 0.2              \\
    $d_{\text{gap}}$ & \qty{1}{\metre}  & $\lambda_1$      & 5                \\
    $\tau$       & 3.33             & $\lambda_2$      & 10               \\
    $\alpha_1$           & 0.4              & $\lambda_3$      & 1               \\
    $N_f$            & 4                    & $\eta$           & 5\%          \\
    \midrule
    \multicolumn{4}{c}{\textit{Planning Parameters}} \\
    \midrule
    $C_u$            & 1.414           &    $v_{\text{max}}$ & \qty{4.0}{\metre\per\second}  \\
    $\omega_{\text{max}}$ & \qty{4.0}{\radian\per\second}            & $\beta_e$             & $-0.2$           \\
    $\rho$                & \qty{20}{\metre}   & & \\
    \bottomrule
    \end{tabular}
    \caption{Parameters used in the experiments.}
    \label{tab:parameters}
\end{table}

We first conduct several modular evaluations, then present large-scale comparative studies to benchmark both the registration module and the overall exploration pipeline.
Tests span four large-scale environments: a garage $[\qtyproduct{192 x 150 x 4}{\metre}]$, a simple cave $[\qtyproduct{500 x 50 x 70}{\metre}]$, a SubT cave $[\qtyproduct{2024 x 330 x 41}{\metre}]$, and an indoor architecture $[\qtyproduct{150 x 110 x 4}{\metre}]$.
The garage features high topological connectivity.
The simple cave scene is collected from a real-world cave environment that has a simple topology with a few branches.
The SubT cave scene is based on the DARPA SubT Cave World Dataset published in~\cite{koval2020subterranean}.
The indoor architecture environment is irregularly shaped with low connectivity.
We design three input sketches for each environment—a precise, a partial, and a deformed prior map—as shown in Figure~\ref{fig:Env_and_prior_map}.
All tests are executed on a computer with an AMD Ryzen 7 8845H CPU and 24 GB of RAM.
Algorithmic parameters are outlined in Table~\ref{tab:parameters}, and the role of each parameter and the basis for its value are discussed in Section~S2 of the supplementary material.

\subsection{MODULE EVALUATION}

\subsubsection{Single-frame Candidate Retrieval Test}
\begin{figure}
  \centering
  \includegraphics[width=\linewidth]{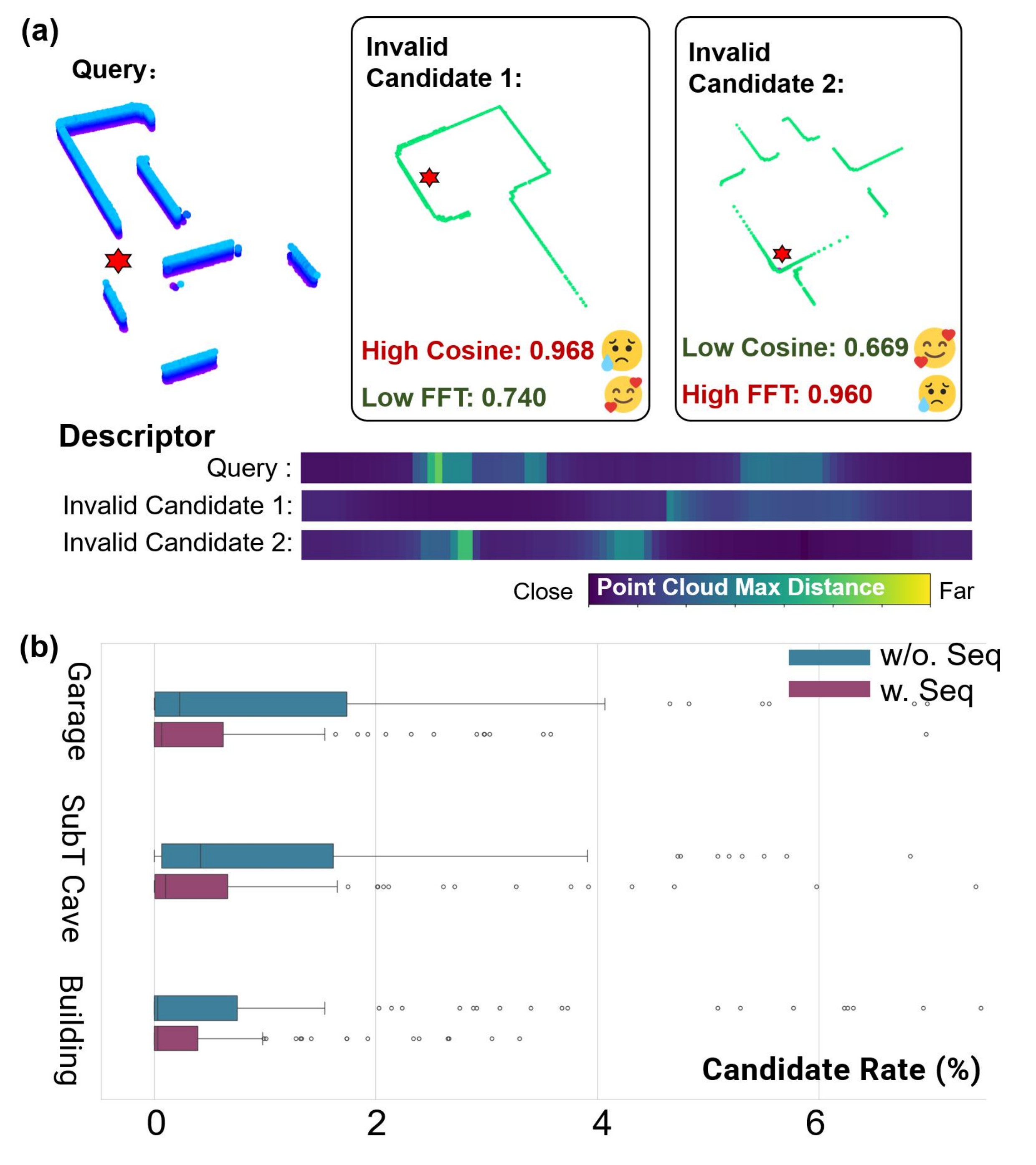}
  \caption{(a) Demonstration case that shows the performance of the cosine similarity and FFT similarity.
The red hexagon represents the UAV's current position and observation viewpoint.
Both Target 1 and Target 2 are outliers.
(b) Candidate ranking score with and without the proposed sequential similarity metric.}
  \label{fig:single_frame_reg}
\end{figure}

\begin{table}[htbp]
  \centering
  \resizebox{\columnwidth}{!}{%
  \begin{tabular}{ccccccc}
  \toprule
  Env. &
    Scene &
    \begin{tabular}[c]{@{}c@{}}Succ.\\ Rate\end{tabular} &
    \begin{tabular}[c]{@{}c@{}}Cos. \\ Sim.\end{tabular} &
    \begin{tabular}[c]{@{}c@{}}FFT\\ Sim.\end{tabular} &
    \begin{tabular}[c]{@{}c@{}}CV\\ Sim.\end{tabular} &
    \begin{tabular}[c]{@{}c@{}}Tot.\\ Sim.\end{tabular} \\ \midrule
  \multirow{3}{*}{Garage}   & Precise & 92.7\% & 0.95 & 0.98 & 0.85 & 0.941 \\
                            & Scaled   & 92.8\% & 0.93 & 0.97 & 0.84 & 0.927 \\
                            & Deformed & 87.5\% & 0.87 & 0.94 & 0.80 & 0.882 \\ \midrule
  \multirow{3}{*}{SubT Cave}     & Precise & 93.0\% & 0.94 & 0.96 & 0.83 & 0.925 \\
                            & Scaled   & 92.7\% & 0.94 & 0.97 & 0.84 & 0.931 \\
                            & Deformed & 88.2\% & 0.85 & 0.95 & 0.78 & 0.874 \\ \midrule
  \multirow{3}{*}{Indoor} & Precise & 91.6\% & 0.94 & 0.98 & 0.84 & 0.935 \\
                            & Scaled   & 92.8\% & 0.93 & 0.97 & 0.85 & 0.929 \\
                            & Deformed & 88.4\% & 0.84 & 0.93 & 0.84 & 0.875 \\ \bottomrule
  \end{tabular}%
  }
  \caption{Success rate of single-frame candidate retrieval and similarity score of each metric in different scenarios.}
  \label{tab:single_frame_success_rate}
\end{table}

We perform the single-frame test at a frequency of \qty{1}{\hertz} during the robot's motion.
Note that this frequency is only used in this specific test context.
If the distance of all selected candidates to the ground truth is larger than the distance threshold $d_{\text{cand}}$, the registration is considered failed.
In practice, we set the distance threshold to $d_{\text{cand}} = 4 \times d_{\text{gap}}$.
We test the performance under precise and deformed prior maps.
Meanwhile, we scale the precise map (by a factor of 2 in practice) to create a scaled prior map.
We record all similarity metrics, total similarity, and success rate under these scenarios in Table~\ref{tab:single_frame_success_rate}.

Table~\ref{tab:single_frame_success_rate} shows that our method achieves a high success rate (above 91\%) with precise and scaled prior maps.
The success rate decreases under deformation but remains above 87\% in all tested environments.

Furthermore, we conduct an ablation study to evaluate the proposed combined similarity metric against using only cosine similarity. Performance is measured by a ranking score, defined as the rank position of the highest-ranked correct candidate (within $d_{\text{cand}}$) among all candidates, where a higher score indicates that correct candidates are ranked closer to the top, reflecting better local environment distinction.
Figure~\ref{fig:single_frame_reg}(a) shows a case where Target 1 and Target 2 are both outliers.
Individual metrics exhibit distinct blind spots: cosine similarity successfully rejects outlier Target 2 but fails on Target 1, whereas FFT similarity exhibits the opposite behavior. Fusing these metrics effectively eliminates most outliers. Quantitatively, Figure~\ref{fig:single_frame_reg}(b) demonstrates that the combined approach yields a better average ranking, lower variance, and a significantly raised lower bound, indicating enhanced discrimination of the local environment.

\begin{figure}
  \centering
  \includegraphics[width=\linewidth]{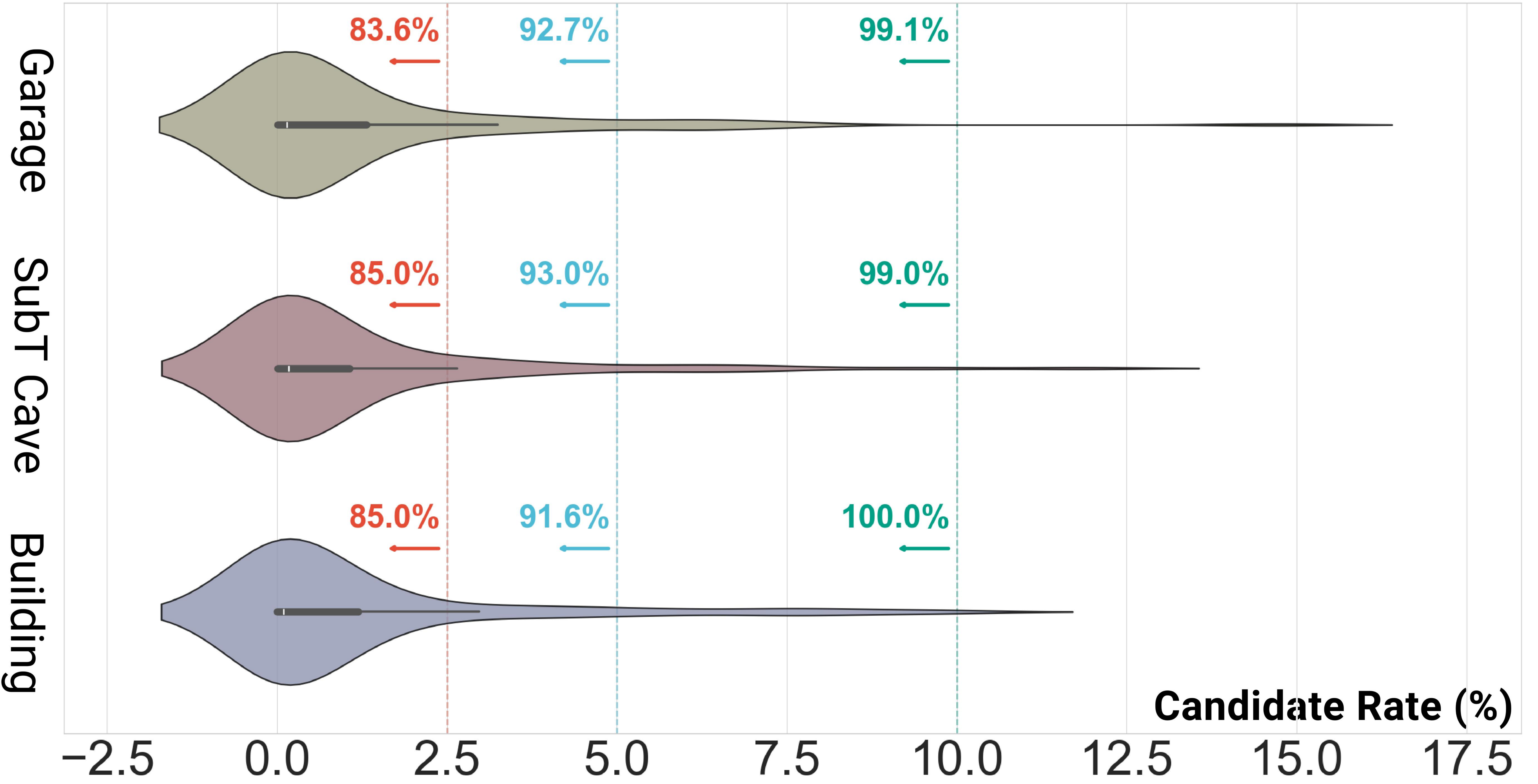}
  \caption{Distribution of the candidate ranking score and success rate under different candidate rates.}
  \label{fig:single_frame_candidate_rate}
\end{figure}

Figure~\ref{fig:single_frame_candidate_rate} further quantifies the distribution of the candidate ranking score and success rate under different candidate rates $\eta$.
The results indicate that true matches consistently rank highly, demonstrating that a relatively low candidate rate is sufficient to maintain a robust registration success rate without incurring excessive computational overhead.

\subsubsection{Multi-frame Verification and Scale-ICP Test}
\begin{figure*}
  \centering
  \includegraphics[width=0.90\linewidth]{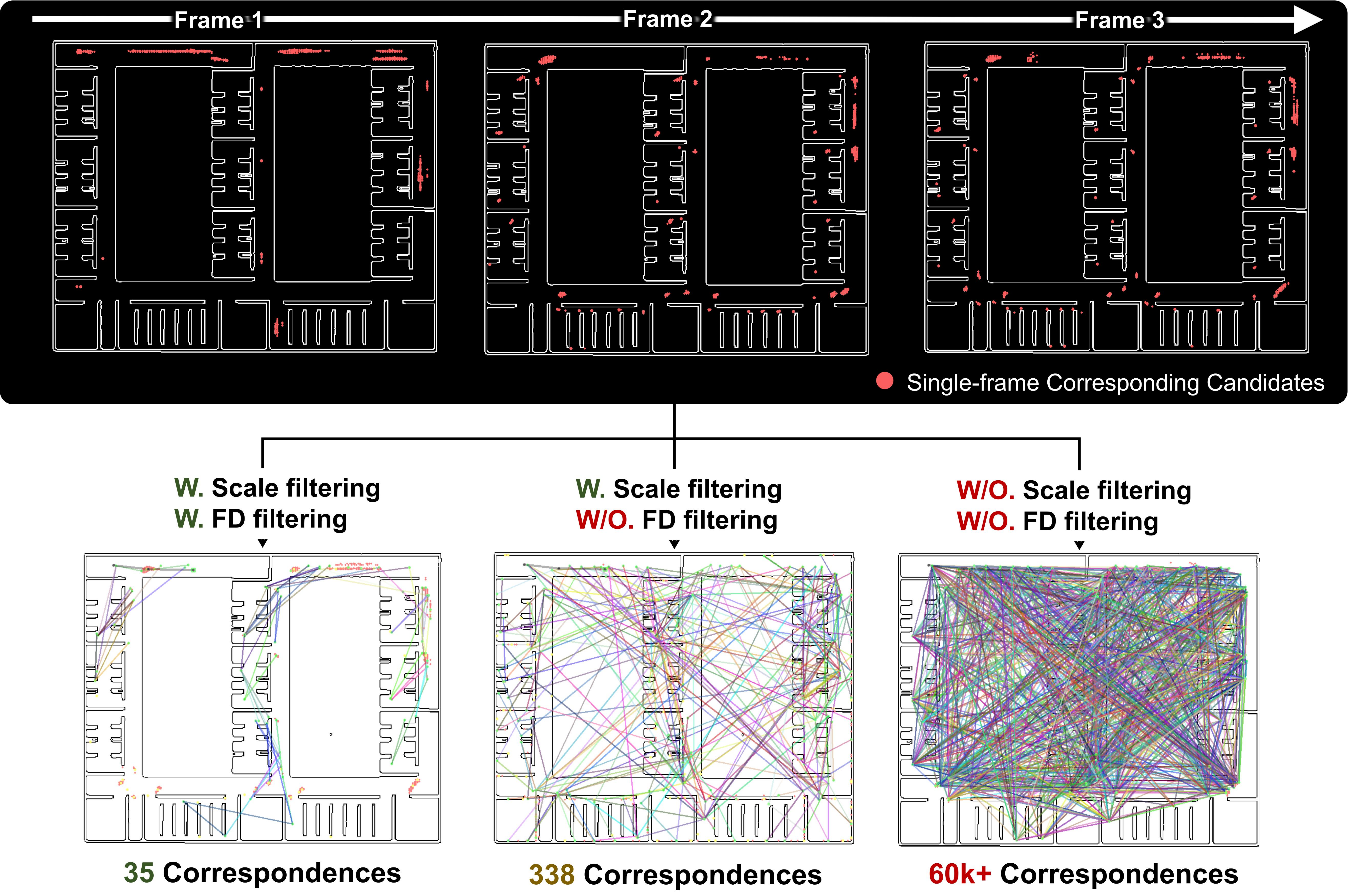}
  \caption{Illustration of the outlier filtering performance in the multi-frame verification process.
  The top row shows the outcomes of single-frame candidate retrieval in the first three frames, while the bottom row presents the results of multi-frame verification.}
  \label{fig:multi_frame_registration}
\end{figure*}

Multi-frame verification is designed to generate coarse registration results for the Scale-ICP refinement.
During the coarse phase, robust outlier rejection is paramount to provide high-quality initial results for refinement.
Our method employs a dual-filtering strategy combining scale and FD filtering.
As illustrated in Figure~\ref{fig:multi_frame_registration}, this verification mechanism effectively prunes over 60,000 initial correspondences down to 35, significantly alleviating the computational burden of the subsequent fine registration.
\begin{figure}[!tb]
  \centering
  \includegraphics[width=\linewidth]{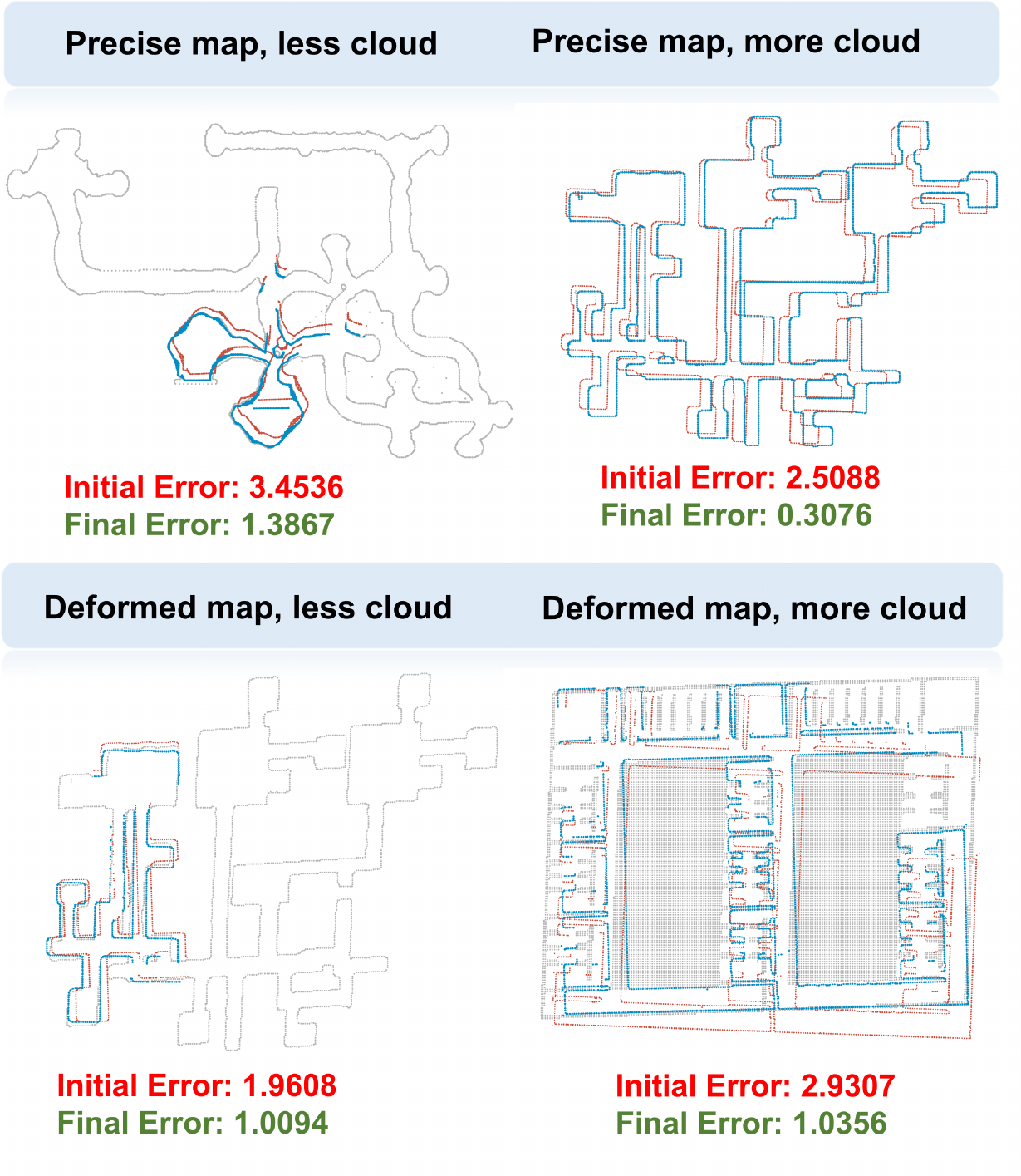}
  \caption{Scale-ICP performance under various scenarios.
The \textcolor{Gray}{gray points} are boundary points from the prior map; the \textcolor{red!70!black}{red points} are the local map under the coarse transformation; the \textcolor{blue!60!white}{blue points} are the local map after Scale-ICP refinement.}
  \label{fig:icp_test}
\end{figure}

Furthermore, we evaluate the proposed Scale-ICP refinement module, which aims to improve coarse registration accuracy and reject incorrect hypotheses.
Figure~\ref{fig:icp_test} demonstrates that Scale-ICP reliably converges to correct results despite considerable initial transformation deviations.
As point cloud data accumulates, the refinement process exhibits increased precision and higher tolerance for erroneous initial estimates.
\begin{table}[]
  \centering
    \resizebox{\columnwidth}{!}{
  \begin{tabular}{cccc}
    \toprule
  \textbf{20 nodes}                       & Optimality & Time(Sequential)/ms & Time(Parallel)/ms \\
  5 iters                        & 100\%      & 30.65           & 10.58             \\
  10 iters                       & 100\%      & 58.14           & 16.75             \\
  20 iters                       & 100\%      & 125.53          & 42.97             \\ \midrule
  \textbf{50 nodes}                       & Optimality & Time(Sequential)/ms    & Time(Parallel)/ms    \\
  5 iters                        & 100\%      & 80.72           & 23.43             \\
  10 iters                       & 100\%      & 159.92          & 45.32             \\
  20 iters                       & 100\%      & 321.02          & 95.62             \\ \midrule
  \textbf{100 nodes} & Optimality & Time(Sequential)/ms    & Time(Parallel)/ms    \\
  5 iters                        & 96.46\%    & 215.14          & 54.36             \\
  10 iters                       & 99.18\%    & 429.07          & 123.24            \\
  20 iters                       & 99.89\%    & 860.54          & 245.50            \\ \bottomrule
  \end{tabular}}
  \vspace{1ex}
  \caption{Optimality and time consumption regarding different node numbers and iteration times.}
  \label{tab:mcts_test}
  \end{table}

\subsubsection{MCTS Solver Test}
We evaluate the MCTS solver's performance for global path planning by analyzing the computational time and path optimality.
In our simulation, distance matrix entries are populated randomly, with 10\% of nodes designated as attached nodes.
We define optimality as the ratio of the pseudo-optimal path length (computed via 500 iterations) to the path length obtained under different iteration budgets.
Furthermore, we track both single-threaded and multi-threaded (4 threads) execution times.

As Table~\ref{tab:mcts_test} demonstrates, the MCTS solver matches the 500-iteration reference solution for the 20-node and 50-node cases under very limited iteration budgets.
When the node number increases to 100, the solver can still achieve 96.46\% optimality under 5 iterations and close to 100\% optimality under 20 iterations.
Furthermore, leveraging its parallel nature, multi-threading significantly reduces the computational time.
In scenarios tested in this paper, around 20-30 prior guidepoints are sufficient for global topological representation.
Therefore, the proposed MCTS solver delivers near-optimal real-time performance and exhibits strong scalability.

\subsection{REGISTRATION BENCHMARK}

In this section, we compare the performance of the proposed registration method with several state-of-the-art methods including \textbf{BIM-reg}~\cite{qiao2025speak} and the 2D feature-based method \textbf{SuperGlue}~\cite{sarlin2020superglue}.

\begin{itemize}
    \item \textbf{BIM-reg:} A 2D-3D registration framework for indoor architecture using Building Information Modeling (BIM) data. It extracts triangle descriptors from line and corner features for constant-time matching, generates transformation candidates via Hough transform and hierarchical voting, and verifies them using an occupancy-aware criterion.
    \item \textbf{SuperGlue:} A 2D feature-based matching method. We project the 3D vertical LiDAR scans onto a 2D plane to extract SuperPoint~\cite{detone2018superpoint} features using pre-trained outdoor weights. The optimal matching is found by solving a transportation problem, and the transformation is estimated using RANSAC~\cite{chum2003locally} to filter outliers.
\end{itemize}

\begin{table}[]
  \centering
  \resizebox{\columnwidth}{!}{%
  \begin{tabular}{cccccccc}
  \toprule
   &
     &
    \multicolumn{3}{c}{Precise} &
    \multicolumn{3}{c}{Deformed} \\
   &
     &
    TTC &
    Err. &
    Succ. &
    TTC &
    Err. &
    Succ. \\
   &
     &
    $\downarrow$    (s) &
    $\downarrow$    (m) &
    $\uparrow$ &
    $\downarrow$    (s) &
    $\downarrow$    (m) &
    $\uparrow$ \\ \midrule
   &
   BIM-reg &
    188.42 &
    0.55 &
    4 &
    - &
    - &
    0 \\
   &
    SuperGlue &
    \cellcolor[HTML]{C8E6C9} 83.67 &
    \cellcolor[HTML]{C8E6C9} 0.38 &
    5 &
    \cellcolor[HTML]{C8E6C9} 90.25 &
    \cellcolor[HTML]{C8E6C9} 2.19 &
    5 \\
  \multirow{-3}{*}{Garage} &
  \textbf{Proposed} &
    \cellcolor[HTML]{E1BEE7}\textbf{38.20} &
    \cellcolor[HTML]{E1BEE7}\textbf{0.26} &
    5 &
    \cellcolor[HTML]{E1BEE7}\textbf{45.21} &
    \cellcolor[HTML]{E1BEE7}\textbf{1.25} &
    5 \\ \midrule
   &
   BIM-reg &
    - &
    - &
    - &
    - &
    - &
    0 \\
   &
    SuperGlue &
    \cellcolor[HTML]{C8E6C9} 173.67 &
    \cellcolor[HTML]{C8E6C9} 0.78 &
    5 &
    \cellcolor[HTML]{C8E6C9} 201.82 &
    \cellcolor[HTML]{C8E6C9} 4.02 &
  5 \\
  \multirow{-3}{*}{SubT Cave} &
  \textbf{Proposed} &
    \cellcolor[HTML]{E1BEE7}\textbf{66.56} &
    \cellcolor[HTML]{E1BEE7}\textbf{0.36} &
    5 &
    \cellcolor[HTML]{E1BEE7}\textbf{89.25} &
    \cellcolor[HTML]{E1BEE7}\textbf{2.92} &
    5 \\ \midrule
   &
   BIM-reg &
    286.20 &
    0.41 &
    3 &
    - &
    - &
    0 \\
   &
    SuperGlue &
    \cellcolor[HTML]{C8E6C9} 66.50 &
    \cellcolor[HTML]{C8E6C9} 0.36 &
    5 &
    \cellcolor[HTML]{C8E6C9} 74.50 &
    \cellcolor[HTML]{C8E6C9} 2.74 &
    5 \\
  \multirow{-3}{*}{Indoor} &
    \textbf{Proposed} &
    \cellcolor[HTML]{E1BEE7}\textbf{33.20} &
    \cellcolor[HTML]{E1BEE7}\textbf{0.22} &
    5 &
    \cellcolor[HTML]{E1BEE7}\textbf{38.45} &
    \cellcolor[HTML]{E1BEE7}\textbf{2.07} &
    5 \\ \bottomrule
  \end{tabular}%
  }
  \caption{Benchmark results of different registration methods in various environments and scenarios.
  TTC: Time To Convergence (s); Err.: Registration Error (m); Succ.: Success count out of 5 tests.
  '-' indicates the method fails to work in this scenario.
  The best and second-best results are highlighted in \textcolor[HTML]{E1BEE7}{\rule{1em}{1em}} and \textcolor[HTML]{C8E6C9}{\rule{1em}{1em}}, respectively.}
  \label{tab:registration_benchmark}
  \end{table}

\begin{table}[]
  \centering
  \resizebox{\columnwidth}{!}{%
  \begin{tabular}{cccc}
  \toprule
            & Scale Difference & Deformation & Multiple Solutions \\ \midrule
  BIM-reg   & \ding{55}        & \ding{55}   & \checkmark         \\
  SuperGlue & \checkmark       & \checkmark  & \ding{55}          \\
  \textbf{Proposed}  & \checkmark       & \checkmark  & \checkmark         \\ \bottomrule
  \end{tabular}%
  }
  \caption{Comparison of different methods in terms of handling scale difference, deformation, and supporting multiple solutions.}
  \label{tab:capability_comparison}
  \end{table}

For each method, we conduct five tests in each environment under both precise maps and deformed maps.
We adopt the same trajectory setup for all tests.
Since our method may compute multiple transformation results, we select the one with the highest confidence score.
As the robot explores the environment, we record three metrics: 1) \textit{Time to Convergence} (TTC), i.e., the time until the first successful registration; 2) \textit{Registration Error}, calculated as the root-mean-square error (RMSE) of closest point-to-point correspondences, consistent with the standard ICP formulation; and 3) \textit{Success Count}.
The registration process is considered successful if the registration error is below a certain threshold.
A trial is marked as a failure if no successful registration occurs after exploring 80\% of the environment.

As shown in Table~\ref{tab:registration_benchmark}, our method outperforms all baselines across all metrics.
BIM-reg relies heavily on precise line and corner features inherent to BIM models, making it unsuitable for unstructured outdoor environments like caves.
Additionally, it depends on accurate prior input to construct triangle descriptors, which makes it sensitive to discrepancies.
This method also requires a large number of line and corner features, leading to poor performance in simpler environments.
SuperGlue performs stably, but it struggles in local-to-global registration tasks as features are extracted and assigned independently while lacking spatial consistency constraints.
Consequently, extracting global features for successful registration requires significantly more time.
\begin{figure*}[ht]
  \centering
  \includegraphics[width=0.98\textwidth]{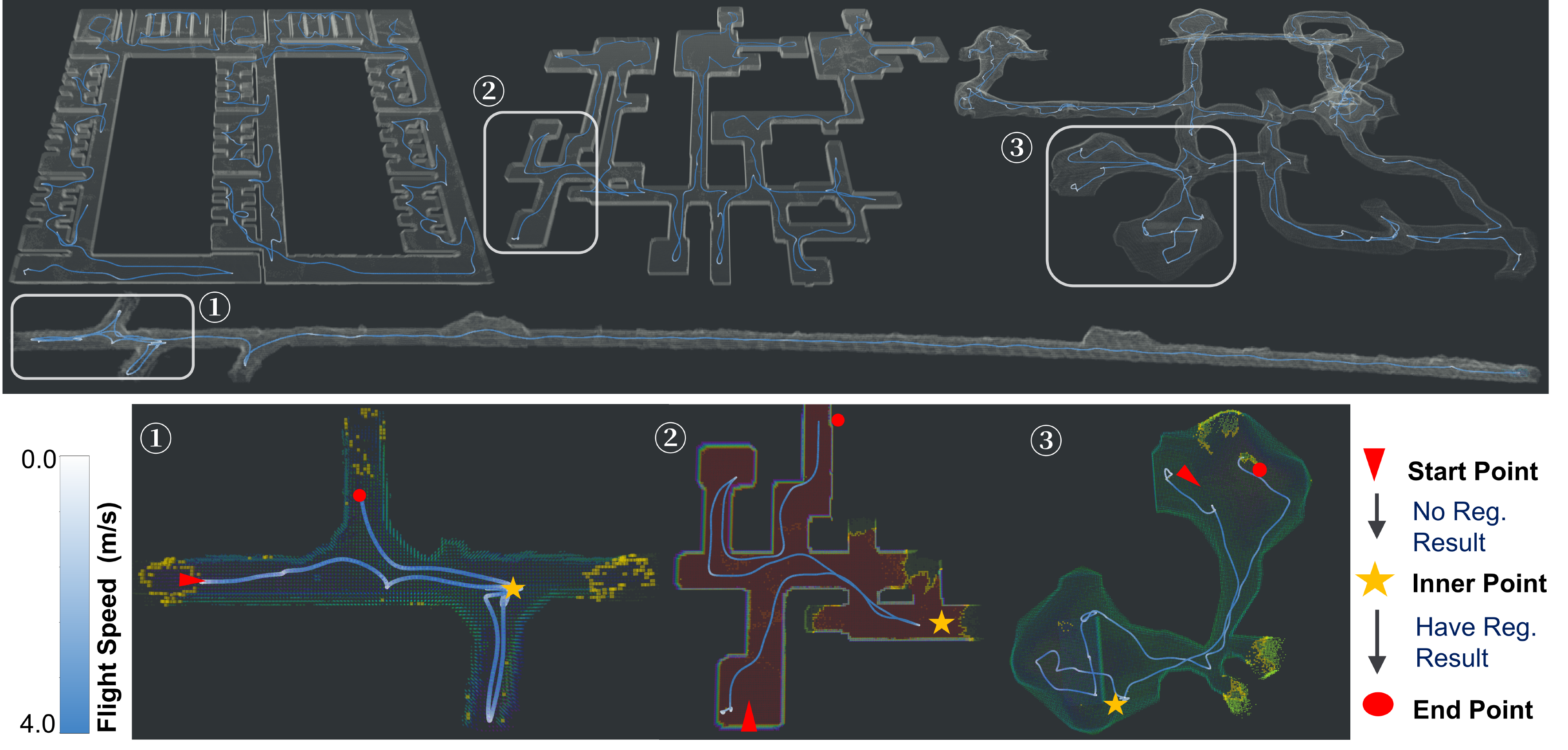}
  \caption{Exploration results in different environments with precise prior maps.
  The top row shows the global exploration trajectories in three environments. The bottom row illustrates the detailed exploration process. Departing from the start point (red triangle), the robot first explores without a valid registration result.
  After obtaining a successful registration result at the interior point (yellow star), the robot replans its exploration trajectory and moves toward the next target position (red dot) selected by the optimized global guidance.}
  \label{fig:exploration_benchmark}
\end{figure*}

Our method achieves the lowest TTC, the lowest registration error, and the highest success rate across all scenarios, while maintaining robustness against discrepancies.
In particular, it is the only method that achieves successful registration from very local environmental observations.

Notably, the method maintains a 100\% registration success rate across all deformed-prior trials.
Both single-frame and multi-frame registration modules inherently accommodate scale variations and structural deformations, while the Scale-ICP backend actively compensates for residual errors.
The robustness enhancement strategy furthermore alleviates the impact of registration failures.

Table~\ref{tab:capability_comparison} summarizes the qualitative capabilities of the compared registration methods. Our method is the only one that can simultaneously handle scale differences and deformations while supporting multiple solutions.

\begin{table*}[!tp]
  \centering
\renewcommand{\arraystretch}{0.92}\setlength{\tabcolsep}{4pt}
  \resizebox{\textwidth}{!}{%
\begin{tabular}{clccccccccccccc}
\toprule
 &  & \multicolumn{4}{c}{Exploration Time (s) $\downarrow$} & \multicolumn{4}{c}{Flight Distance (m) $\downarrow$} & \multicolumn{4}{c}{Avg. Velocity (m/s) $\uparrow$} &  \\
\multirow{-2}{*}{Scene} & \multirow{-2}{*}{Method} & Avg & Std & Max & Min & Avg & Std & Max & Min & Avg & Std & Max & Min & \multirow{-2}{*}{Fin.} \\ \midrule
 & FUEL & 1626.10 & 21.75 & 1645.23 & 1595.20 & 3091.71 & 47.49 & 3153.68 & 3037.93 & 1.90 & \cellcolor[HTML]{E1BEE7}0.01 & 1.92 & 1.89 & 4 \\
 & Luperto et al.+EPIC & 848.73 & 99.61 & 958.00 & 719.30 & 2900.67 & 370.81 & 3311.11 & 2410.44 & 3.41 & 0.08 & 3.52 & 3.34 & 5 \\
 & EPIC & 708.60 & 13.24 & 724.60 & 689.30 & 2346.42 & 64.06 & 2432.37 & 2271.05 & 3.31 & 0.08 & 3.40 & 3.22 & 5 \\
\cmidrule(lr){2-2}
 & \textbf{Proposed(Prec.)} & \cellcolor[HTML]{E1BEE7}\textbf{601.28} & 12.56 & \cellcolor[HTML]{E1BEE7}618.20 & \cellcolor[HTML]{E1BEE7}587.00 & \cellcolor[HTML]{E1BEE7}\textbf{2052.09} & \cellcolor[HTML]{E1BEE7}22.71 & \cellcolor[HTML]{E1BEE7}2068.90 & \cellcolor[HTML]{C8E6C9}2015.60 & 3.41 & 0.05 & 3.47 & 3.35 & 5 \\
 & \textbf{Proposed(Part.)} & 625.03 & \cellcolor[HTML]{E1BEE7}9.34 & 637.26 & 613.60 & 2187.94 & 58.16 & 2239.91 & 2107.63 & \cellcolor[HTML]{E1BEE7}\textbf{3.50} & 0.06 & \cellcolor[HTML]{E1BEE7}3.55 & \cellcolor[HTML]{E1BEE7}3.41 & 5 \\
 & \textbf{Proposed(Deform.)} & 609.38 & 10.98 & \cellcolor[HTML]{C8E6C9}624.81 & 596.00 & 2110.75 & 66.11 & 2206.37 & 2044.07 & \cellcolor[HTML]{C8E6C9}3.46 & 0.06 & \cellcolor[HTML]{C8E6C9}3.53 & 3.39 & 5 \\
\cmidrule(lr){2-2}
 & Proposed w/o MRH & \cellcolor[HTML]{C8E6C9}608.87 & 14.76 & 629.24 & \cellcolor[HTML]{C8E6C9}592.12 & \cellcolor[HTML]{C8E6C9}2063.29 & 56.51 & \cellcolor[HTML]{C8E6C9}2131.83 & \cellcolor[HTML]{E1BEE7}1991.18 & 3.39 & 0.02 & 3.41 & 3.36 & 5 \\
 & Proposed w/o FE-TSP & 652.65 & \cellcolor[HTML]{C8E6C9}9.59 & 663.79 & 641.96 & 2190.22 & 38.32 & 2230.65 & 2140.54 & 3.36 & \cellcolor[HTML]{C8E6C9}0.01 & 3.37 & 3.33 & 5 \\
 & Proposed w/o MCTS & 677.76 & 10.00 & 686.10 & 661.10 & 2312.30 & \cellcolor[HTML]{C8E6C9}33.95 & 2338.10 & 2253.90 & 3.41 & 0.01 & 3.44 & \cellcolor[HTML]{C8E6C9}3.40 & 5 \\
\cmidrule(lr){2-2}
\multirow{-10}{*}{Garage} & \cellcolor{black!6}Proposed w/ Oracle Reg.\textsuperscript{$\dagger$} & \cellcolor{black!6}592.97 & \cellcolor{black!6}8.58 & \cellcolor{black!6}603.05 & \cellcolor{black!6}583.50 & \cellcolor{black!6}2020.36 & \cellcolor{black!6}28.64 & \cellcolor{black!6}2058.42 & \cellcolor{black!6}1989.29 & \cellcolor{black!6}3.41 & \cellcolor{black!6}0.02 & \cellcolor{black!6}3.43 & \cellcolor{black!6}3.38 & \cellcolor{black!6}5 \\
\midrule
 & FUEL & - & - & - & - & - & - & - & - & - & - & - & - & 0 \\
 & Luperto et al.+EPIC & 257.94 & \cellcolor[HTML]{E1BEE7}2.96 & 261.40 & 254.50 & 960.04 & \cellcolor[HTML]{E1BEE7}5.65 & 963.98 & 950.16 & \cellcolor[HTML]{E1BEE7}\textbf{3.72} & 0.06 & \cellcolor[HTML]{E1BEE7}3.78 & \cellcolor[HTML]{E1BEE7}3.65 & 5 \\
 & EPIC & 276.36 & 6.48 & 284.30 & 267.00 & 986.50 & \cellcolor[HTML]{C8E6C9}7.35 & 996.47 & 976.88 & \cellcolor[HTML]{C8E6C9}3.57 & 0.06 & \cellcolor[HTML]{C8E6C9}3.66 & \cellcolor[HTML]{C8E6C9}3.50 & 5 \\
\cmidrule(lr){2-2}
 & \textbf{Proposed(Prec.)} & \cellcolor[HTML]{C8E6C9}181.96 & 6.65 & \cellcolor[HTML]{E1BEE7}189.20 & \cellcolor[HTML]{C8E6C9}173.20 & \cellcolor[HTML]{C8E6C9}614.43 & 10.73 & \cellcolor[HTML]{E1BEE7}622.93 & \cellcolor[HTML]{C8E6C9}598.35 & 3.38 & 0.07 & 3.45 & 3.28 & 5 \\
 & \textbf{Proposed(Part.)} & \cellcolor[HTML]{E1BEE7}\textbf{181.62} & \cellcolor[HTML]{C8E6C9}5.50 & \cellcolor[HTML]{C8E6C9}189.30 & 176.20 & \cellcolor[HTML]{E1BEE7}\textbf{612.34} & 11.13 & \cellcolor[HTML]{C8E6C9}627.80 & 599.27 & 3.37 & \cellcolor[HTML]{E1BEE7}0.04 & 3.41 & 3.32 & 5 \\
\cmidrule(lr){2-2}
 & Proposed w/o MRH & 186.94 & 14.54 & 208.40 & \cellcolor[HTML]{E1BEE7}170.40 & 617.84 & 17.02 & 637.48 & \cellcolor[HTML]{E1BEE7}597.20 & 3.32 & 0.17 & 3.50 & 3.06 & 5 \\
 & Proposed w/o FE-TSP & 219.01 & 11.31 & 230.80 & 203.70 & 753.63 & 32.78 & 784.89 & 705.13 & 3.44 & \cellcolor[HTML]{C8E6C9}0.04 & 3.50 & 3.39 & 5 \\
 & Proposed w/o MCTS & 278.87 & 15.13 & 296.40 & 258.70 & 991.31 & 33.64 & 1029.20 & 943.00 & 3.56 & 0.08 & 3.65 & 3.47 & 5 \\
\cmidrule(lr){2-2}
\multirow{-9}{*}{Simple Cave} & \cellcolor{black!6}Proposed w/ Oracle Reg.\textsuperscript{$\dagger$} & \cellcolor{black!6}173.78 & \cellcolor{black!6}8.21 & \cellcolor{black!6}184.00 & \cellcolor{black!6}164.00 & \cellcolor{black!6}575.50 & \cellcolor{black!6}28.50 & \cellcolor{black!6}610.00 & \cellcolor{black!6}540.50 & \cellcolor{black!6}3.31 & \cellcolor{black!6}0.01 & \cellcolor{black!6}3.33 & \cellcolor{black!6}3.30 & \cellcolor{black!6}5 \\
\midrule
 & FUEL & 1543.48 & 87.15 & 1658.98 & 1447.68 & 2457.32 & 134.95 & 2631.05 & 2304.21 & 1.59 & \cellcolor[HTML]{E1BEE7}0.02 & 1.61 & 1.58 & 4 \\
 & Luperto et al.+EPIC & 668.66 & 65.28 & 726.00 & 564.30 & 2324.52 & 244.56 & 2537.52 & 1937.73 & 3.47 & 0.03 & 3.50 & 3.43 & 5 \\
 & EPIC & 525.20 & 10.89 & 536.00 & 509.00 & 1868.07 & 36.75 & 1916.76 & 1832.24 & 3.56 & 0.04 & 3.60 & 3.49 & 5 \\
\cmidrule(lr){2-2}
 & \textbf{Proposed(Prec.)} & \cellcolor[HTML]{E1BEE7}\textbf{474.30} & \cellcolor[HTML]{E1BEE7}7.41 & \cellcolor[HTML]{E1BEE7}482.00 & \cellcolor[HTML]{E1BEE7}464.00 & 1722.46 & 30.89 & 1770.00 & 1693.10 & \cellcolor[HTML]{E1BEE7}\textbf{3.63} & 0.05 & \cellcolor[HTML]{E1BEE7}3.67 & \cellcolor[HTML]{E1BEE7}3.56 & 5 \\
 & \textbf{Proposed(Part.)} & \cellcolor[HTML]{C8E6C9}475.71 & 8.97 & \cellcolor[HTML]{C8E6C9}483.55 & \cellcolor[HTML]{C8E6C9}466.00 & \cellcolor[HTML]{E1BEE7}\textbf{1707.88} & \cellcolor[HTML]{C8E6C9}30.44 & \cellcolor[HTML]{E1BEE7}1738.93 & \cellcolor[HTML]{C8E6C9}1674.89 & \cellcolor[HTML]{C8E6C9}3.59 & \cellcolor[HTML]{C8E6C9}0.02 & 3.60 & \cellcolor[HTML]{C8E6C9}3.56 & 5 \\
 & \textbf{Proposed(Deform.)} & 482.81 & 16.06 & 509.00 & 466.00 & 1735.02 & 60.71 & 1832.24 & 1675.71 & 3.59 & 0.02 & 3.61 & 3.56 & 5 \\
\cmidrule(lr){2-2}
 & Proposed w/o MRH & 477.73 & \cellcolor[HTML]{C8E6C9}8.53 & 489.20 & 467.00 & \cellcolor[HTML]{C8E6C9}1712.74 & \cellcolor[HTML]{E1BEE7}28.69 & \cellcolor[HTML]{C8E6C9}1750.70 & \cellcolor[HTML]{E1BEE7}1673.46 & 3.59 & 0.05 & \cellcolor[HTML]{C8E6C9}3.67 & 3.54 & 5 \\
 & Proposed w/o FE-TSP & 518.56 & 14.48 & 536.00 & 499.80 & 1849.99 & 41.04 & 1916.76 & 1806.45 & 3.57 & 0.05 & 3.61 & 3.49 & 5 \\
 & Proposed w/o MCTS & 572.86 & 32.95 & 620.60 & 534.60 & 2033.32 & 120.71 & 2221.80 & 1919.60 & 3.55 & 0.04 & 3.59 & 3.51 & 5 \\
\cmidrule(lr){2-2}
\multirow{-10}{*}{Indoor} & \cellcolor{black!6}Proposed w/ Oracle Reg.\textsuperscript{$\dagger$} & \cellcolor{black!6}469.54 & \cellcolor{black!6}7.23 & \cellcolor{black!6}478.31 & \cellcolor{black!6}461.38 & \cellcolor{black!6}1703.89 & \cellcolor{black!6}32.34 & \cellcolor{black!6}1738.99 & \cellcolor{black!6}1668.24 & \cellcolor{black!6}3.63 & \cellcolor{black!6}0.02 & \cellcolor{black!6}3.66 & \cellcolor{black!6}3.60 & \cellcolor{black!6}5 \\
\midrule
 & FUEL & - & - & - & - & - & - & - & - & - & - & - & - & 0 \\
 & Luperto et al.+EPIC & 1091.95 & 62.20 & 1170.30 & 1021.00 & 3611.87 & 223.91 & 3890.44 & 3354.00 & \cellcolor[HTML]{E1BEE7}\textbf{3.31} & \cellcolor[HTML]{C8E6C9}0.02 & \cellcolor[HTML]{E1BEE7}3.32 & \cellcolor[HTML]{E1BEE7}3.29 & 5 \\
 & EPIC & 1032.85 & 24.65 & 1065.20 & 1005.20 & 3059.12 & \cellcolor[HTML]{E1BEE7}37.99 & 3105.20 & 3012.38 & 2.96 & 0.06 & 3.05 & 2.92 & 5 \\
\cmidrule(lr){2-2}
 & \textbf{Proposed(Part.)} & \cellcolor[HTML]{E1BEE7}\textbf{897.04} & \cellcolor[HTML]{E1BEE7}15.08 & \cellcolor[HTML]{E1BEE7}910.20 & \cellcolor[HTML]{C8E6C9}875.52 & \cellcolor[HTML]{E1BEE7}\textbf{2692.07} & \cellcolor[HTML]{C8E6C9}50.36 & \cellcolor[HTML]{E1BEE7}2750.20 & \cellcolor[HTML]{C8E6C9}2633.60 & 3.00 & 0.03 & 3.04 & 2.97 & 5 \\
 & \textbf{Proposed(Deform.)} & \cellcolor[HTML]{C8E6C9}906.18 & 28.85 & 936.20 & \cellcolor[HTML]{E1BEE7}869.55 & \cellcolor[HTML]{C8E6C9}2700.27 & 83.38 & \cellcolor[HTML]{C8E6C9}2782.70 & \cellcolor[HTML]{E1BEE7}2587.86 & 2.98 & \cellcolor[HTML]{E1BEE7}0.01 & 3.00 & 2.97 & 5 \\
\cmidrule(lr){2-2}
 & Proposed w/o MRH & 910.94 & \cellcolor[HTML]{C8E6C9}22.28 & \cellcolor[HTML]{C8E6C9}934.60 & 879.70 & 2727.13 & 75.02 & 2807.54 & 2634.70 & 2.99 & 0.02 & 3.01 & 2.98 & 5 \\
 & Proposed w/o FE-TSP & 1077.23 & 69.68 & 1165.78 & 995.67 & 3351.13 & 239.11 & 3607.23 & 3066.58 & \cellcolor[HTML]{C8E6C9}3.11 & 0.04 & \cellcolor[HTML]{C8E6C9}3.16 & \cellcolor[HTML]{C8E6C9}3.07 & 5 \\
 & Proposed w/o MCTS & 1168.18 & 91.11 & 1245.67 & 1023.45 & 3622.49 & 285.82 & 3896.49 & 3172.40 & 3.10 & 0.03 & 3.13 & 3.06 & 5 \\
\cmidrule(lr){2-2}
\multirow{-9}{*}{SubT Cave} & \cellcolor{black!6}Proposed w/ Oracle Reg.\textsuperscript{$\dagger$} & \cellcolor{black!6}883.84 & \cellcolor{black!6}17.18 & \cellcolor{black!6}899.80 & \cellcolor{black!6}862.10 & \cellcolor{black!6}2631.68 & \cellcolor{black!6}61.13 & \cellcolor{black!6}2706.39 & \cellcolor{black!6}2549.44 & \cellcolor{black!6}2.98 & \cellcolor{black!6}0.02 & \cellcolor{black!6}3.01 & \cellcolor{black!6}2.96 & \cellcolor{black!6}5 \\
\bottomrule
\end{tabular}%
  }
  \caption{Statistics of the exploration benchmark. "Fin." denotes the number of successfully finished trials out of 5. The best and second-best results are highlighted in \textcolor[HTML]{E1BEE7}{\rule{1em}{1em}} and \textcolor[HTML]{C8E6C9}{\rule{1em}{1em}}, respectively. Two scenarios are not tested, namely simple cave with a deformed map, because a meaningful deformation is hard to design for so simple a structure, and SubT cave with a precise map, because that environment has vertical features that a 2D map cannot represent. \textsuperscript{$\dagger$}\,The shaded \emph{Proposed w/ Oracle Reg.} row is excluded from the best and second-best ranking because it is a hypothetical study rather than an ablation study.}
  \label{tab:exploration_benchmark}
\end{table*}

\subsection{EXPLORATION BENCHMARK}

In this section, we compare the performance of the proposed exploration framework with several baseline methods including \textbf{FUEL}~\cite{zhou2021fuel}, \textbf{EPIC}~\cite{geng2025epic}, and \textbf{Luperto et al.}~\cite{luperto2020robot}.
We tested our method's performance under different levels of discrepancies between the prior map and the real-world environment.

\begin{itemize}
    \item \textbf{FUEL:} This approach proposes a frontier information structure (FIS) to represent the environment and guide the robot to explore unknown areas.
Candidate viewpoints are then sampled uniformly within the cylindrical coordinate system to fully cover the frontier cluster.
An ATSP problem is solved to determine the optimal global tour visiting all viewpoints.
    \item \textbf{EPIC:} This approach generates frontiers on point cloud instead of grid cells and solves an ATSP among sampled viewpoints.
We adopt the same realization of all other modules, thus the improvement of our method over EPIC is primarily attributable to the registration and the proposed planning framework.
    \item \textbf{Luperto et al.:} This method improves robot exploration by using prior knowledge (e.g., floor plans) to estimate how much new area a location will reveal.
Instead of optimistically assuming all unknown space is reachable, it checks the prior map to see if the area is free and visible, enabling smarter, faster navigation.
We extend the method into 3D space using EPIC's interface, thus the comparison between our method and this method for handling prior maps is fair.
\end{itemize}

We also conduct several ablation and hypothetical studies to evaluate the effectiveness of each planning module.

\begin{itemize}
  \item \textbf{Proposed w/o MRH:} We only consider one registration result to guide the planning module. The risk-aware selector is removed, and the registration hypothesis with the highest confidence is directly selected for planning.
  \item \textbf{Proposed w/o FE-TSP:} We remove the FE-TSP module and directly solve the unconstrained ATSP problem among selected viewpoints.
  \item \textbf{Proposed w/o MCTS:} We remove the MCTS-based prior guidepoints sequencing process and greedily choose the nearest and second nearest prior guidepoints as the next two targets. Since a greedy rule of this kind never evaluates a complete sequence, no travel risk can be computed for it, and we therefore also disable the multi-hypothesis strategy and commit to the highest-confidence hypothesis in this ablation. The two choices are consistent, because taking the highest-confidence hypothesis is itself a greedy decision that ignores global information.
  \item \textbf{Proposed w/ Oracle Reg.:} We assume the ground-truth registration is available from the beginning of the exploration, so that we can evaluate the performance of the proposed planning framework without the influence of the registration module. This is a hypothetical study rather than an ablation study, and is therefore excluded from the best and second-best rankings in Table~\ref{tab:exploration_benchmark}.
\end{itemize}

We use the MARSIM~\cite{kong2023marsim} simulator to provide identical, high-fidelity sensor input for all methods.
The input sketches include precise, partial, and deformed prior maps.
Each method runs five times in each environment with a maximum velocity of $\qty{4.0}{\metre\per\second}$, map resolution of $\qty{0.3}{\metre}$ (if using a grid map), and a time limit of $\qty{2000}{\second}$.
A test fails if the explored area is less than 95\% or the time runs out.

Notably, two scenarios are not tested, namely simple cave with a deformed map and SubT cave with a precise map.
The former is due to the simple structure of the environment, which makes it difficult to design a geometrically deformed map. The latter is because the environment has vertical features that cannot be represented in a 2D map. More discussion can be found in Section~\ref{sec:limitations}.

Table~\ref{tab:exploration_benchmark} shows that the proposed method reduces exploration time and flight distance in all tested settings.
Furthermore, our approach remains efficient with deformed or partial maps. Deformations minimally impact the global topological guidance, with minor performance drops mainly stemming from the additional registration time.
For partial maps, the framework prioritizes exploring prior-missing regions and seamlessly re-establishes the guidance once missing areas are covered.
As long as the core topology is preserved, the prior remains effective at reducing unnecessary revisits.
In addition, by consistently following the global path, our method avoids random exploratory behaviors, yielding stable performance with low variance across most metrics.
The average flight velocity remains similar or slightly lower compared to EPIC and Luperto et al., because following the global guidance may involve more deceleration maneuvers to ensure thorough local exploration. The reduced exploration time mainly comes from more efficient global coverage rather than higher flight speed.

Each planning module contributes to the overall performance.
The MCTS-based prior guidepoints sequencing process effectively enhances performance by providing more efficient global ordering, while the FE-TSP module ensures that the selected viewpoints are visited in an optimal order. The risk-aware registration hypothesis selector improves the robustness of the planning framework. By scoring every candidate
sequence with the confidence weights and the resulting travel risk, it chooses the most reliable sequence based on global information and enhances the stability of the exploration process.
Section~S5 of the supplementary material reports the settings of these ablations and the resulting trajectories, and Section~S6 examines how faithfully the prior-map planning cost predicts the realized flight cost under precise, partial, and deformed priors.

Notably, although the method of Luperto et al. also uses prior maps, it performs poorly in our tests.
Its greedy strategy maximizes the expected revealed area based on the prior map, which only improves exploration efficiency in the early stages.
Because it focuses only on local information gain and ignores the global topology, the UAV subsequently spends significantly more time detouring to cover small missed areas.
In contrast, our method uses a global coverage path throughout the process, achieving better overall results.

The trajectory visualizations are shown in Figure~\ref{fig:exploration_benchmark}.
A per-module runtime and memory breakdown, measured both in simulation and on the on-board computer, is given in Section~S8 of the supplementary material.
In the early exploration phase, the UAV may fly in suboptimal directions before a successful registration. However, once registered, it quickly corrects its path and follows the global guidance.
Our framework ensures early registration, which minimizes wasted time and improves overall efficiency.

For instance, in environments like the simple cave, the lack of global guidance forces the baselines into long detours to revisit missed areas. Our method avoids these detours, achieving a 34.2\% reduction in exploration time and a 37.9\% reduction in flight distance.
However, in highly connected environments like garages, the performance gap is smaller because less backtracking is needed.
Nevertheless, the consistent global guidance still minimizes unnecessary revisits and yields better overall performance.

\section{FIELD EXPERIMENTS}

\begin{figure*}
  \centering
  \includegraphics[width=\linewidth]{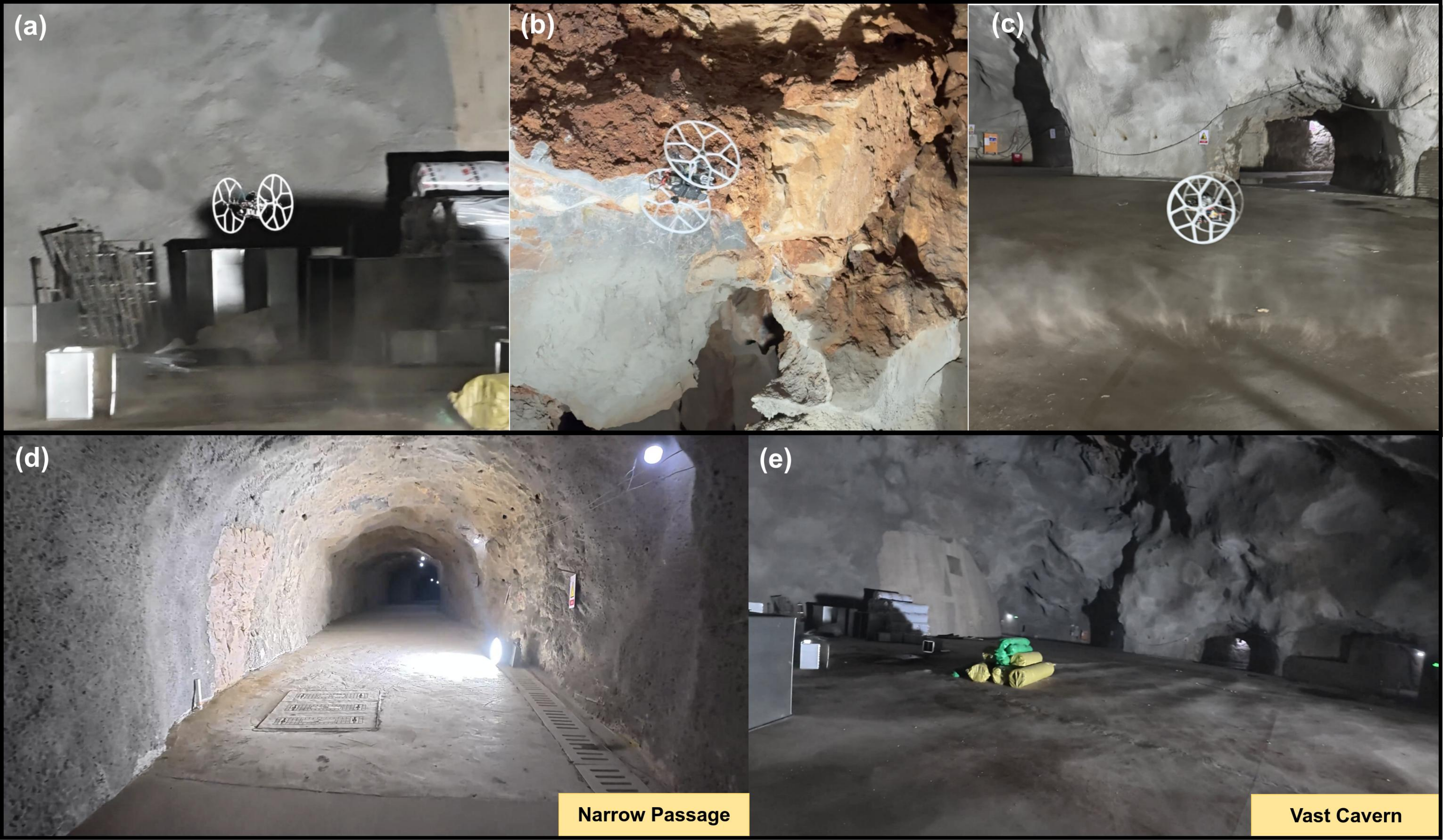}
  \caption{Experimental environment. (a)(b)(c) Third person view of the UAV flying in the wild cave.
(d)(e) Onboard view of the UAV flying in the wild cave.}
  \label{fig:map_for_exploration}
\end{figure*}
\begin{figure}
  \centering
  \includegraphics[width=\linewidth]{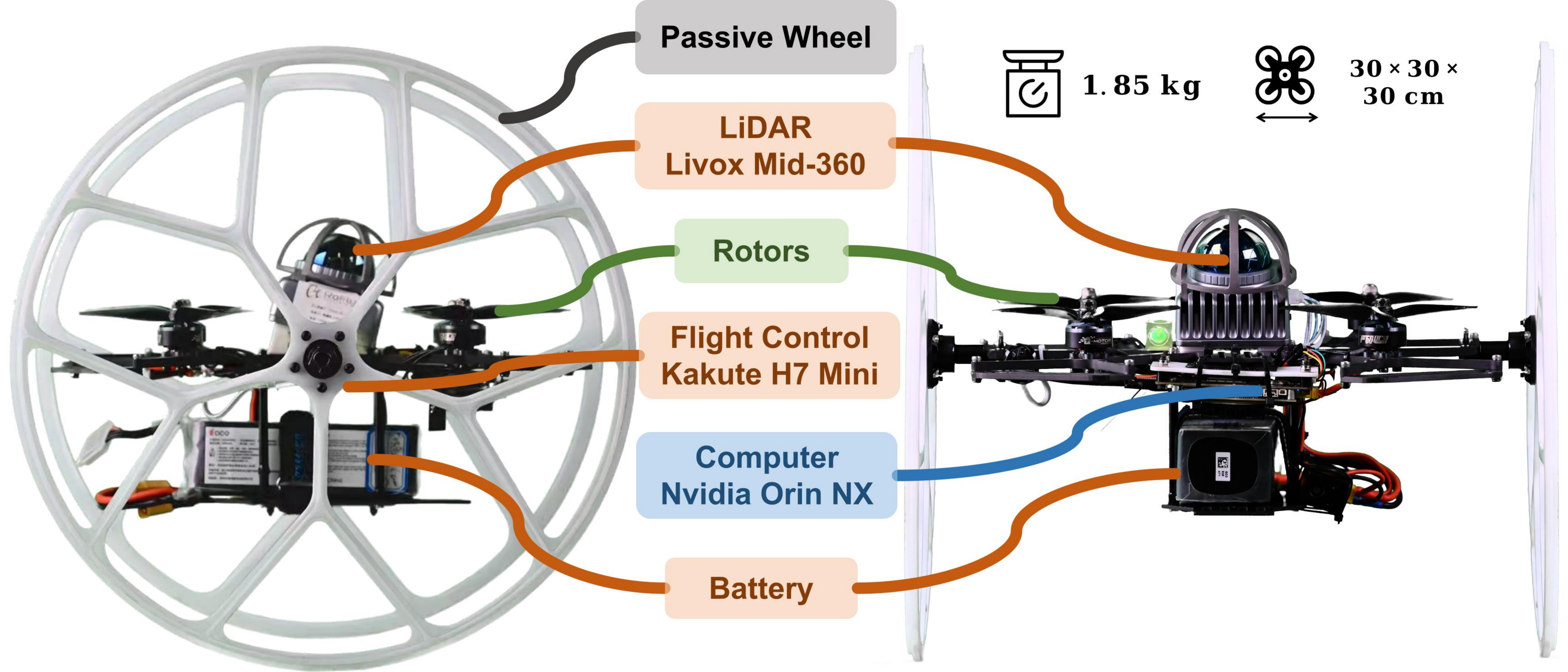}
  \caption{The LiDAR-based UAV platform used in field experiments.
(a) The left view.
(b) The front view.}
  \label{fig:UAV_platform}
\end{figure}

We conduct field experiments in a large-scale wild cave to validate the proposed registration and exploration framework.
Three main experiments are conducted: single-frame candidate retrieval tests, overall registration tests, and autonomous exploration tests under precise, partial, and deformed maps.
\subsection{EXPERIMENTAL SETUP AND IMPLEMENTATION DETAILS}

We utilize a LiDAR-based quadrotor platform equipped with a Kakute H7 Mini flight controller and a Livox Mid-360 LiDAR mounted at a \qty{15}{\degree} pitch angle.
Two passive wheels are mounted to protect the vehicle body during contact with the cave surface and do not provide any locomotion capability.
The onboard computer is an NVIDIA Jetson Orin NX, which processes the SLAM, registration, exploration, and control algorithms in real time.
FAST-LIO2~\cite{xu2022fast} is adopted for localization and mapping.
We use an extended Kalman filter (EKF) to fuse IMU measurements and odometry to generate high-frequency pose estimations.
The control algorithm proposed in~\cite{omari2013nonlinear} is adopted for high-accuracy onboard $SE(3)$ trajectory tracking.
We also adopt the differential flatness property~\cite{wang2022robust} for feedforward compensation of control variables.
The maximum speed of the UAV is limited to $\qty{3.0}{\metre\per\second}$ for safety considerations.
The UAV platform is shown in Figure~\ref{fig:UAV_platform}.

\begin{figure*}[!tp]
  \centering
  \includegraphics[width=\textwidth]{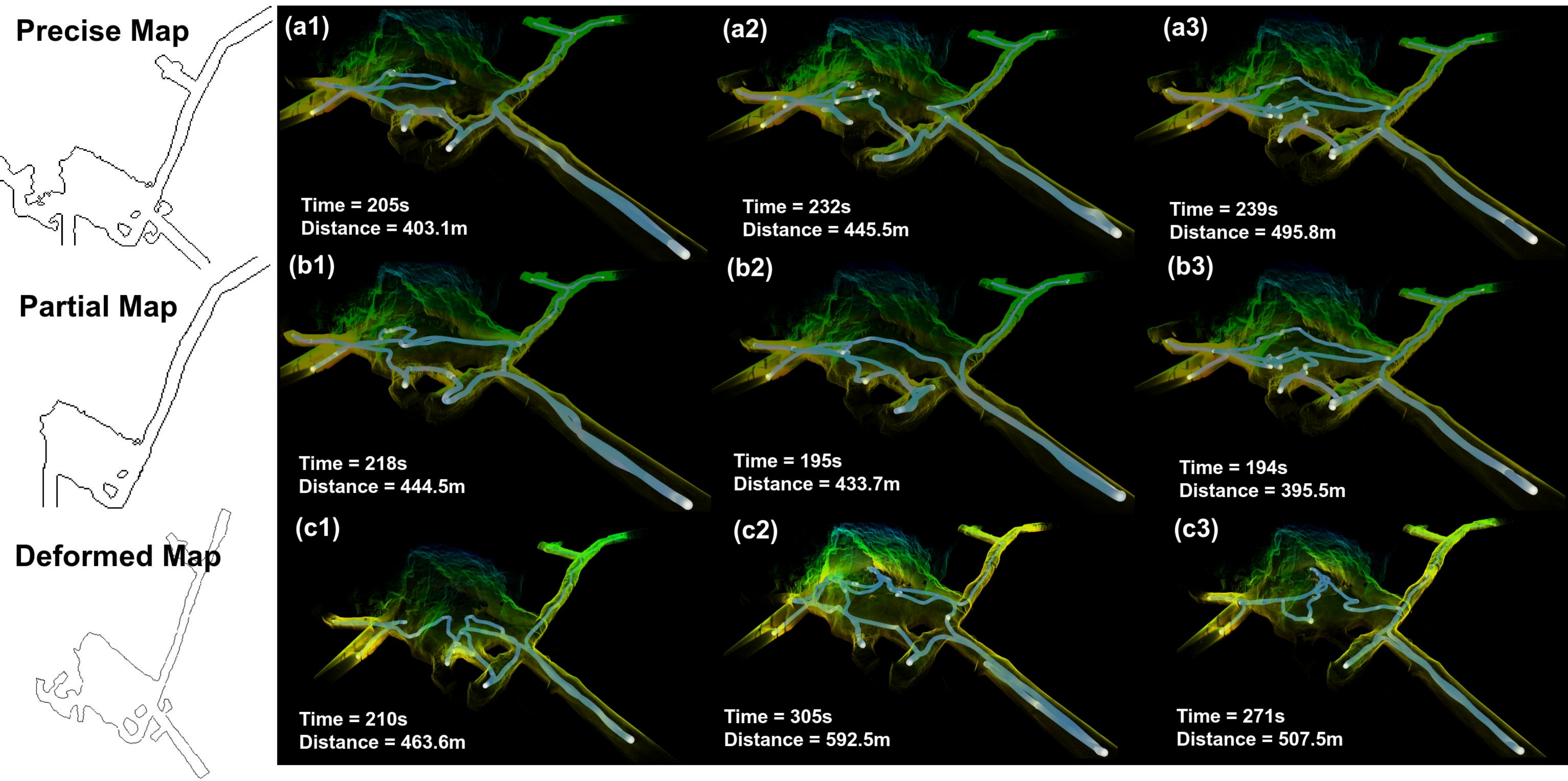}
  \caption{The experimental results under different prior maps.
(a1)-(a3) Results under precise prior map.
(b1)-(b3) Results under partial prior map.
(c1)-(c3) Results under deformed prior map.
The exploration time and distance may vary because of different layouts of obstacles in the cave.}
  \label{fig:environment_and_maps_exp}
\end{figure*}

The experimental environment is a natural wild cave with an approximate range of $[\qtyproduct{70 x 110 x 3}{\metre}]$.
Figure~\ref{fig:map_for_exploration} shows representative third-person and onboard views of the experimental environment.
Three testing conditions are provided for evaluation: a precise map created by scanning projection; a deformed map derived from the original blueprint with certain scaling and morphological discrepancies; and a partial map obtained by removing specific topological regions from the precise map.
The prior maps and the experimental results are shown in Figure~\ref{fig:environment_and_maps_exp}.

\subsection{SINGLE-FRAME CANDIDATE RETRIEVAL TEST}
\begin{table}[]
\centering
\resizebox{\columnwidth}{!}{%
\begin{tabular}{cccc}
\toprule
\multicolumn{2}{c}{Precise Map} & \multicolumn{2}{c}{Deformed Map} \\
Candidate Rate & Succ. Rate $\uparrow$ & Candidate Rate & Succ. Rate $\uparrow$ \\ \midrule
10\% & 96.08\% & 10\% & 92.97\% \\
5\% & 89.74\% & 5\% & 82.15\% \\
2.50\% & 76.92\% & 2.50\% & 71.43\% \\ \bottomrule
\end{tabular}%
}
\caption{Single-frame candidate retrieval results under precise and deformed prior maps with different candidate rates.}
\label{tab:single_frame_exp}
\end{table}

We collect data by letting the UAV platform explore the cave environment while recording its position and point cloud data.
After ten rounds of data collection, we perform the single-frame test at a frequency of \qty{1}{\hertz} and finally obtain 2348 frames of candidate data.
Since the cave was under construction during the experiments, obstacles changed frequently and were not present in the prior map.
We use the same criteria as in the simulation test to evaluate the success rate of the candidate retrieval.
As shown in Table~\ref{tab:single_frame_exp}, the tests on both precise and deformed maps achieve a high success rate.
The choice of candidate rate also affects the success rate.
A higher candidate rate improves the success rate but may increase the computational time, so a trade-off between the two should be considered when setting the candidate rate.
In practice, owing to the effective clustering and filtering, a 10\% candidate proportion still yields real-time performance while ensuring a high single-frame success rate and stable overall registration.

\begin{figure}
  \centering
  \includegraphics[width=\linewidth]{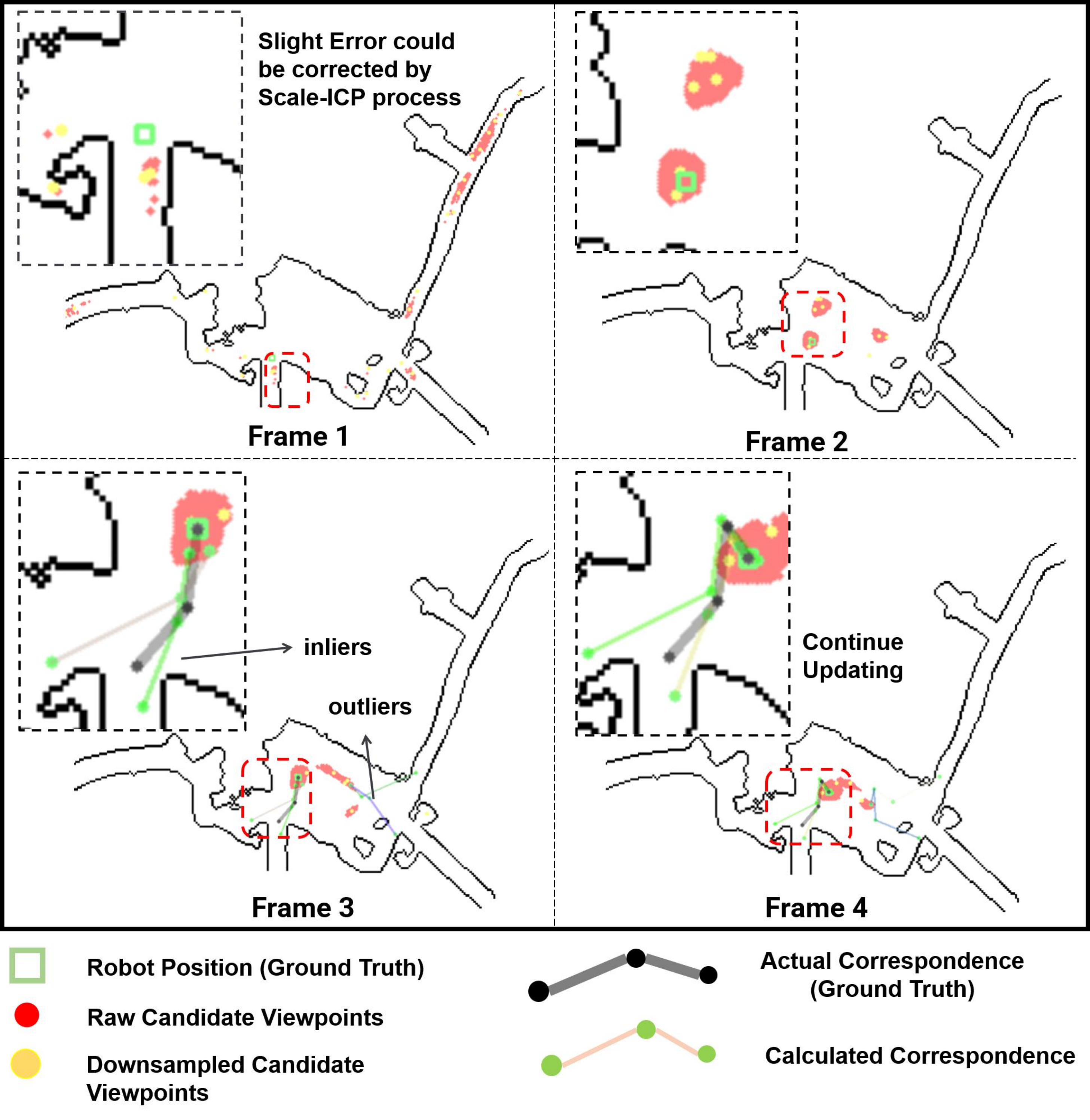}
  \caption{Visualization of the registration process in the field experiment.
}
  \label{fig:exp_reg_result}
\end{figure}

\subsection{OVERALL REGISTRATION TEST}

\begin{table}[]
\centering
\resizebox{\columnwidth}{!}{%
\begin{tabular}{cccc}
\toprule
 & TTC $\downarrow$ (s) & Err. $\downarrow$ (m) & Succ. Rate $\uparrow$ \\ \midrule
Precise Map & 32.4 & 1.02 & 100\% \\
Partial Map & 32.9 & 1.27 & 100\% \\
Deformed Map & 39.5 & 1.57 & 100\% \\ \bottomrule
\end{tabular}%
}
\caption{Overall registration results under different types of prior maps.}
\label{tab:overall_registration_exp}
\end{table}
\begin{figure*}[!h]
  \centering
  \includegraphics[width=\linewidth]{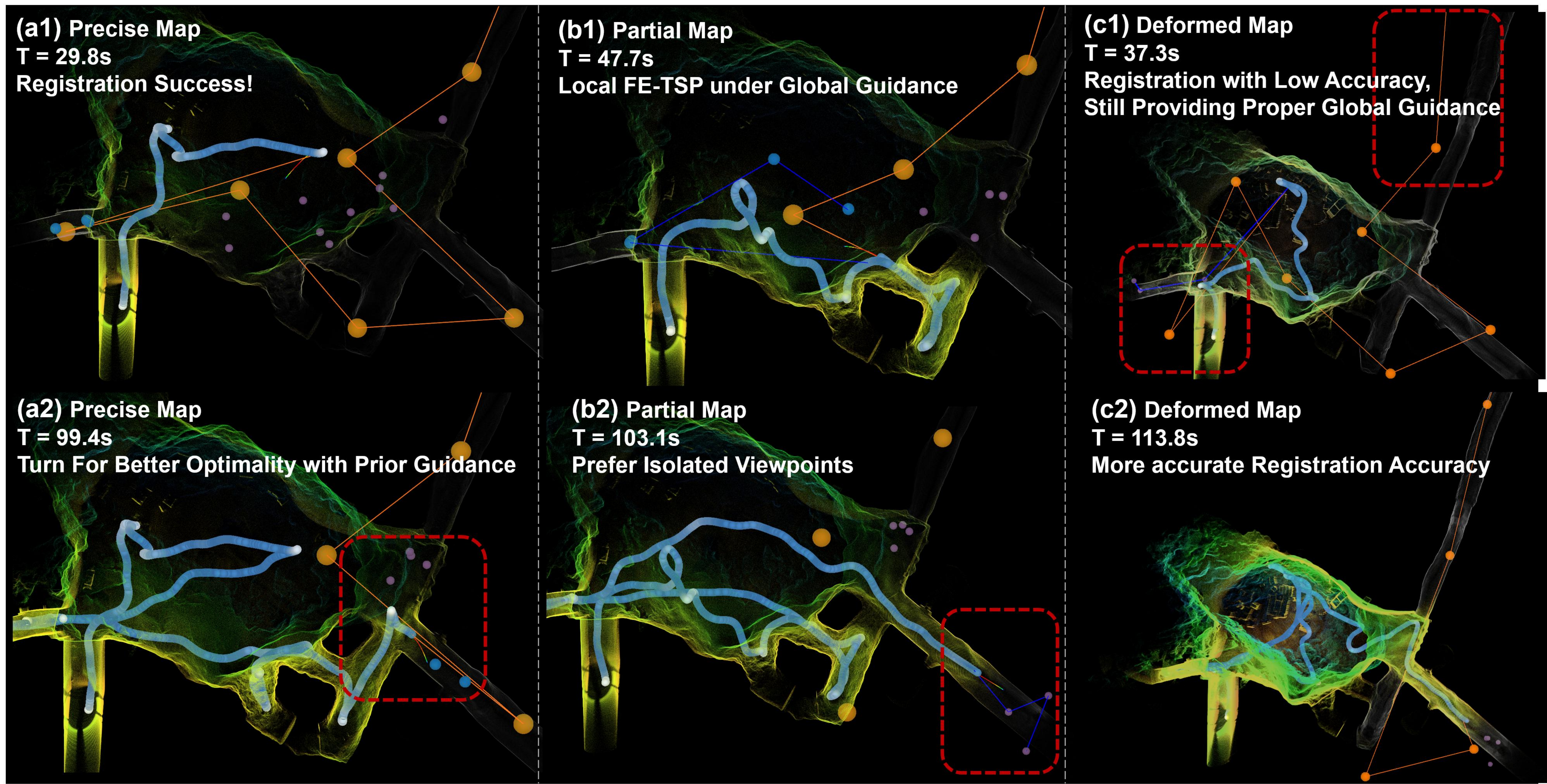}
  \caption{Visualization of representative real-world exploration cases.
(a1)(a2) Exploration under a precise prior map.
(b1)(b2) Exploration under a partial prior map.
(c1)(c2) Exploration under a deformed prior map.}
  \label{fig:explore_exp_explain}
\end{figure*}

We utilize the ten collected point cloud datasets to perform the overall registration tests.
To obtain more diverse testing conditions, we sample $N_s = 10$ frames uniformly from all frames as starting frames to simulate ten different testing rounds with varying initial conditions and trajectories.
This results in 100 valid overall registration rounds.
We adopt the same evaluation metrics as in the simulation test.
Table~\ref{tab:overall_registration_exp} shows that our method achieves a 100\% success rate across all types of prior maps.
The TTC also remains consistently low at $\qtyrange[range-phrase = \sim]{32.4}{39.5}{\second}$.
The registration accuracy is slightly lower than in the simulation test, primarily due to the inevitable discrepancies between the prior map and the real-world environment, as the cave's rough and irregular surfaces limit the precision of the sketch.

We visualize an example of the registration process in Figure~\ref{fig:exp_reg_result}.
In Frames 1 and 2, only the single-frame module is applied, which successfully selects the correct candidates.
Despite certain errors (Frame 1), the multi-frame verification mechanism identifies the inliers and refines them via the Scale-ICP process.
In Frame 3, the coarse registration results are established, and in Frame 4, the results are continuously updated for a better initial guess.

Figure~\ref{fig:cover}(d) illustrates the effectiveness of the Scale-ICP algorithm. Registration errors are significantly reduced, demonstrating robustness to coarse transformation inaccuracies and prior-map discrepancies.

\subsection{AUTONOMOUS EXPLORATION TEST}
We conduct autonomous exploration tests using the UAV platform with different types of prior maps.
The exploration process is recorded, and the exploration time, flight distance, and average speed are calculated for evaluation.
Three trials are conducted for each type of prior map, resulting in a total of nine trials.
The travel trajectories, exploration times, and flight distances are shown in Figure~\ref{fig:environment_and_maps_exp}.
Details of the exploration process are shown in Figure~\ref{fig:explore_exp_explain}.
More details of the exploration experiment can be found in the video attachment.

\begin{figure}
  \centering
  \includegraphics[width=\linewidth]{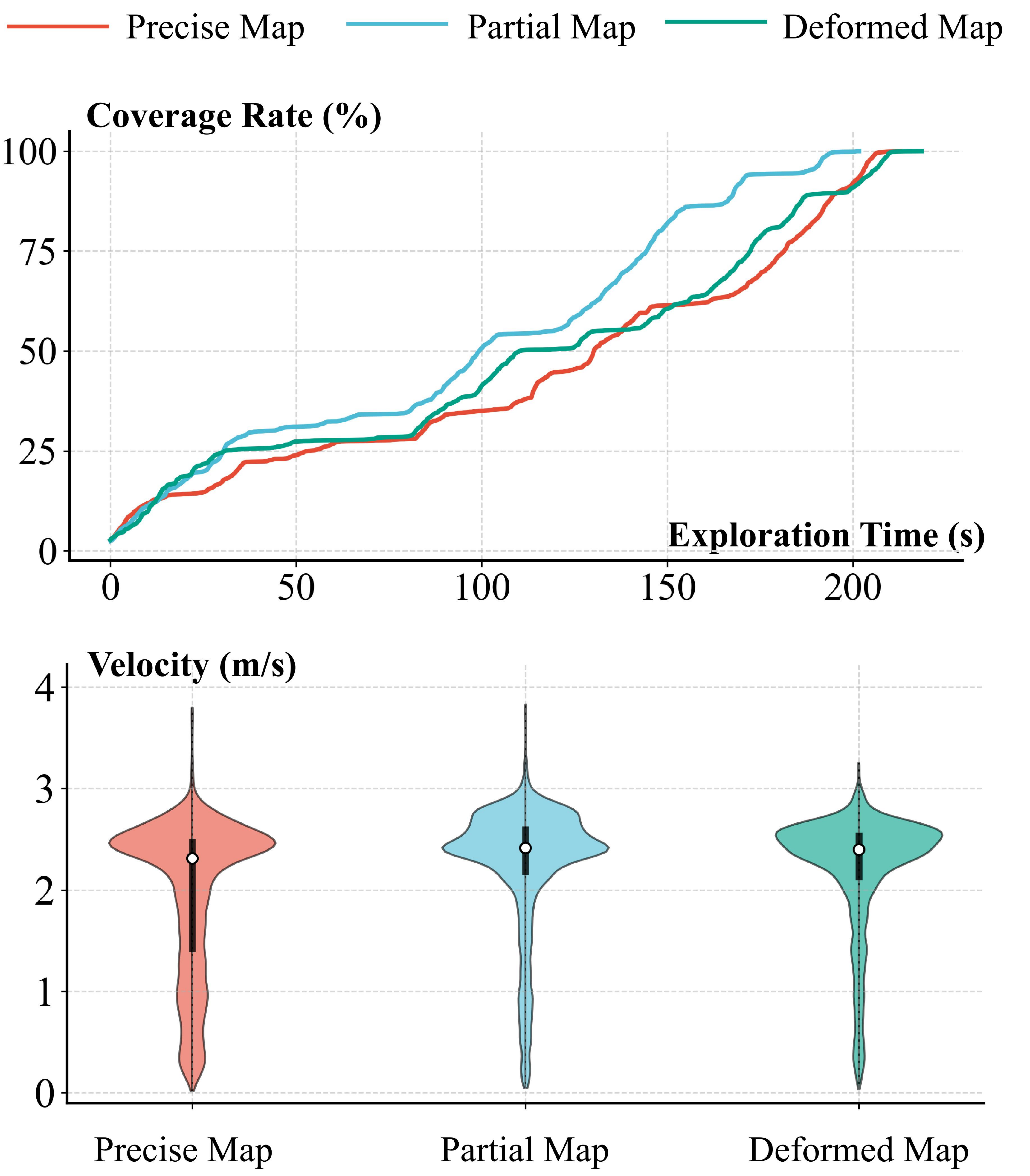}
\caption{Coverage rate and velocity distribution in field experiments with different prior maps.
The top plot shows the coverage rate over exploration time, and the bottom plot shows the UAV velocity distributions.}
  \label{fig:exp_vel}
\end{figure}

The exploration time and flight distance may vary across rounds due to changing obstacle layouts during the cave's construction, as complex layouts require more coverage time.
Despite this, our UAV consistently generates efficient global guidance paths from registration results and follows them.

Furthermore, the exploration path is steady and efficient under all prior maps. Figure~\ref{fig:exp_vel} shows that the coverage rate trends are similar across the three prior-map types, indicating that the planner maintains consistent exploration behavior despite prior-map discrepancies. The coverage rates also increase steadily without long pauses, and the velocity distributions are concentrated around high values, reflecting few detours and stable, efficient motion.

As shown in Figure~\ref{fig:explore_exp_explain}(a1), at the beginning of the exploration process, no reliable registration result is available, causing the UAV to head in a suboptimal direction (the optimal direction is leftward).
After exploring for \qty{29.8}{\second}, a reliable registration result is obtained, allowing the UAV to correct its trajectory and fly toward the left area.
At $T = \qty{99.4}{\second}$, where the UAV reaches a crossroad in the cave, it quickly recognizes the unexplored right side and changes its direction to explore it (Figure~\ref{fig:explore_exp_explain}(a2)).

The effect of the local path planning is demonstrated with partial prior maps, where the local planning module guides the UAV to prioritize exploring prior-missing areas instead of blindly following the global guidance.
This is particularly evident in Figure~\ref{fig:explore_exp_explain}(b1), where the local FE-TSP solver detours to explore distant areas (missing areas in the partial prior map) while guided by the global coverage path, and in Figure~\ref{fig:explore_exp_explain}(b2), where the UAV heads to the isolated viewpoints to cover the missing areas.
After covering the missing areas, the UAV continues to explore the remaining areas efficiently under the global guidance.

Provided with the deformed map, the registration process can no longer achieve the same high accuracy as with a precise map.
Nevertheless, the generated global guidance path remains effective despite certain positional errors.
The low accuracy of registration results is particularly evident in the early stages, when the local map is small and deformations in local structures can significantly impact the positional accuracy of distant prior guidepoints (Figure~\ref{fig:explore_exp_explain}(c1)).
However, since distant prior guidepoints only offer coarse guidance without affecting local path planning, the UAV can still effectively follow the global guidance path to explore the environment.
As exploration continues, the global guidance path is continuously updated with increasingly precise registration results as more point cloud information is obtained, thereby providing more accurate guidance (Figure~\ref{fig:explore_exp_explain}(c2)).

\section{LIMITATIONS AND DISCUSSIONS}\label{sec:limitations}

The main limitation of the proposed method lies in the potential failure of registration and inefficiency of exploration when the prior map is significantly deformed.
Because 2D maps provide only sparse information, significant deformations render the prior too unreliable for registration.
Different types of discrepancy also affect the performance of the proposed method differently.
When the deformation affects only local areas while the overall topology remains intact, the registration process can still succeed.
In our experiments, deformations primarily consisted of slight image warping or scale changes, which our method handled effectively.
However, for deformations like severely elongated or shortened corridors, the Scale-ICP process fails to achieve precise registration.
Moreover, if the global topological information is highly erroneous, the resulting guidance path may become suboptimal or even misleading, reducing exploration efficiency.
For example, unmapped roadblocks can alter the true topological structure and mislead the global guidance.
The main reason is that the proposed method relies on global topological information to generate optimal global routes and is sensitive to the topological correctness.
The framework has no mechanism for identifying such topological errors and abandoning the prior map, and may therefore suffer from reduced exploration efficiency in such cases.
Section~S7 of the supplementary material evaluates this failure mode on five controlled topological faults, and identifies the case in which the prior makes the planner perform worse than a prior-free baseline.

Another limitation is the exploration ability in highly vertical environments, since the prior map lacks vertical information. We can only handle prior maps that represent the environment as a 2D top-down sketch, while prior maps that illustrate vertical features (e.g., multi-floor maps) are not supported.
For prior representations that can offer 3D information such as semantic maps or stacked 2D maps, our method still needs further adaptation.

Future work will improve robustness against deformation by incorporating semantic features and temporal consistency into the registration process, and by designing more resilient global guidance strategies that can tolerate topological errors.
We also plan to extend our method to handle more complex environments such as multi-floor scenarios and diverse prior formats, including 3D or semantic maps.

\section{CONCLUSION}

This paper presents a prior-guided LiDAR-based UAV exploration framework for large-scale environments with sparse, unaligned, and potentially discrepant 2D prior maps. The framework combines online 2D-3D point cloud registration with hierarchical viewpoint planning to generate efficient local exploration paths under uncertain prior-map guidance.

For registration, we introduced the GeoContext descriptor for single-frame candidate retrieval and a multi-frame geometric verification strategy to establish robust correspondences between online LiDAR observations and prior maps. The resulting transformations are further refined using Scale-ICP, enabling reliable registration under partial observations and moderate map discrepancies.
For planning, we developed a hierarchical strategy that first determines a global traversal sequence over prior guidepoints and then generates local viewpoint traversal paths consistent with the selected guidance. The proposed framework incorporates registration uncertainty into the planning process and maintains effective exploration behavior under incomplete or deformed priors.
Simulation benchmarks and field experiments demonstrated reduced exploration time and flight distance compared with baseline methods, with consistent performance across precise, partial, and deformed prior maps.

The robustness demonstrated here covers priors that are incomplete or mildly deformed, for which the global topology remains intact and the guidance stays useful. It does not extend to priors whose topology is itself erroneous, where the guidance may become misleading rather than merely imprecise, nor to highly vertical environments that a 2D top-down map cannot represent. Future work will focus on enhancing robustness against deformations, incorporating semantic features, and extending the framework to support 3D prior maps.

\bibliographystyle{IEEEtran}
\bibliography{reference}

\begin{IEEEbiography}[{\includegraphics[width=1in,height=1.25in,clip,keepaspectratio]{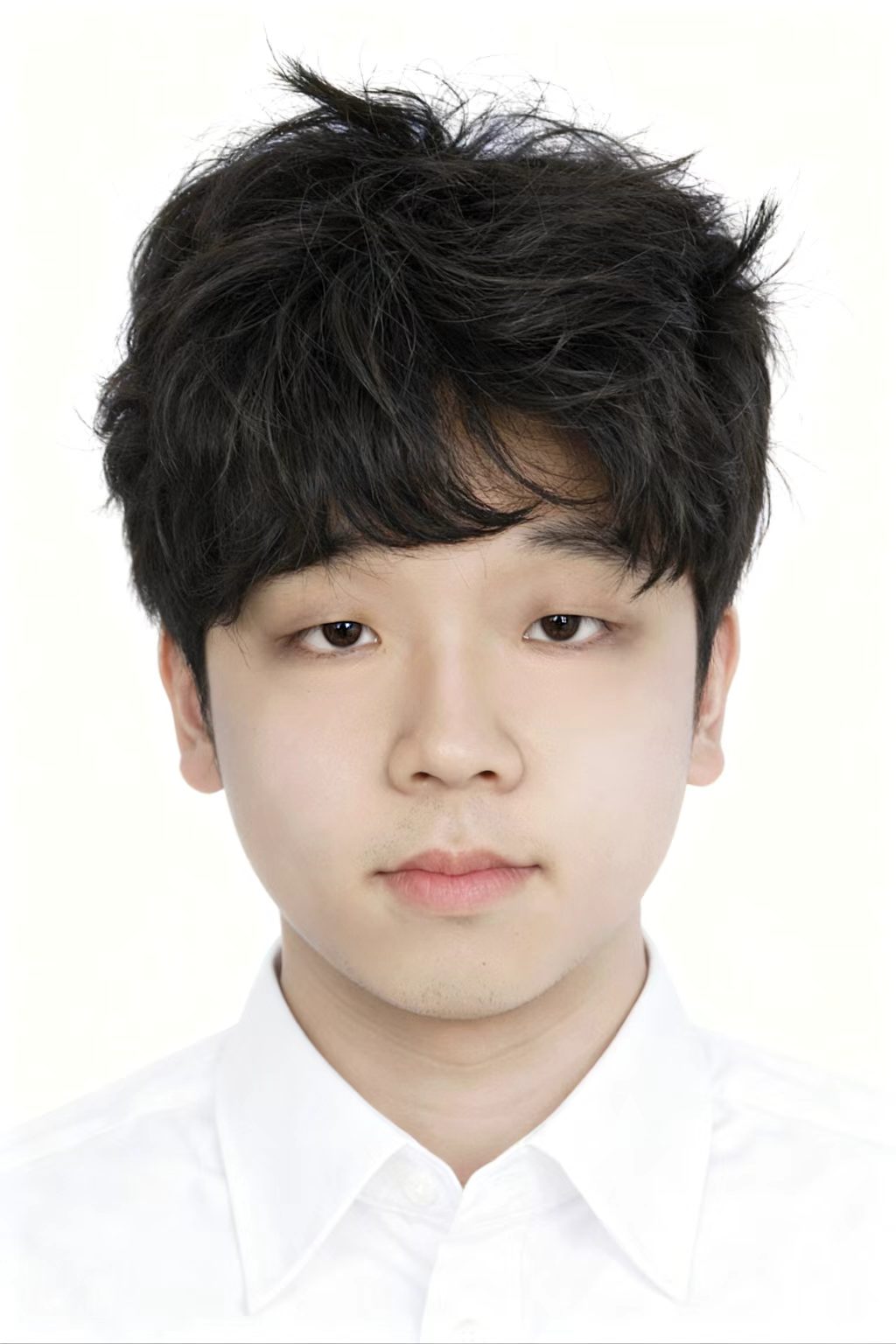}}]
{Tiancheng Lai}~received the B.Eng. degree in Automation and Engineering from Harbin Institute of Technology, Shenzhen, China, in 2024.  He is currently pursuing the M.S. degree in control science and engineering in the FastLab at Zhejiang University.
His research interests include autonomous exploration and reinforcement learning for robotics.
\end{IEEEbiography}

\begin{IEEEbiography}[{\includegraphics[width=1in,height=1.25in,clip,keepaspectratio]{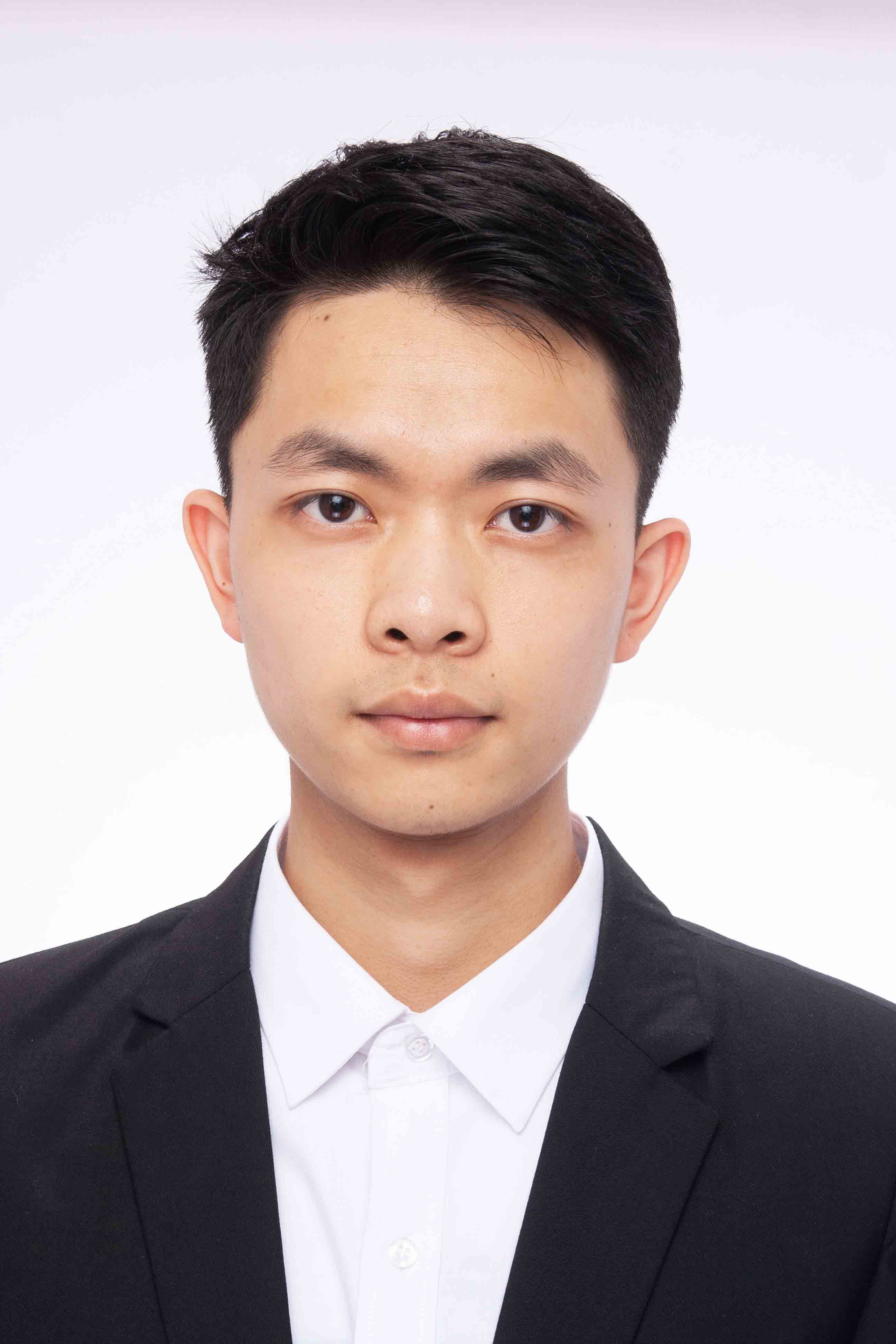}}]
{Yuman Gao}~received the B.Eng. degree in control science and engineering from Zhejiang University, Hangzhou, China, in 2021, where he was also a member of the Advanced Honor Class of Engineering Education (ACEE) at Chu Kochen Honors College. He received the Ph.D. degree in robotics from Zhejiang University in 2026, advised by Prof. Fei Gao at the FAST Lab. During his Ph.D. studies, he was a visiting scholar with the Hybrid Robotics Group and the Berkeley Artificial Intelligence Research (BAIR) Lab, University of California, Berkeley, CA, USA, advised by Prof. Koushil Sreenath.
His research interests include motion planning and autonomous navigation for multi-robot systems.
\end{IEEEbiography}

\begin{IEEEbiography}[{\includegraphics[width=1in,height=1.25in,clip,keepaspectratio]{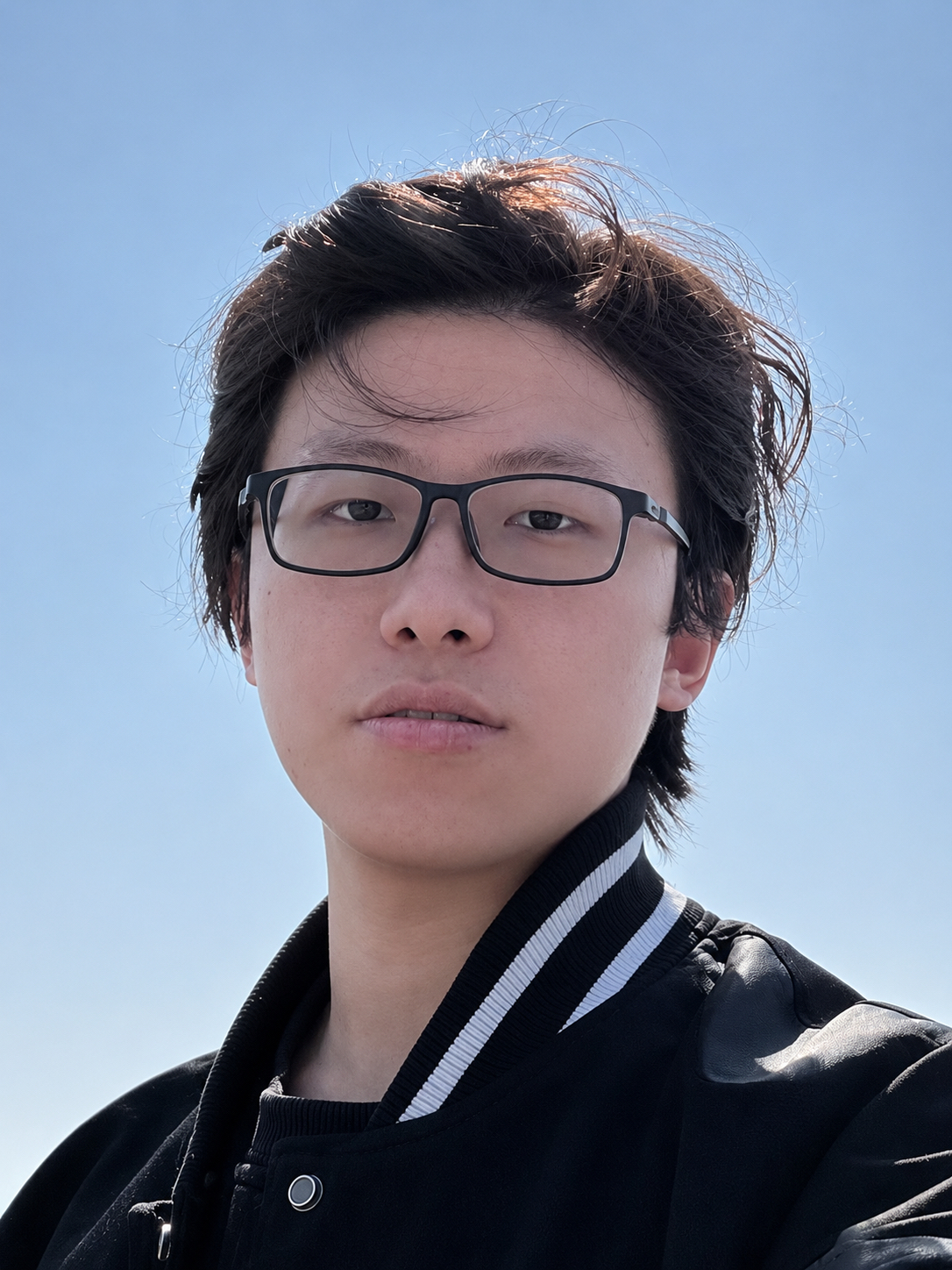}}]
{Xiangyu Li}~received the B.S. degree in electronic information engineering from China Jiliang University, Hangzhou, China, in 2023. He is currently working toward the M.S. degree in
electronic information with Zhejiang University, Hangzhou.
He is a Member of the FAST Lab with Zhejiang University, Hangzhou. His research interests include mechatronic system design, the control of novel robotic platforms, reinforcement
learning, and world models for robotics.
\end{IEEEbiography}

\begin{IEEEbiography}[{\includegraphics[width=1in,height=1.25in,clip,keepaspectratio]{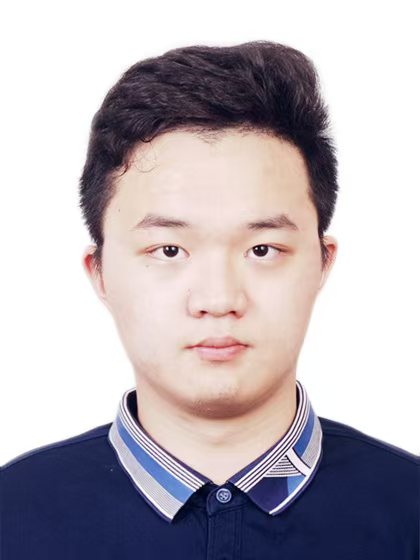}}]
{Ruitian Pang}~received the B.S. degree in electronic information engineering from China Jiliang University, Hangzhou, China, in 2023. He is currently working toward the M.S. degree in
electronic information with Zhejiang University, Hangzhou.
He is a Member of the FAST Lab with Zhejiang University, Hangzhou. His research interests include mechatronic system design, the control of novel robotic platforms, reinforcement
learning, and world models for robotics.
\end{IEEEbiography}

\begin{IEEEbiography}[{\includegraphics[width=1in,height=1.25in,clip,keepaspectratio]{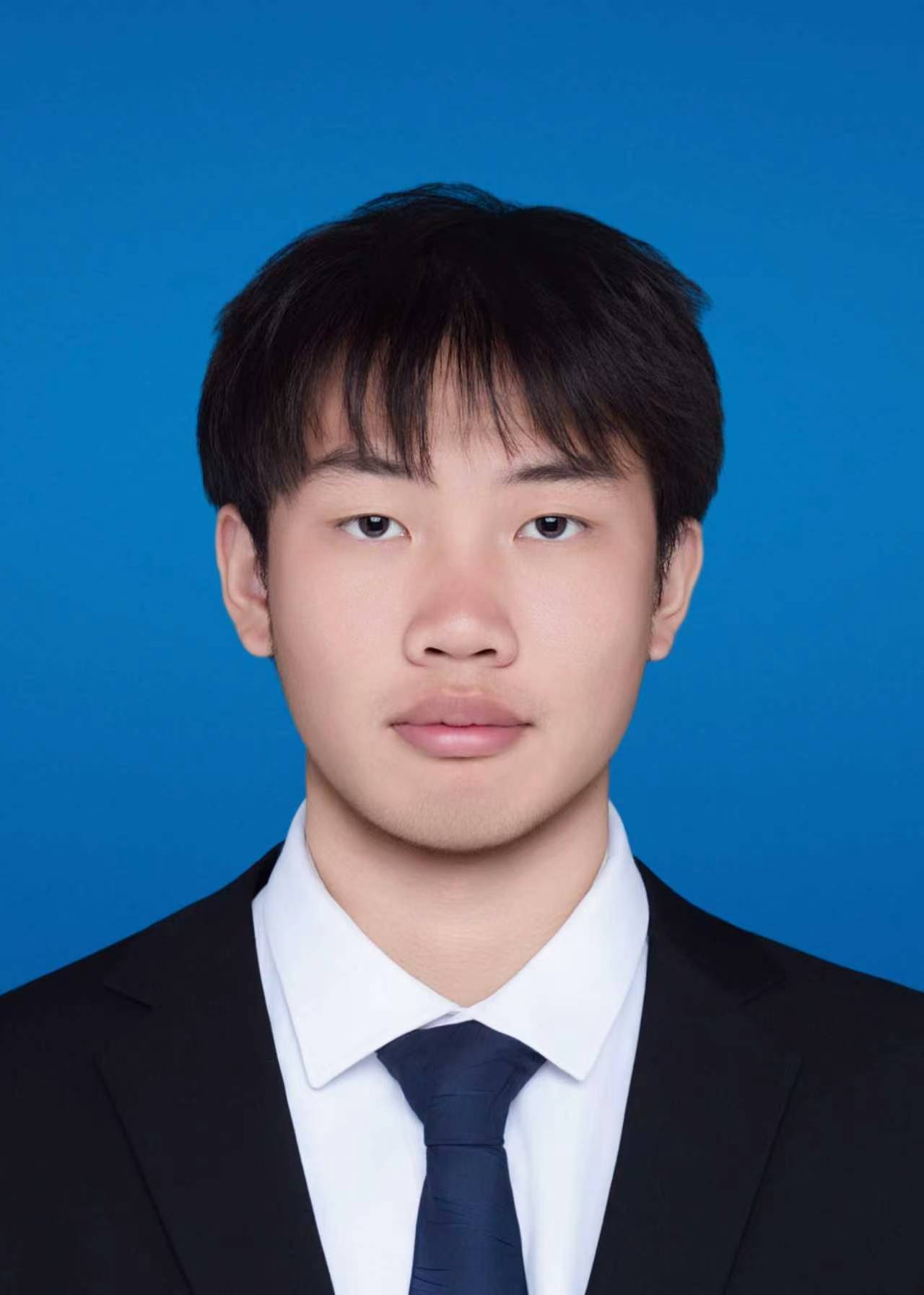}}]
{Xingpeng Wang}~is currently pursuing the M.S. degree in the College of Control Science and Engineering, Zhejiang University, Hangzhou, China. His research interests include robotic localization and state estimation, high-precision GNSS positioning, multi-sensor fusion, and autonomous navigation.
\end{IEEEbiography}

\begin{IEEEbiography}[{\includegraphics[width=1in,height=1.25in,clip,keepaspectratio]{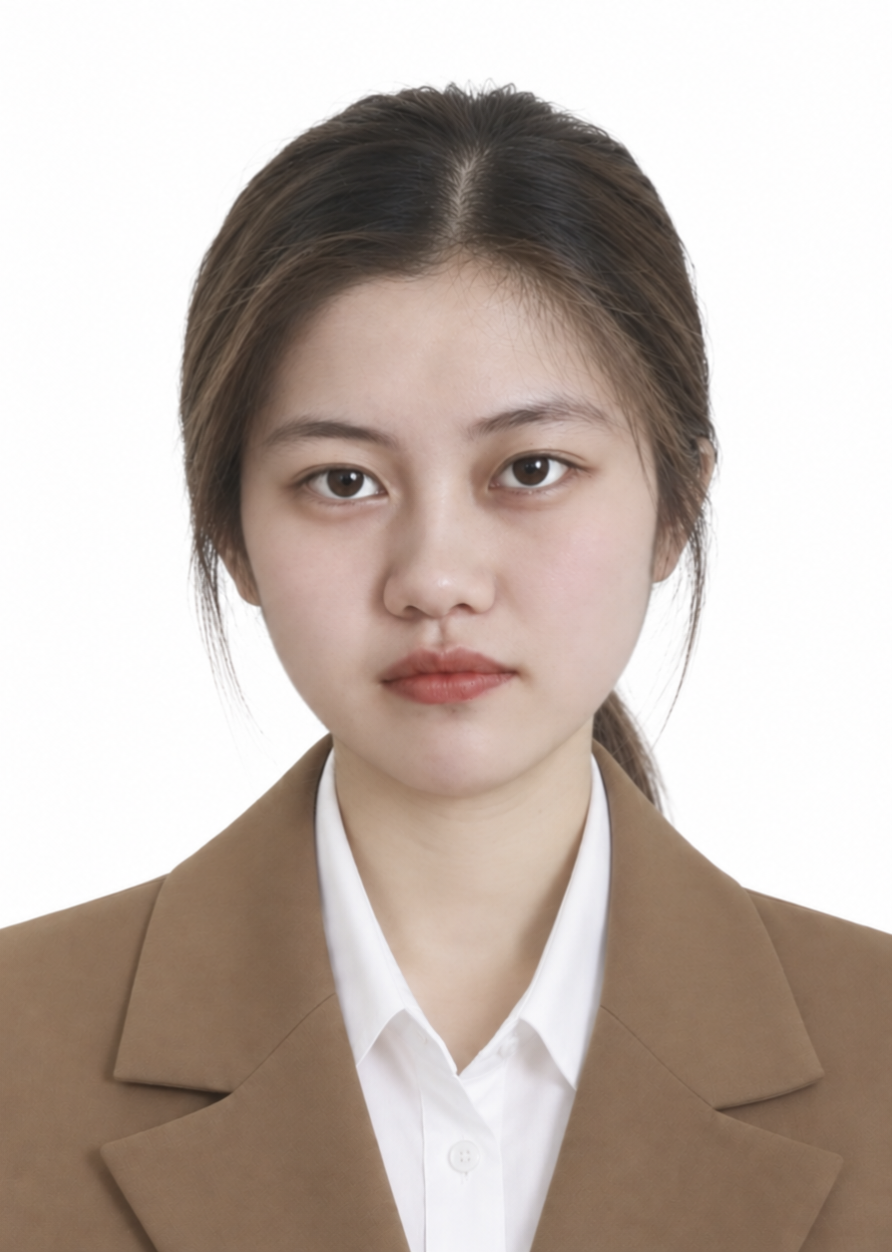}}]
{Siqi Shen}~received the B.E. degree in agricultural engineering from Zhejiang University, Hangzhou, China, in 2025. She is currently pursuing the M.E. degree in electronic information at the College of Control Science and Engineering, Zhejiang University, where she is a member of the FASTLab. Her current research interests include motion planning and robot learning.
\end{IEEEbiography}

\begin{IEEEbiography}[{\includegraphics[width=1in,height=1.25in,clip,keepaspectratio]{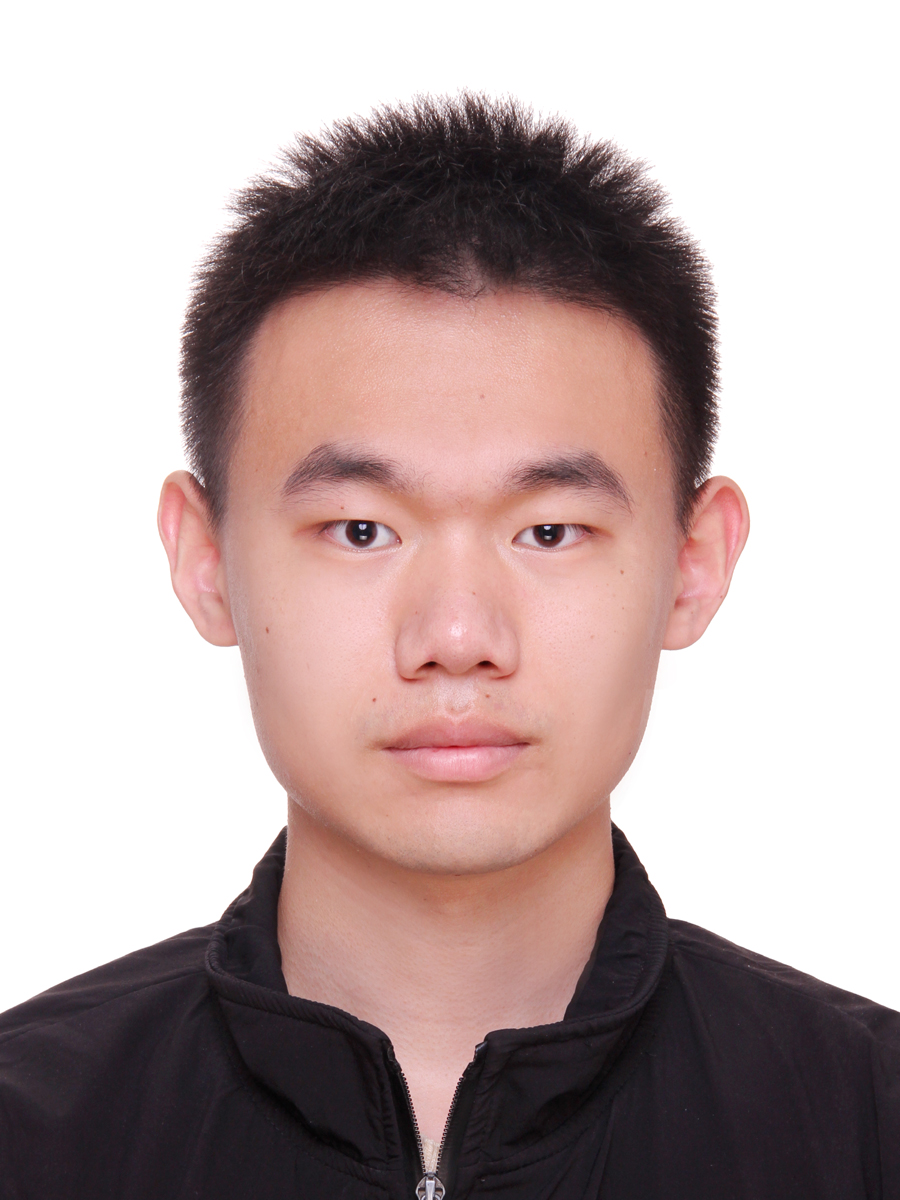}}]
{Mengke Zhang}~received the Doctor of Engineering degree in Electronic Information from Zhejiang University, Hangzhou, China, in 2026. He conducted his doctoral research at the Fast Lab, Zhejiang University. His research interests include optimization and motion planning.
 \end{IEEEbiography}

 \begin{IEEEbiography}[{\includegraphics[width=1in,height=1.25in,clip,keepaspectratio]{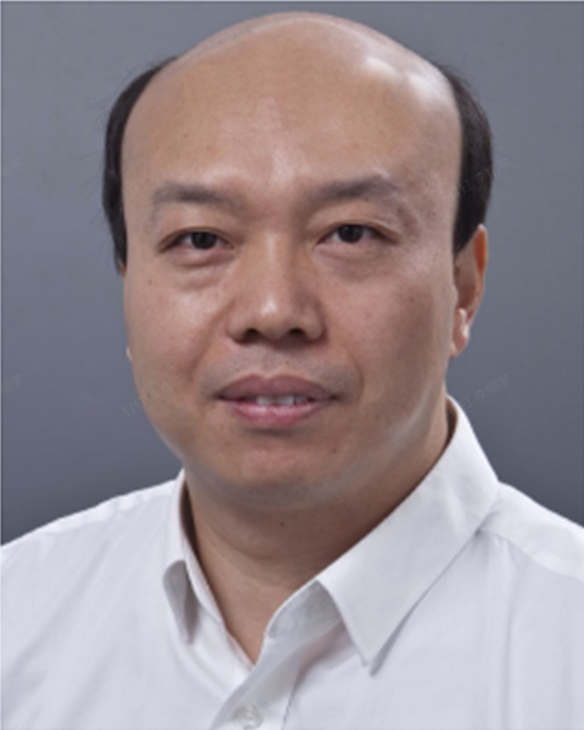}}]
{Zaisheng Pan}~received the B.S.and master's degrees in industrial automation from Zhejiang Univer-sity, Zhejiang, China, in 1993 and 1996, respectively.He is currently a Senior Engineer with the Instituteof Cyber-Systems and Control. Zheiiang University.His main research interests include the Internet ofThings, industrial big data, and industrial networks.
 \end{IEEEbiography}

\begin{IEEEbiography}[{\includegraphics[width=1in,height=1.25in,clip,keepaspectratio]{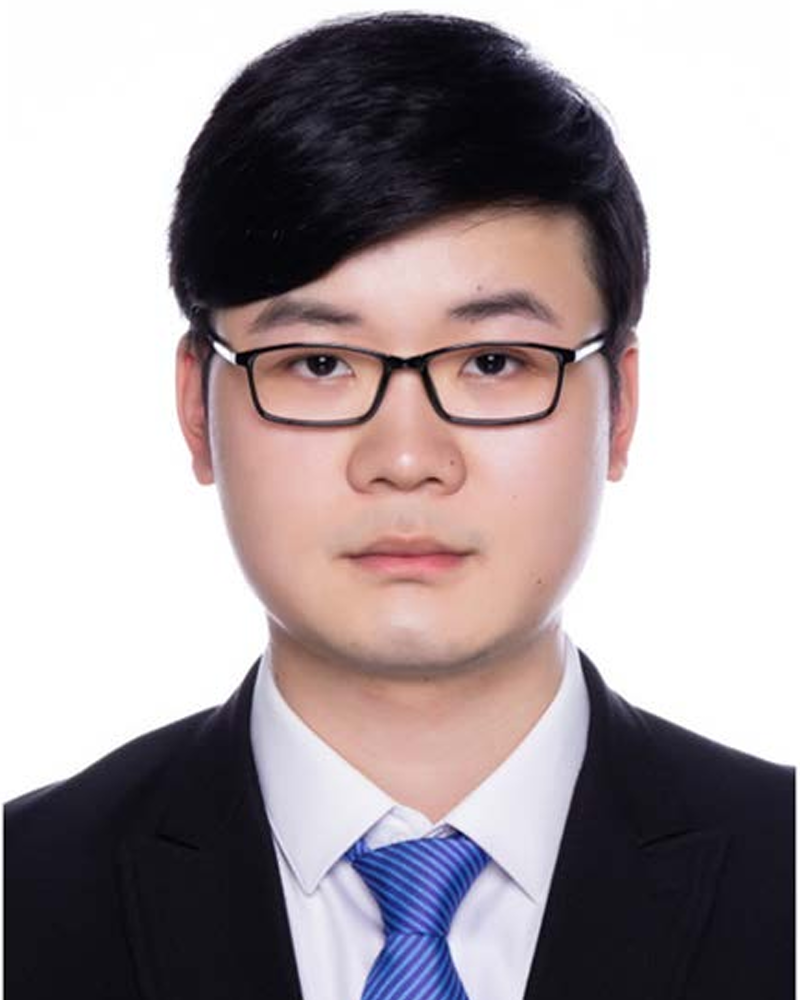}}]
{Fei Gao}~(Member, IEEE)~received the Ph.D. degree
in electronic and computer engineering from the
Hong Kong University of Science and Technology,
Hong Kong, in 2019. He is currently a tenured Associate Professor in the College of Control Science
and Engineering, Zhejiang University, Hangzhou,
China, where he leads the Aerial Embodied Intelligence Laboratory. His research interests include
aerial robots, autonomous navigation, motion planning, embodied intelligence, and swarm intelligence.
He is also the founder of Differential Robotics.
 \end{IEEEbiography}

 \begin{IEEEbiography}[{\includegraphics[width=1in,height=1.25in,clip,keepaspectratio]{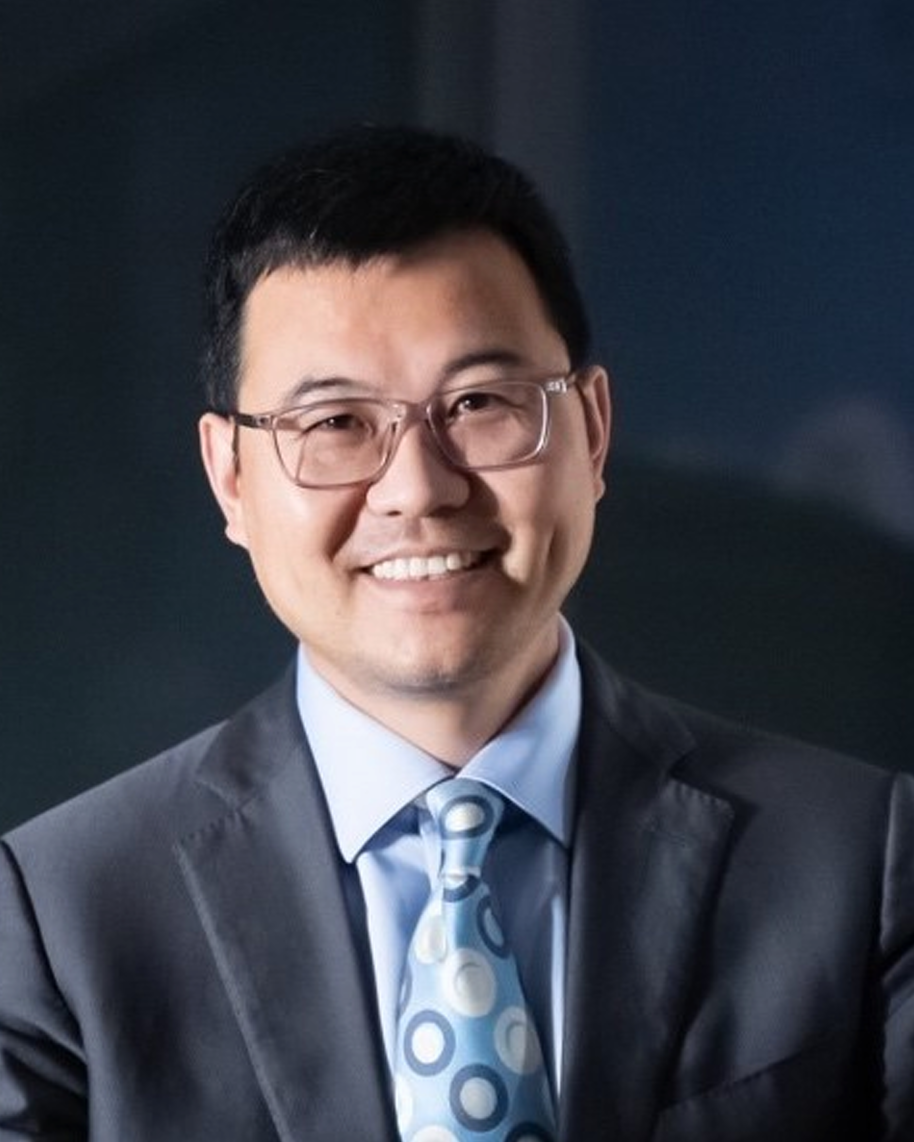}}]
{Chao Xu}~(Senior Member, IEEE)~received the Ph.D. degree in mechanical engineering from Lehigh University in 2010. He is currently the Associate Dean and a Professor with the College of Control Science and Engineering, Zhejiang University. He is the Inaugural Dean of ZJU Huzhou Institute. His research expertise is flying robotics and control-theoretic learning. He has published over 100 articles in international journals, including Science Robotics and Nature Machine Intelligence. He will join the organization committee of the IROS-2025 in Hangzhou.
 \end{IEEEbiography}

\begin{IEEEbiography}[{\includegraphics[width=1in,height=1.25in,clip,keepaspectratio]{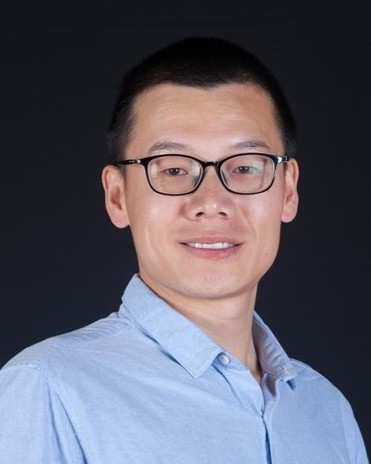}}]
{Yanjun Cao}~received his Ph.D. degree in computer and software engineering from the University of Montreal, Polytechnique Montreal, Canada, in 2020. He is currently an associate researcher at the Huzhou Institute of Zhejiang University, as a PI in the Center of Swarm Navigation. He leads the Field Intelligent Robotics Engineering group of the Field Autonomous System and Computing Lab. His research focuses on key challenges in multi-robot systems, such as collaborative localization, autonomous navigation, perception and communication.
 \end{IEEEbiography}

\clearpage

\makeatletter
\@twosidefalse
\includepdf[pages=-,offset=5.3pc -6.75pc]{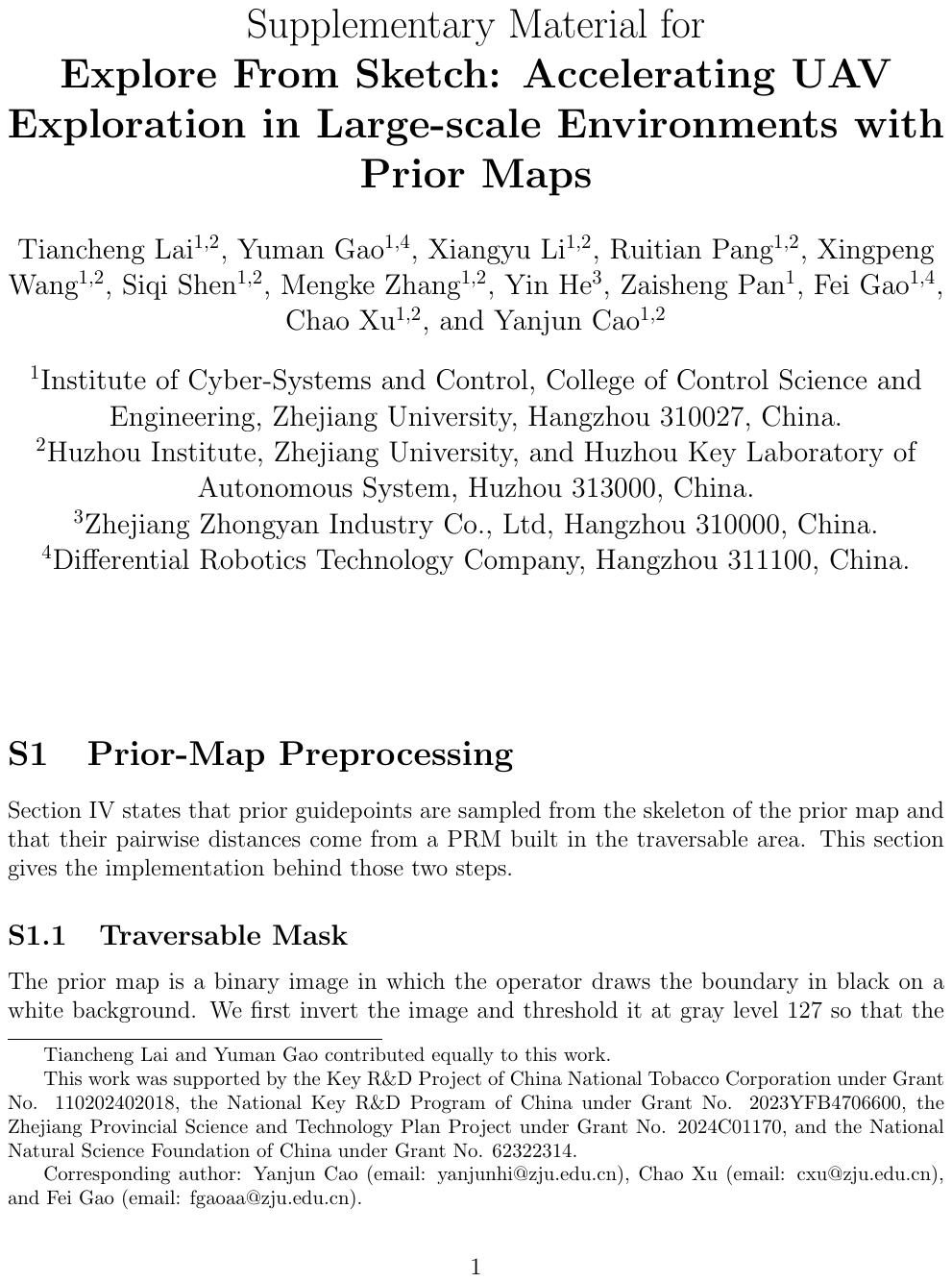}
\@twosidetrue
\makeatother

\end{document}